\documentclass[10pt]{article}
\usepackage[preprint]{tmlr}  %

\usepackage{amsmath,amsfonts,bm}

\def\eqref#1{equation~\ref{#1}}

\def\1{\bm{1}}

\def\rx{{\textnormal{x}}}
\def\ry{{\textnormal{y}}}

\DeclareMathAlphabet{\mathsfit}{\encodingdefault}{\sfdefault}{m}{sl}
\SetMathAlphabet{\mathsfit}{bold}{\encodingdefault}{\sfdefault}{bx}{n}

\usepackage{graphicx}
\usepackage{booktabs}
\usepackage[section]{placeins}  %
\usepackage{tikz}
\usepackage{pgfplots}
\pgfplotsset{compat=1.18}
\usetikzlibrary{arrows.meta, positioning, fit, backgrounds, calc}
\usepackage{hyperref}
\usepackage{url}

\newcommand{\method}{GaussianRenderFormer}

\title{GaussianRenderFormer: Transformer-based Rendering of 3D Gaussian Splats}

\author{\name Shahaf Valencia Levy \email shahaf.valenciale@mail.huji.ac.il \\
      \addr The Hebrew University of Jerusalem
      \AND
      \name Sagie Benaim \email sagie.benaim@mail.huji.ac.il \\
      \addr The Hebrew University of Jerusalem}

\def\month{MM}
\def\year{YYYY}
\def\openreview{\url{https://openreview.net/forum?id=XXXX}}

\begin{document}

\maketitle

\newcommand{\teaserimg}[1]{\raisebox{-.5\height}{\includegraphics[width=0.13\textwidth]{figures/teaser/#1}}}
\begin{figure}[!ht]
  \centering
  \setlength{\tabcolsep}{1.2pt}
  \renewcommand{\arraystretch}{0.6}
  \footnotesize
  \begin{tabular}{@{}c@{\hspace{3pt}}cccccc@{\hspace{5pt}}c@{}}
    \rotatebox[origin=c]{90}{Reference} &
    \teaserimg{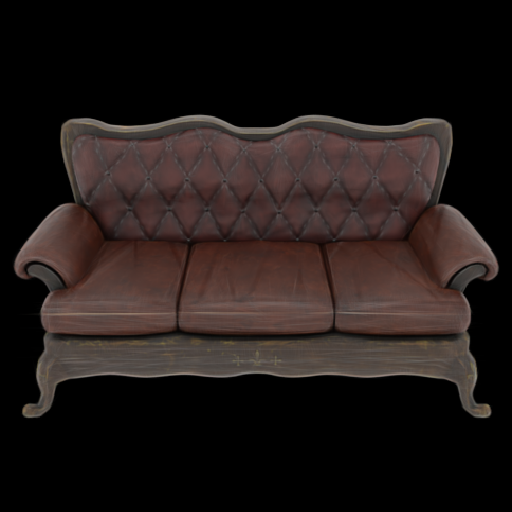} & \teaserimg{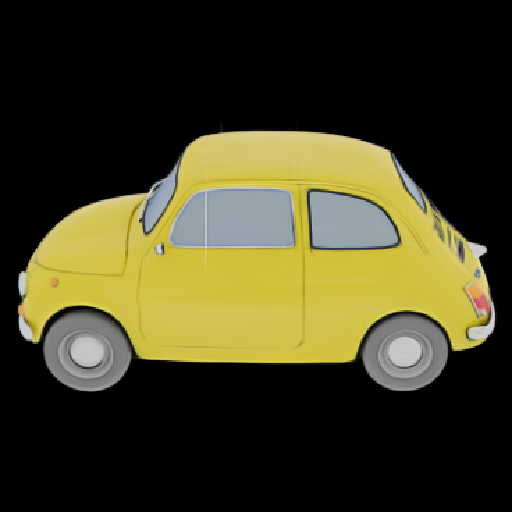} & \teaserimg{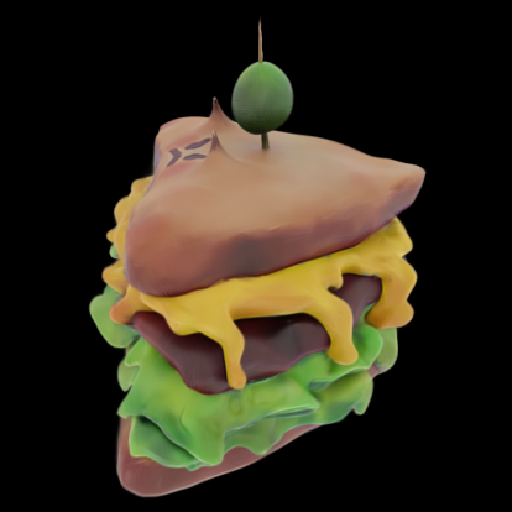} &
    \teaserimg{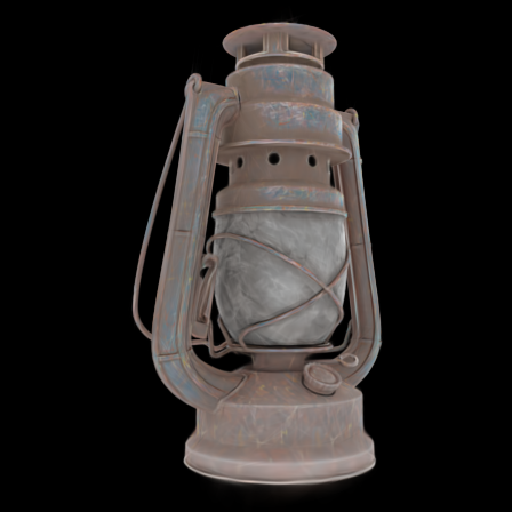} & \teaserimg{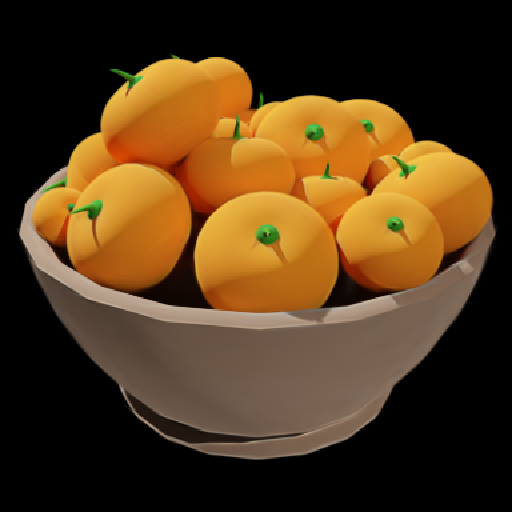} & \teaserimg{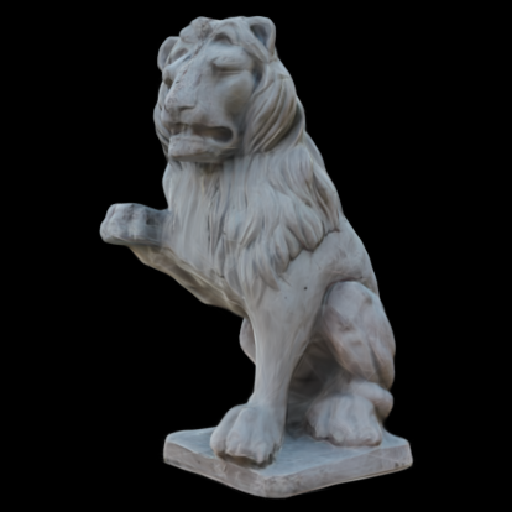} &
    \teaserimg{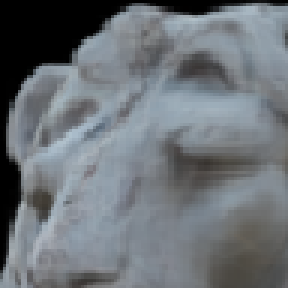} \\[24pt]
    \rotatebox[origin=c]{90}{Ours} &
    \teaserimg{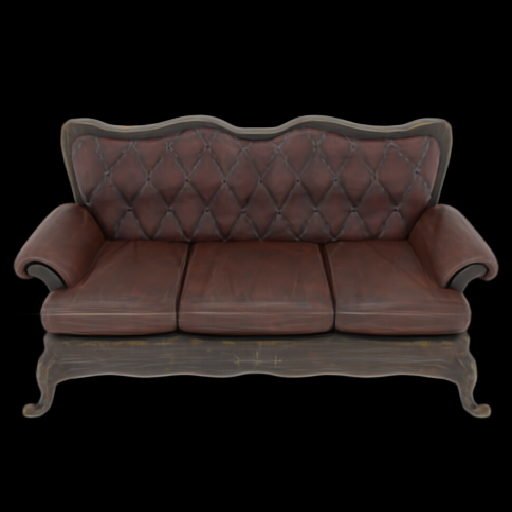} & \teaserimg{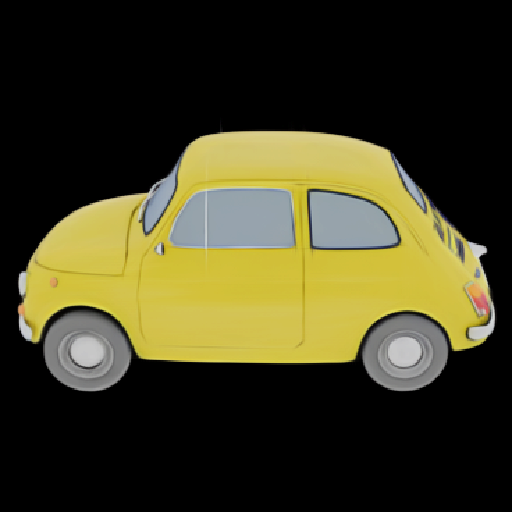} & \teaserimg{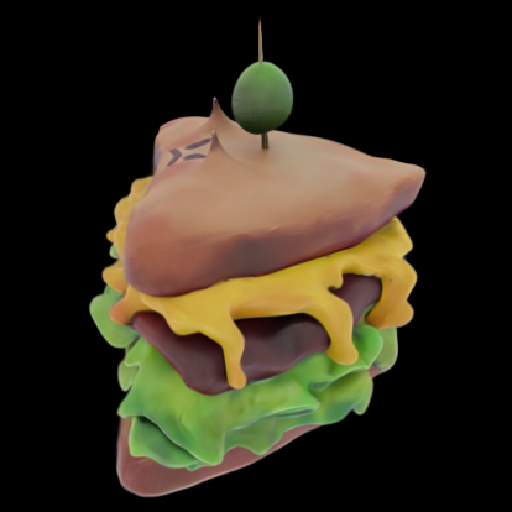} &
    \teaserimg{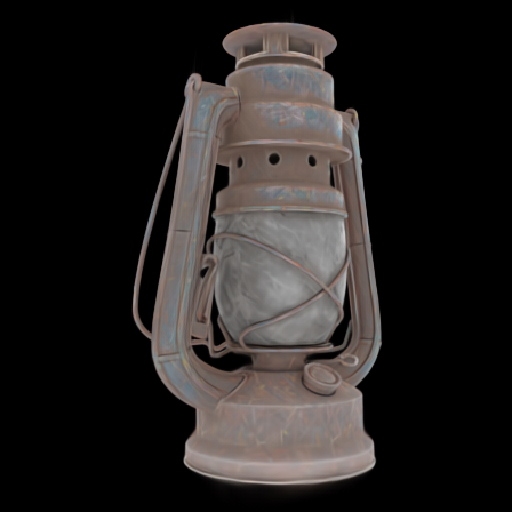} & \teaserimg{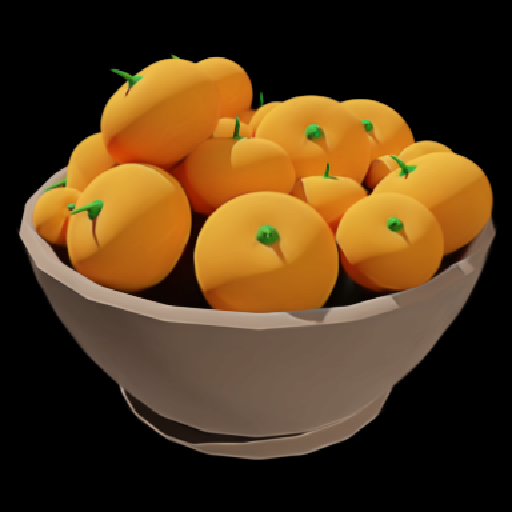} & \teaserimg{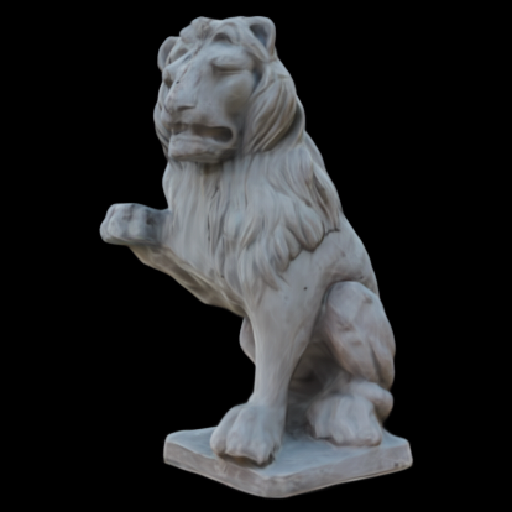} &
    \teaserimg{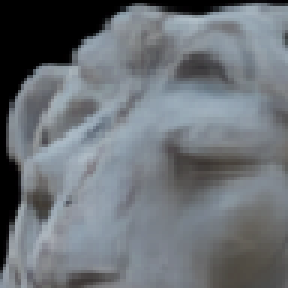} \\[24pt]
  \end{tabular}
  \caption{\method{} renders 3D Gaussian splats with a neural network, without a rasterizer. Top: the reference, a
  rasterization of each object's full splat. Bottom (ours): \method{} from a 20,000-Gaussian input, on six objects never seen
  in training, chosen by hand for variety. Right: a $3\times$ zoom of the most detailed region of the last object,
  where the model softens the stone's texture.}
  \label{fig:teaser}
  \vspace{-8pt}  %
\end{figure}

\begin{abstract}
Can a neural network learn to render 3D Gaussian splats on its own, with no rasterizer in the renderer? We present GaussianRenderFormer, a transformer that takes a set of Gaussians and a camera and outputs the image directly. Each Gaussian becomes one token in a view-independent encoder, and a view-dependent decoder turns small patches of camera rays into pixels by attending to those tokens. Trained on 26,820 Objaverse objects pruned to 20,000 Gaussians each, GaussianRenderFormer renders 300 unseen objects at 38.08 dB PSNR against the full, unpruned splats, while rasterizing the same pruned input reaches 44.02 dB. The gap is not one of generalization, and our diagnostics are consistent with a bottleneck in the view decoder: telling it the projected footprint of each Gaussian, which it could in principle infer, adds 1.4 dB. Under-training is not ruled out. Code, trained models and videos: \url{https://svlwoof.github.io/gaussianrenderformer}.

\end{abstract}

\section{Introduction}
\label{sec:introduction}

3D Gaussian Splatting (3DGS)~\citep{kerbl20233dgs} has become a common way to represent real scenes captured from
photos. A 3DGS scene is a set of Gaussians, each with a position, scale, rotation, color and opacity, and every image
of it is produced by a rasterizer: a hand-written algorithm that projects the Gaussians onto the image plane and blends
them. In this paper we replace that algorithm with a neural network.\footnote{Large language models assisted
both the research behind this paper, including code and the analysis of experiments, and its writing. The authors
take full responsibility for its content.} Given a set of Gaussians and a camera, our model
outputs the image directly, without per-scene optimization and with no rasterizer in the renderer
(Figure~\ref{fig:teaser}).

Learned rendering has made significant progress for other kinds of scene descriptions.
RenderFormer~\citep{zeng2025renderformer} renders triangle meshes with global illumination using a transformer, and
RenderFormer-V2~\citep{zeng2026renderformerv2} extends it to more kinds of primitives, lights and media. For Gaussians,
to our knowledge, no learned renderer exists. Methods that reconstruct, refine, relight or analyze Gaussians still hand the final image to a hand-written renderer, usually the 3DGS rasterizer itself
(Section~\ref{sec:related_work}).

A learned renderer of Gaussians could offer what a rasterizer does not. It could learn light transport from data, as
RenderFormer does for meshes, where relightable Gaussian methods rely on a hand-written shading model and approximate
the indirect light~\citep{gao2024relightable, liang2024gsir, chen2025gigs}. It could render captured splats together
with meshes and lights in one model, as RenderFormer-V2 does for meshes, media and lights. It has no sorting step, so it
avoids the popping that sorting causes as the camera moves. And it could learn a prior that fills in what its input lacks. None of this is
shown here: a network first has to render Gaussians at all.

It is also not obvious that the approach that works for meshes carries over to Gaussians. Gaussians have no surface
or material, are semi-transparent and overlap heavily. An object typically needs tens of
thousands, while RenderFormer was trained on scenes of up to 4,096 triangles. A learned renderer has to
resolve which Gaussians are visible and how they blend at every pixel, which a rasterizer does by sorting
them.

We present \method{}, a transformer that renders Gaussians directly. Each Gaussian becomes one token, and a
view-independent encoder lets these tokens exchange information. A view-dependent decoder then represents each small
patch of the image by a token built from the camera rays passing through it. These tokens attend to the Gaussians and
are told where in the image each Gaussian lands and how large it appears there, both computed per Gaussian, while
everything per pixel is learned. We start from RenderFormer's pretrained weights and train on 26,820
objects from Objaverse~\citep{deitke2023objaverse, ma2024shapesplat}, each pruned to 20,000 Gaussians.

We treat this as a feasibility study at the object level. On 300 objects never seen in training, \method{} reaches
38.08 dB PSNR against renders of the full, unpruned splats, while rasterizing the same pruned input reaches 44.02 dB
(Section~\ref{sec:experiments}). For context, LVSM~\citep{jin2025lvsm}, the closest rasterizer-free transformer,
reaches 32.36 dB with its decoder-only variant on Google Scanned Objects at 512 pixels from four input images, and GS-LRM~\citep{zhang2024gslrm},
which rasterizes, 30.52 dB, both as reported in LVSM's Table~1. They start from images and we from the splat, so the numbers are not directly comparable,
but they show the range in which learned renderers of objects work. The gap to rasterization is not one of
generalization, and our diagnostics are consistent with a bottleneck in the view decoder, though under-training is
not ruled out.

Our contributions are (1) \method{}, to our knowledge the first feed-forward neural renderer for 3D Gaussian splats
with no rasterizer in the renderer, (2) three changes that adapt a mesh renderer's view decoder to Gaussians, a 2D
rotary embedding of where each Gaussian lands in the image, its projected footprint on the attention keys and smaller
patches, each ablated at moderate scale and again at full scale, together with a schedule that trains the new input
layer first, and (3) an analysis of the remaining gap to rasterization, which is not a generalization gap and is
consistent with a bottleneck in the view decoder, and of how the model behaves outside its training setup. The code and the trained models are available at \url{https://github.com/SVLwoof/gaussianformer}.

\section{Related Work}
\label{sec:related_work}

\noindent\textbf{Rendering 3D Gaussians.} 3D Gaussian Splatting (3DGS) renders a scene by projecting each Gaussian onto the image plane and alpha-blending the
sorted projections, a process evolved from EWA splatting~\citep{zwicker2002ewa}. Later works improve this rasterizer
against aliasing~\citep{yu2024mipsplatting}, with flat disks~\citep{huang20242dgs} or with per-pixel
sorting~\citep{radl2024stopthepop}, and gsplat~\citep{ye2025gsplat}\footnote{\url{https://github.com/nerfstudio-project/gsplat}}
is the optimized open-source implementation we use to render our ground truth. Other works ray-trace Gaussians~\citep{moenneloccoz20243dgrt}
or replace them with ellipsoids that can be ray-traced exactly~\citep{mai2025ever}, and Sort-free Gaussian Splatting~\citep{hou2025sortfree} learns
weights that replace the sorting but keeps the projection and blending of a rasterizer. All of these are hand-written
algorithms, while \method{} renders with a network.

\noindent\textbf{Learned renderers for explicit scenes.} The works closest to ours render an explicit description of a scene with a network instead of an algorithm.
RenderFormer renders triangle meshes with global illumination (Section~\ref{sec:background_renderformer}), and
RenderFormer-V2~\citep{zeng2026renderformerv2} and RenderFormer++~\citep{du2026renderformerpp} extend it to more primitives, lights and media and to
object-level tokens, but none of them takes Gaussians. Pointersect~\citep{chang2023pointersect} renders point clouds
with a transformer per ray, \citet{xu2026lighttransport} predict indirect illumination from sampled points, and
RenderNet~\citep{nguyenphuoc2018rendernet} rendered voxel grids and the Generative Query
Network~\citep{eslami2018gqn} a learned scene code. Most other neural rendering methods keep a conventional renderer and
are optimized per scene: deferred neural rendering~\citep{thies2019deferred} and neural point-based
graphics~\citep{aliev2020npbg, ruckert2022adop, franke2024trips} rasterize learned features before a network produces
the image, and Point-NeRF~\citep{xu2022pointnerf} ray-marches neural points. To our knowledge, \method{}
is the first model to render a given set of 3D Gaussians with a network alone.

\noindent\textbf{Networks around Gaussian splatting.} Most learned models that involve 3D Gaussians produce them rather than render them. Feed-forward reconstruction
methods predict Gaussians from a set of images in a single pass and then rasterize
them~\citep{szymanowicz2024splatter, charatan2024pixelsplat, chen2024mvsplat, tang2024lgm, zhang2024gslrm,
xu2025freesplatter, chen2025longlrm, jiang2025anysplat}. Other models take Gaussians as input tokens for other tasks,
such as self-supervised pretraining~\citep{ma2024shapesplat}, scene understanding~\citep{li2025scenesplat},
generative modeling~\citep{gao2025can3tok} or refining the Gaussians themselves~\citep{chen2025splatformer}, and
relightable methods shade them with a hand-written model~\citep{gao2024relightable, jiang2024gaussianshader,
liang2024gsir}, which GI-GS~\citep{chen2025gigs} extends with an approximation of the indirect light. Wherever these
methods produce an image, a rasterizer still renders it, while for \method{} the Gaussians are the input and the image is the output.

\noindent\textbf{View synthesis without explicit geometry.} Another line of work synthesizes new views without any explicit 3D representation, mostly with transformers. The Scene
Representation Transformer~\citep{sajjadi2022srt} decodes each pixel from a ray that attends to latent tokens of posed
images, Light Field Networks~\citep{sitzmann2021lfn}, Light Field Neural Rendering~\citep{suhail2022lfnr} and
GNT~\citep{varma2023gnt} map rays to colors directly or with attention along each ray, and LVSM~\citep{jin2025lvsm}
and RayZer~\citep{jiang2025rayzer} scale this idea to large models. These are the closest relatives of \method{}: most of them decode
scene tokens with ray queries, and none uses a rasterizer. Their scene, however, is given only through images, while ours
is explicit Gaussians, so our output can be compared directly with rasterizing the same input.

\section{Background}
\label{sec:background}

\subsection{3D Gaussian Splatting}
\label{sec:background_3dgs}

3DGS represents a scene as a set of $N$ Gaussians, fitted to photos by gradient descent. Gaussian $i$ has a center
$\boldsymbol{\mu}_i \in \mathbb{R}^3$, a covariance $\boldsymbol{\Sigma}_i = \mathbf{R}_i \mathbf{S}_i
\mathbf{S}_i^\top \mathbf{R}_i^\top$ built from a rotation $\mathbf{R}_i$ (a unit quaternion $\mathbf{q}_i$) and
per-axis scales $\mathbf{S}_i = \mathrm{diag}(\mathbf{s}_i)$, an opacity $o_i \in [0, 1]$ and a color $\mathbf{c}_i$.
To render an image, the rasterizer projects each Gaussian to a 2D Gaussian with center $\boldsymbol{\mu}'_i$ and
covariance $\boldsymbol{\Sigma}'_i = \mathbf{J} \mathbf{W} \boldsymbol{\Sigma}_i \mathbf{W}^\top \mathbf{J}^\top$, where
$\mathbf{W}$ is the camera's rotation and $\mathbf{J}$ the Jacobian of the projection, sorts the Gaussians by depth
and blends them front to back at every pixel $\mathbf{x}$:
\begin{equation}
    \mathbf{C}(\mathbf{x}) = \sum_{i=1}^{N} \mathbf{c}_i \, \alpha_i(\mathbf{x}) \prod_{j=1}^{i-1} \bigl(1 -
    \alpha_j(\mathbf{x})\bigr), \qquad
    \alpha_i(\mathbf{x}) = o_i \exp\Bigl(-\tfrac{1}{2} (\mathbf{x} - \boldsymbol{\mu}'_i)^\top
    \boldsymbol{\Sigma}'^{-1}_i (\mathbf{x} - \boldsymbol{\mu}'_i)\Bigr).
    \label{eq:splatting}
\end{equation}
This is the computation \method{} replaces. Of the color, which 3DGS stores as spherical harmonics that change with
the viewing direction, we use only the view-independent part, a single RGB value, in the input and the ground truth.

\subsection{RenderFormer}
\label{sec:background_renderformer}

RenderFormer renders a triangle mesh with a transformer in two stages. In the view-independent stage, each triangle
becomes one token, built by a linear layer from its vertex normals and its material, and 12 layers of self-attention
over these tokens and 16 register tokens model how light travels between triangles. Positions enter only through a
rotary position embedding~\citep{su2021roformer} of the triangles' vertices in the attention. In the view-dependent stage, each patch of
$8 \times 8$ pixels becomes one token, built by a linear layer from the directions of its 64 rays. Six layers of
cross-attention to the triangle tokens, with the triangles' positions in the camera's
frame, each followed by self-attention among the ray-bundle tokens, gather what every patch sees, and a DPT
head~\citep{ranftl2021dpt} turns them into pixels of high dynamic range radiance, encoded as $\log_{10}(1 + x)$.
RenderFormer is trained on scenes of up to 4,096 triangles, first at 256 and then at 512 pixels.
\method{} keeps this layout and its pretrained weights (Section~\ref{sec:method}).

\providecommand{\todo}[1]{\textcolor{red}{[TODO: #1]}}
\section{Method}
\label{sec:method}

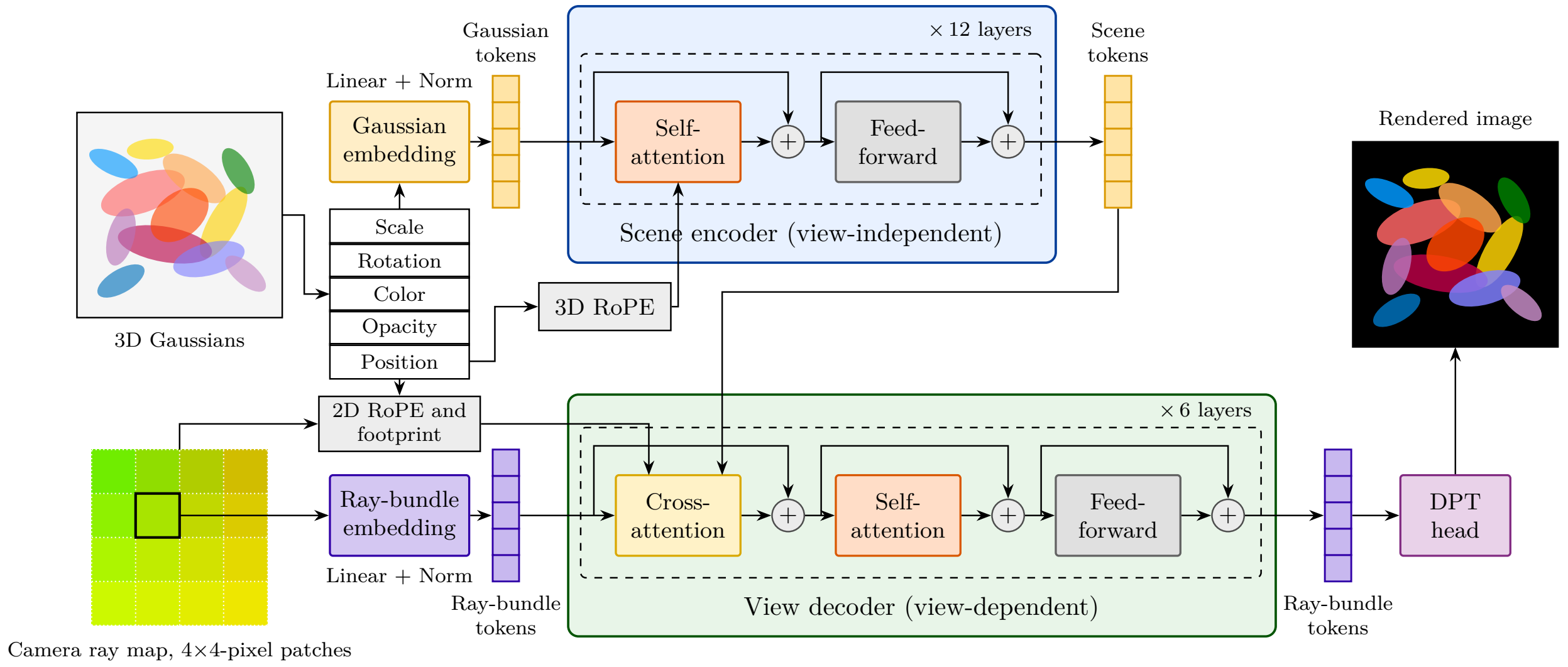
\begin{figure}[!htb]
  \centering
  \resizebox{\textwidth}{!}{\begin{tikzpicture}[
    x=1mm, y=1mm,
    font=\small,
    >={Stealth[length=2mm, width=1.5mm]},
    flow/.style={->, line width=0.6pt},
    block/.style={draw=#1!85!black, fill=#1!22, line width=0.8pt, rounded corners=1.2pt, align=center,
                  minimum width=17mm, minimum height=11mm, inner sep=1.5pt},
    plain/.style={draw, line width=0.6pt, align=center, inner sep=2pt},
    cell/.style={draw, line width=0.6pt, minimum width=19mm, minimum height=4.8mm, inner sep=0pt, outer sep=0pt,
                 font=\footnotesize},
    plus/.style={draw=black!70, fill=black!8, line width=0.6pt, circle, inner sep=0.5pt, minimum size=4.4mm},
    stage/.style={draw=#1!60!black, fill=#1!9, line width=0.9pt, rounded corners=4pt, inner sep=0pt},
    repeatbox/.style={draw, dashed, line width=0.6pt, rounded corners=2pt, inner sep=0pt},
    note/.style={font=\footnotesize, align=center},
    pics/tokens/.style n args={2}{code={
        \foreach \i in {-2,...,2}
            \node[draw=#1!85!black, fill=#1!#2, line width=0.7pt, minimum size=3.6mm, inner sep=0pt, outer sep=0pt]
                at (0, \i*3.6) {};
        \node[minimum width=3.6mm, minimum height=18mm, inner sep=0pt, outer sep=0.4pt] (-box) at (0, 0) {};
    }},
]
\colorlet{cGauss}{orange!60!yellow}   %
\colorlet{cRay}{blue!55!violet}       %
\colorlet{cSelf}{orange!80!red}       %
\colorlet{cCross}{orange!45!yellow}   %
\colorlet{cFFN}{black!55}             %
\colorlet{cEnc}{blue!60!cyan}         %
\colorlet{cDec}{green!55!black}       %
\colorlet{cHead}{violet!80}           %

\begin{scope}[shift={(0,-10)}]
  \draw[line width=0.6pt, fill=black!4] (-14,-14) rectangle (14,14);
  \foreach \x/\y/\a/\rx/\ry/\c in {
      -5/3/20/6/2.6/red!60, 2/5/-35/5/2.2/orange!70, 6/-1/60/5.5/2/yellow!70!orange, -2/-4/-10/6.5/2.4/red!45!violet,
      -8/-3/75/4/1.8/violet!55, 4/-6/15/5/2.2/blue!50, 0/0/40/4.5/3/orange!55!red, 8/6/-60/3.5/1.6/green!50!black,
      -9/7/-25/3.6/1.5/blue!40!cyan, -4/9/5/3.2/1.4/yellow!80!orange, 9/-8/-40/3.4/1.5/violet!45, -8/-9/30/3.6/1.6/cyan!50!blue}
    \fill[\c, opacity=0.62, rotate around={\a:(\x,\y)}] (\x,\y) ellipse [x radius=\rx, y radius=\ry];
\end{scope}
\node[note] at (0,-27) {3D Gaussians};

\node[cell] (aScale)    at (30,-11.6) {Scale};
\node[cell, below=-0.6pt of aScale]    (aRot)   {Rotation};
\node[cell, below=-0.6pt of aRot]      (aColor) {Color};
\node[cell, below=-0.6pt of aColor]    (aOpac)  {Opacity};
\node[cell, below=-0.6pt of aOpac]     (aPos)   {Position};
\node[block=cGauss, minimum width=19mm] (gemb) at (30,0) {Gaussian\\embedding};
\draw[flow] (aScale.north) -- (gemb.south);
\node[note, above=0.3mm of gemb] {Linear + Norm};
\draw[flow] (14,-10) -- (17,-10) |- (aColor.west);

\pic (gtok) at (44.5,0) {tokens={cGauss}{35}};
\node[note, above=0.6mm of gtok-box] {Gaussian\\tokens};
\draw[flow] (gemb.east) -- (gtok-box.west);

\node[block=cSelf] (esa)  at (68,0)  {Self-\\attention};
\node[plus]        (ep1)  at (83,0)  {$+$};
\node[block=cFFN]  (effn) at (98,0)  {Feed-\\forward};
\node[plus]        (ep2)  at (113,0) {$+$};
\pic (stok) at (128,0) {tokens={cGauss}{35}};
\node[note, above=0.6mm of stok-box] {Scene\\tokens};

\draw[flow] (gtok-box.east) -- (esa.west);
\draw[flow] (esa) -- (ep1);   \draw[flow] (ep1) -- (effn);
\draw[flow] (effn) -- (ep2);  \draw[flow] (ep2) -- (stok-box.west);
\draw[flow] (56.5,0) |- (68,9.5) -| (ep1.north);
\draw[flow] (87.5,0) |- (98,9.5) -| (ep2.north);

\begin{scope}[on background layer]
  \node[stage=cEnc, fit={(53,18.5) (119.5,-16.5)}] (enc) {};
  \node[repeatbox, fit={(54.7,12) (117.5,-8.5)}] (encrep) {};
\end{scope}
\node[font=\normalsize] at (86.2,-12.7) {Scene encoder (view-independent)};
\node[note, anchor=east] at (117.5,15.2) {$\times$\,12 layers};

\node[plain, fill=black!7, minimum width=18mm, minimum height=6.5mm] (rope3) at (58,-22.5) {3D RoPE};
\node[plain, fill=black!7, minimum width=22mm, minimum height=7.5mm, font=\footnotesize] (proj) at (30,-38.5)
     {2D RoPE and\\footprint};
\draw[flow] (aPos.east) -- ++(3.5,0) |- (rope3.west);
\draw[flow] (aPos.south) -- (proj.north);
\draw[flow] (rope3.east) -| (68,-5.5);

\begin{scope}[yshift=6mm]
\begin{scope}[shift={(-12,-72)}]
  \foreach \i in {0,...,3} \foreach \j in {0,...,3} {
      \pgfmathtruncatemacro{\g}{80 - 18*\i}
      \pgfmathtruncatemacro{\o}{25 + 15*\j}
      \fill[green!\g!orange!\o!yellow] (\i*6, \j*6) rectangle ++(6,6);
  }
  \draw[white, line width=0.5pt, densely dotted] (0,0) grid[step=6] (24,24);
  \draw[line width=1pt] (6,12) rectangle ++(6,6);
  \coordinate (patch) at (12,15);
\end{scope}
\node[note] at (0,-75.5) {Camera ray map, $4{\times}4$-pixel patches};

\node[block=cRay, minimum width=19mm] (remb) at (30,-57) {Ray-bundle\\embedding};
\draw[flow] (patch) -- (17,-57) -- (remb.west);
\node[note, below=0.3mm of remb] {Linear + Norm};
\draw[flow] (0,-48) |- (proj.west);

\pic (rtok) at (44.5,-57) {tokens={cRay}{28}};
\node[note, below=0.6mm of rtok-box] {Ray-bundle\\tokens};
\draw[flow] (remb.east) -- (rtok-box.west);

\node[block=cCross] (dxa)  at (68,-57)  {Cross-\\attention};
\node[plus]         (dp1)  at (83,-57)  {$+$};
\node[block=cSelf]  (dsa)  at (98,-57)  {Self-\\attention};
\node[plus]         (dp2)  at (113,-57) {$+$};
\node[block=cFFN]   (dffn) at (128,-57) {Feed-\\forward};
\node[plus]         (dp3)  at (143,-57) {$+$};
\pic (otok) at (158,-57) {tokens={cRay}{28}};
\node[note, below=0.6mm of otok-box] {Ray-bundle\\tokens};
\node[block=cHead, minimum width=15mm] (head) at (174,-57) {DPT\\head};

\draw[flow] (rtok-box.east) -- (dxa.west);
\draw[flow] (dxa) -- (dp1);   \draw[flow] (dp1) -- (dsa);   \draw[flow] (dsa) -- (dp2);
\draw[flow] (dp2) -- (dffn);  \draw[flow] (dffn) -- (dp3);  \draw[flow] (dp3) -- (otok-box.west);
\draw[flow] (otok-box.east) -- (head.west);
\draw[flow] (56.5,-57) |- (68,-47.5) -| (dp1.north);
\draw[flow] (87.5,-57) |- (98,-47.5) -| (dp2.north);
\draw[flow] (117.5,-57) |- (128,-47.5) -| (dp3.north);

\begin{scope}[on background layer]
  \node[stage=cDec, fit={(53,-40.5) (149.5,-73.5)}] (dec) {};
  \node[repeatbox, fit={(54.7,-45) (147.5,-65.5)}] (decrep) {};
\end{scope}
\node[font=\normalsize] at (101.2,-69.7) {View decoder (view-dependent)};
\node[note, anchor=east] at (147.5,-42.8) {$\times$\,6 layers};

\draw[flow] (stok-box.south) |- (74,-26.5) -- (74,-51.5);
\draw[flow] (proj.east) -| (64,-51.5);

\end{scope}

\begin{scope}[shift={(174,-14)}]
  \draw[line width=0.6pt, fill=black] (-14,-14) rectangle (14,14);
  \foreach \x/\y/\a/\rx/\ry/\c in {
      -5/3/20/6/2.6/red!60, 2/5/-35/5/2.2/orange!70, 6/-1/60/5.5/2/yellow!70!orange, -2/-4/-10/6.5/2.4/red!45!violet,
      -8/-3/75/4/1.8/violet!55, 4/-6/15/5/2.2/blue!50, 0/0/40/4.5/3/orange!55!red, 8/6/-60/3.5/1.6/green!50!black,
      -9/7/-25/3.6/1.5/blue!40!cyan, -4/9/5/3.2/1.4/yellow!80!orange, 9/-8/-40/3.4/1.5/violet!45, -8/-9/30/3.6/1.6/cyan!50!blue}
    \fill[\c, opacity=0.85, rotate around={\a:(\x,\y)}] (\x,\y) ellipse [x radius=\rx, y radius=\ry];
\end{scope}
\node[note] at (174,3) {Rendered image};
\draw[flow] (head.north) -- (174,-28);  %
\end{tikzpicture}}
  \caption{Overview of \method{}. Each Gaussian's 14 parameters are embedded as one token, and a view-independent
  scene encoder (top) lets these tokens exchange information, using a 3D rotary embedding of their centers. For each
  view, the image is split into $4 \times 4$-pixel patches, and each patch becomes a token built from its camera
  rays. The view decoder (bottom) lets these tokens attend to the scene tokens, using a 2D rotary embedding of where
  each Gaussian lands in the image and an embedding of its projected footprint, and a DPT head turns them into
  pixels.}
  \label{fig:architecture}
  \vspace{-8pt}  %
\end{figure}

\method{} keeps RenderFormer's two-stage layout (Section~\ref{sec:background_renderformer}) and starts from its
pretrained weights. We change the first stage so that each token is a Gaussian instead of a
triangle, and how the second stage looks at the scene, so that it can handle tens of thousands of
semi-transparent Gaussians: it is told where each Gaussian lands in the image and how large it appears there.
Figure~\ref{fig:architecture} shows the full model.

\subsection{Gaussian tokens and the scene encoder}
\label{sec:method_encoder}

Each Gaussian is described by a vector of 14 numbers,
\begin{equation}
    \mathbf{g}_i = \bigl[\, \boldsymbol{\mu}_i, \; \mathbf{s}_i, \; \mathbf{q}_i, \; \mathbf{c}_i, \; o_i \,\bigr]
    \in \mathbb{R}^{14},
    \label{eq:gaussian_vector}
\end{equation}
holding its center (3), scales (3), rotation quaternion (4), color (3) and opacity (1). The token of Gaussian $i$ is
\begin{equation}
    \mathbf{t}_i = \mathbf{e} + \mathrm{RMSNorm}(\mathbf{A} \mathbf{g}_i + \mathbf{b}),
    \label{eq:gaussian_token}
\end{equation}
where $\mathbf{A} \in \mathbb{R}^{768 \times 14}$ and $\mathbf{b}$ are a linear layer followed by RMSNorm~\citep{zhang2019rmsnorm}, and $\mathbf{e}$ is a learned
embedding shared by all Gaussian tokens. The token embeds all 14 numbers, the center included, so it changes with the
random rotation of every training sample (Section~\ref{sec:method_training}). This input layer replaces RenderFormer's
triangle embedding, and it is the only new part of the model apart from the footprint layers
(Section~\ref{sec:method_footprint}), which start at zero. In addition, the encoder's
self-attention applies a 3D rotary position embedding of each Gaussian's center, where RenderFormer used the
triangle's vertices. The 16 register tokens sit at the mean
center of the Gaussians. The rest of the scene encoder, 12 self-attention layers, is unchanged and runs once
per object, whatever the number of views.

\subsection{The view decoder}
\label{sec:method_decoder}

The view decoder follows RenderFormer: each image patch is a token built from the directions of the rays through its
pixels, and six layers of cross-attention to the scene tokens and self-attention among the patches turn these tokens
into pixels through a DPT head. Inside it, the Gaussians' centers are expressed in the camera's frame and enter the
attention through the same 3D rotary embedding. We make three changes.

\noindent\textbf{2D rotary embedding of projected positions.} A 3D position embedding tells the decoder where a Gaussian is relative to the camera, but not where it lands in the
image, which is what a patch needs to know. We therefore project each Gaussian's center into the view,
\begin{equation}
    u_i = \frac{1}{p} \Bigl(\frac{W}{2} + f \frac{x_i}{z_i}\Bigr), \qquad
    v_i = \frac{1}{p} \Bigl(\frac{H}{2} - f \frac{y_i}{z_i}\Bigr),
    \label{eq:projection}
\end{equation}
where $(x_i, y_i, z_i)$ is the center in camera coordinates, $f$ the focal length in pixels, $W \times H$ the image
size and $p$ the patch size, so $(u_i, v_i)$ is measured in patches. In the cross-attention, the keys of each
Gaussian are rotated by a 2D rotary embedding of $(u_i, v_i)$ and the query of each patch by the same embedding of
the patch's center, so the attention between a patch and a Gaussian depends on the distance between them in the
image. The patches' self-attention uses the same 2D embedding. It rotates 8 channel pairs per image axis in each
attention head, pairs that the 3D embedding leaves unrotated, at 8 frequencies from 0.25 to 1.75 radians per patch,
so the pretrained 3D embedding is kept as it is.

\noindent\textbf{Smaller patches.} RenderFormer uses patches of $8 \times 8$ pixels. We use $4 \times 4$ pixels, which gives the decoder four times as
many tokens per image and lets it resolve finer detail. To reuse the pretrained ray embedding, which expects the 64
ray directions of an $8 \times 8$ patch, we upsample each $4 \times 4$ patch's ray directions to $8 \times 8$ by
nearest-neighbor interpolation before embedding them. This is the same as an embedding of the 48 numbers of a
$4 \times 4$ patch whose weights are the pretrained ones summed over each $2 \times 2$ block. The DPT head upsamples
its output to the image size whatever the patch size.

\noindent\textbf{Projected footprint on the keys.} \label{sec:method_footprint} The rotary embeddings tell the decoder where each Gaussian's center lands, but not how far the Gaussian reaches in
the image. Its scale and rotation reach the decoder only through the scene tokens, which are the same for every view,
so the decoder would have to work out each Gaussian's apparent size and shape for every camera. We give it directly.
With $\boldsymbol{\Sigma}_i = \mathbf{R}_i \operatorname{diag}(\mathbf{s}_i)^2 \mathbf{R}_i^\top$ the Gaussian's
covariance in camera coordinates and $\mathbf{J}_i$ the Jacobian of the projection of Eq.~\ref{eq:projection} at its
center, its footprint in the image, in patches, is the $2 \times 2$ covariance
\begin{equation}
    \boldsymbol{\Sigma}'_i = \mathbf{J}_i \boldsymbol{\Sigma}_i \mathbf{J}_i^\top + \frac{0.3}{p^2} \mathbf{I},
    \qquad
    \mathbf{J}_i = \frac{f}{p\, z_i} \begin{pmatrix} 1 & 0 & -x_i / z_i \\ 0 & -1 & y_i / z_i \end{pmatrix},
    \label{eq:footprint}
\end{equation}
the same projection that 3DGS applies before rasterizing~\citep{kerbl20233dgs}, where the small isotropic term
(0.3 square pixels) is the low-pass filter of the gsplat rasterizer~\citep{ye2025gsplat}. Without it the model is only 0.08 dB
worse (Section~\ref{sec:exp_ablations}). We describe it by four numbers,
the logarithms of its two standard deviations and $(\cos 2\theta, \sin 2\theta)$ for the angle $\theta$ of its major
axis, each clipped to $[-6, 6]$. In each decoder layer, a linear layer maps these four numbers to a vector that is
added to the Gaussian's key in the cross-attention, so the footprint changes which Gaussians a patch attends to but
not what it reads from them. Adding it to the values as well helps on our ablation harness, but we keep it in the
keys only for the reasons given in Section~\ref{sec:exp_ablations}. Register tokens and Gaussians behind the camera get zeros. These linear layers are new
and initialized to zero, so the model starts from exactly the function it had without them.

The projected center and footprint are computed in closed form for each Gaussian from its parameters and the camera,
a change of coordinates like expressing each triangle's vertices in the camera's frame for RenderFormer's decoder. Nothing is
computed per pixel: which Gaussians cover a pixel, which of them are visible, how they blend and what color results
are left to the network.

The first two changes add no parameters and the third only 23,040, initialized to zero, but the first two change what
the pretrained layers compute, which the short stage 3 at $256 \times 256$ pixels recovers
(Section~\ref{sec:method_training}).

\subsection{Training}
\label{sec:method_training}

The model is trained to reproduce renders of the full, unpruned splat from a version pruned to 20,000 Gaussians
(Section~\ref{sec:data}). The model outputs colors in the log encoding $\log_{10}(1 + \mathbf{I})$ inherited from
RenderFormer, and the loss combines an L1 term in that encoding with LPIPS~\citep{zhang2018lpips} on a VGG backbone:
\begin{equation}
    \mathcal{L} = (1 - \lambda) \, \mathcal{L}_1^{w}\bigl(\hat{\mathbf{I}}, \log_{10}(1 + \mathbf{I})\bigr)
    + \lambda \, \mathcal{L}_{\mathrm{LPIPS}}\bigl(10^{\hat{\mathbf{I}}} - 1, \mathbf{I}\bigr), \qquad \lambda = 0.5,
    \label{eq:loss}
\end{equation}
where $\hat{\mathbf{I}}$ is the model's output and $\mathbf{I}$ the target image. $\mathcal{L}_1^{w}$ is the L1
distance averaged over pixels with a weight of 1 on the object and $0.05$ on the background, since an object covers
only a few percent of the image. Every training sample rotates the object and its camera by the same random rotation, which leaves the image
unchanged but keeps the model from relying on a fixed orientation.

Training runs in four stages (Table~\ref{tab:stages}). The final model's first two stages differ slightly from this
recipe (Appendix~\ref{sec:appendix_reproducibility}). The first trains only the new input layer, with the rest of
RenderFormer frozen. The second trains the whole model at
$256 \times 256$ pixels. The third switches on the three changes above for a short recovery at the same resolution,
and the fourth trains at $512 \times 512$ pixels with LPIPS. We use AdamW~\citep{loshchilov2019adamw} with a weight decay of 0.01, a batch of 8
views on 8 GPUs, and a learning rate of $5 \times 10^{-5}$ ($10^{-3}$ in the first stage), held constant after a short
warmup and decayed with a cosine over the last steps of each stage.

\begin{table}[t]
  \centering
  \caption{Training stages. Stage 3 switches on the changes of Section~\ref{sec:method_decoder}.}
  \label{tab:stages}
  \begin{tabular}{@{}llrll@{}}
    \toprule
    Stage & Trains & Steps & Resolution & Loss \\
    \midrule
    1 & input layer only & 33.5k & 256 & log-L1 \\
    2 & whole model & 402k & 256 & log-L1 \\
    3 & whole model, new decoder changes & 3k & 256 & log-L1, background weight \\
    4 & whole model & 385k & 512 & log-L1 + LPIPS, background weight \\
    \bottomrule
  \end{tabular}
  \vspace{-8pt}  %
\end{table}

\section{Data}
\label{sec:data}

\noindent\textbf{Source.} We train on Objaverse\_Splats\footnote{\url{https://huggingface.co/datasets/ShapeSplats/Objaverse_Splats}}, the
subset of ShapeSplat with a trained 3DGS splat per Objaverse object. We use only splats that fit their object well according to
ShapeSplat's metadata, with a PSNR of at least 32 dB and an LPIPS of at most 0.06, and that have exactly 50,000
Gaussians. From these we draw 30,000 objects for training and 2,000 for validation, and drop objects that are
degenerate or nearly transparent (maximum opacity below 0.4). This leaves 26,820 training objects and 1,806
validation objects.

\noindent\textbf{Normalization.} Each splat is centered on the median position of its Gaussians, rotated so that its up axis is $+y$, and scaled
uniformly so that its largest coordinate is 0.45, which places every object in $[-0.45, 0.45]^3$. We keep only the
view-independent color of each Gaussian (Section~\ref{sec:background_3dgs}).

\noindent\textbf{Ground truth.} For every object we render 28 views of the full, unpruned splat with gsplat at $512 \times 512$
pixels, with a $45^\circ$ field of view and a black background. The cameras orbit the object and look at its center: 14 at
a distance of 1.7, and 7 each at 1.15 and 2.45. Consecutive cameras alternate between looking down at the object and
slightly up at it, to see its top and its bottom. Every object has the same 28 cameras.

\noindent\textbf{Input preparation.} The full splats have 50,000 Gaussians, while the model takes 20,000. We keep the 20,000 Gaussians with the highest
importance score adapted from LightGaussian~\citep{fan2024lightgaussian}, which sums each Gaussian's opacity times its
projected area over 64 orbit views and slightly favors larger Gaussians. Pruning alone leaves holes and darkens the object, so we then fine-tune the kept
Gaussians for 1,500 steps to match renders of the full splat from those 64 views, with the usual 3DGS loss (L1 plus
0.2 times D-SSIM~\citep{wang2004ssim}) and without adding or removing any Gaussians. This takes about 10 seconds per object on one GPU.
The 64 views lie at the same distance as the main evaluation views, but at different angles. The model never sees
the full splat, only the pruned one, and is trained to reproduce the renders of the full splat. A rasterizer can
render the pruned input directly. We call this gsplat on the input, and use it as the bar for the model throughout
Section~\ref{sec:experiments}, while the reference is always the render of the full splat.

\noindent\textbf{Splits.} Of the 1,806 validation objects, 100 monitor the validation
loss during training, and 300 others, never seen in training, are the held-out objects on which we evaluated every
design decision. We call them the development held-out set. For the main result, we draw 300 further objects at
random from the remaining 1,406, never used during development: the fresh held-out set. We also evaluate
300 training objects.

\providecommand{\todo}[1]{\textcolor{red}{[TODO: #1]}}

\section{Experiments}
\label{sec:experiments}

\noindent\textbf{Setup.} All images are $512 \times 512$ pixels with a $45^\circ$ field of view. Both the model's input and the ground truth use
only the view-independent color of each Gaussian. We report PSNR, SSIM and LPIPS~\citep{zhang2018lpips}, with LPIPS computed on an
AlexNet backbone (the training loss uses VGG). The main result is on the fresh held-out objects, over all 28 views of
the training setup. Unless stated otherwise, every other number is the mean over the 300 development held-out
objects, 4 views each at distance 1.7. Since the object covers only a small part of the image, every metric is
computed on the evaluation crop: the square around the object in the reference (pixels brighter than 0.02), with a
12-pixel margin, resized to $512 \times 512$. The black background inside the crop is easy for both renderers, so
we also report masked PSNR, over the pixels where the full splat's opacity exceeds 0.5, on the whole image. We
compare against two references. The first is a rasterization of the full, unpruned splat, which is the image the
model is trained to produce. The second is a rasterization of the pruned 20,000-Gaussian input, which is exactly
what the input encodes.

\subsection{Main result}
\label{sec:exp_main}

Table~\ref{tab:main} shows the main result, on 300 held-out objects that we never used during development. Over
all 28 views at the three training distances, \method{} reaches 38.08 dB against the unpruned render, while
rasterizing the same pruned input reaches 44.02 dB. The gap depends on the camera distance, from 3.9 dB at 1.15 to
6.8 dB at 1.7. Rasterizing the input is best at 1.7, the distance of the 64 views the pruned input was fine-tuned on
(Section~\ref{sec:data}), while the model is nearly flat across distances. Masked PSNR, which leaves out the black
background, is about 5 dB lower for both, with about the same gap. Figure~\ref{fig:qualitative} shows three
held-out objects, at the 10th, 50th and 90th percentile of the model's score. The model's errors concentrate on edges
and fine texture, while its colors and shapes are right.

The gap is not one of generalization. On the development held-out objects, on which we chose our design, the model
scores the same, 38.02 dB, so choosing on them did not inflate the result. On 300 training objects, over their 28
training views, it scores only 0.17 dB higher than on unseen objects. These are the exact images the model was
trained to produce, from the exact same input, and it still stays 5.9 dB below rasterizing that input. Nor does the
gap come from what pruning removed: measured against a rasterization of its own input instead of the unpruned
render, the model scores only 0.3 dB higher (4 views at distance 1.7). Its error is in rendering the input it has.
Each training image was seen only about four times in the final stage, though, so an under-trained model would show
the same. Sections~\ref{sec:exp_scaling} and~\ref{sec:exp_gap} ask where the gap comes from.

\newcommand{\qualimg}[1]{\includegraphics[width=0.152\textwidth]{figures/qualitative/#1}}
\begin{figure}[t]
  \centering
  \setlength{\tabcolsep}{1pt}
  \footnotesize
  \begin{tabular}{@{}ccccc@{}}
    Reference & gsplat on the input & Ours & Error, gsplat & Error, Ours \\
    \qualimg{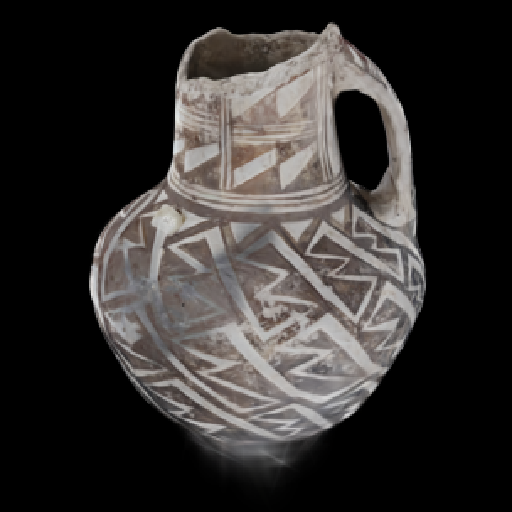} & \qualimg{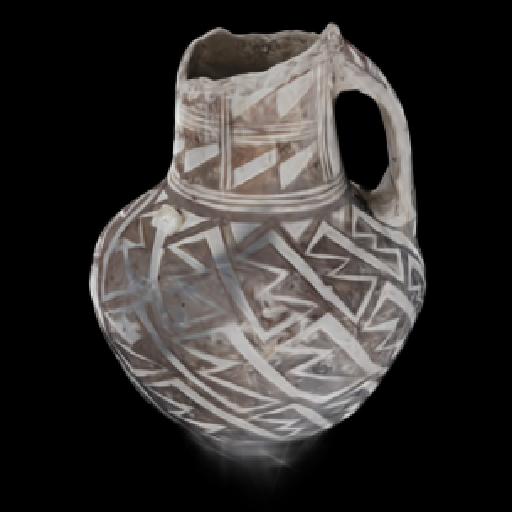} & \qualimg{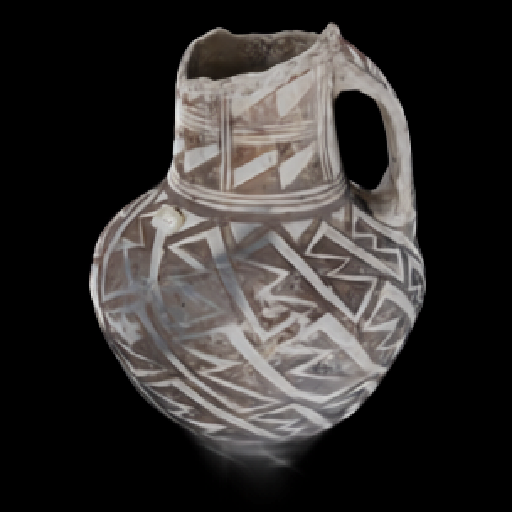} & \qualimg{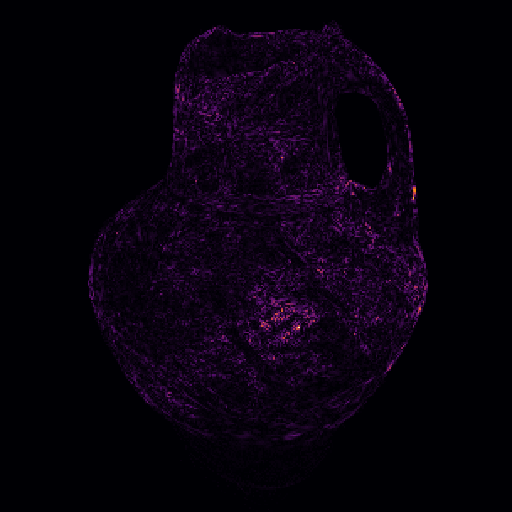} & \qualimg{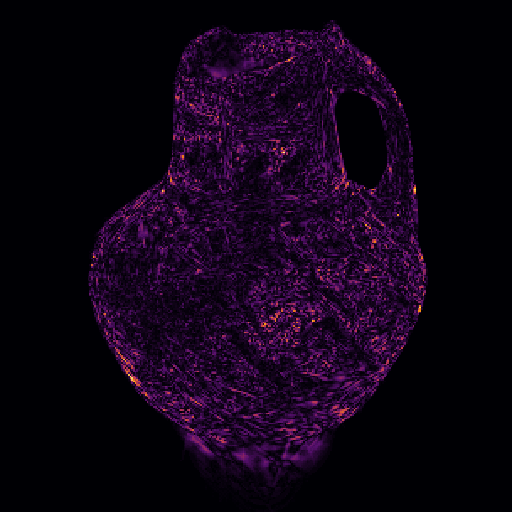} \\
    \qualimg{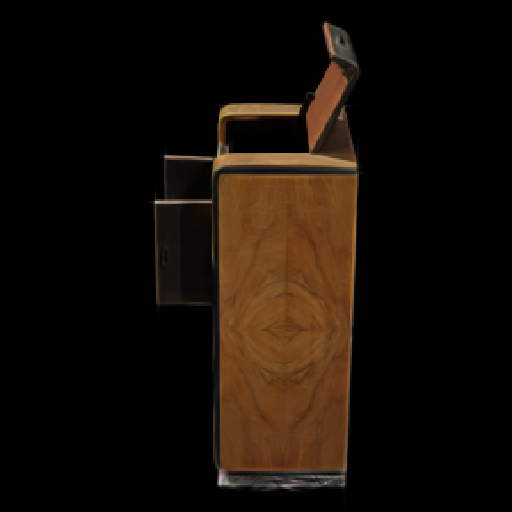} & \qualimg{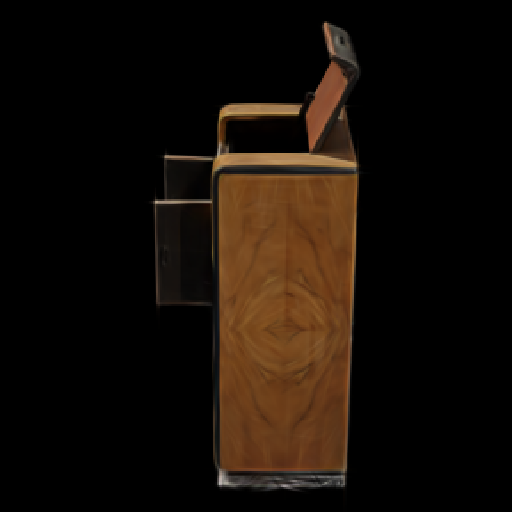} & \qualimg{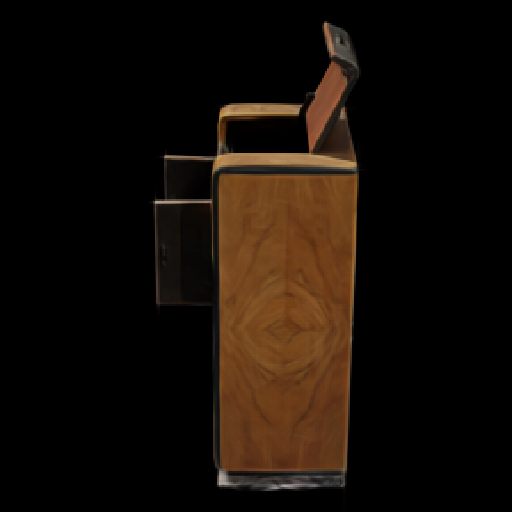} & \qualimg{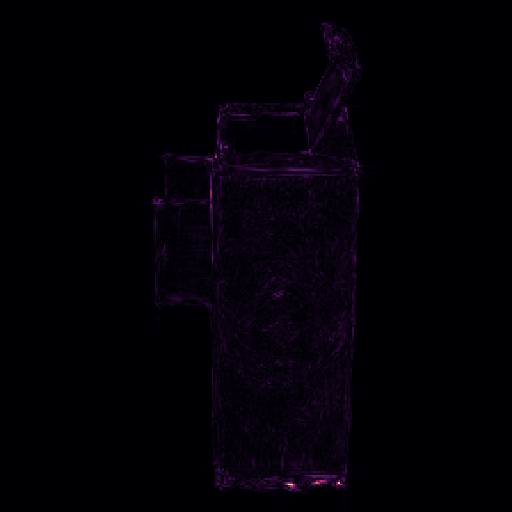} & \qualimg{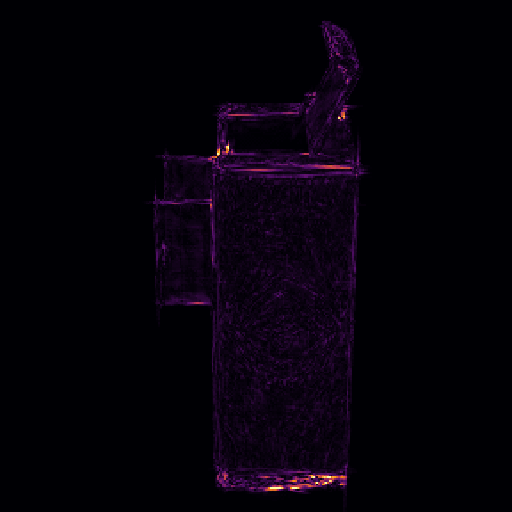} \\
    \qualimg{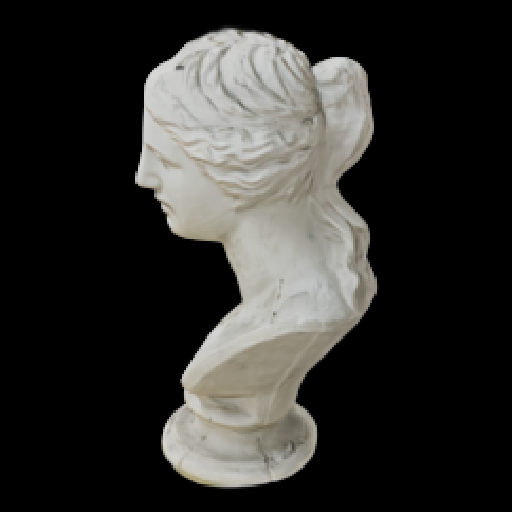} & \qualimg{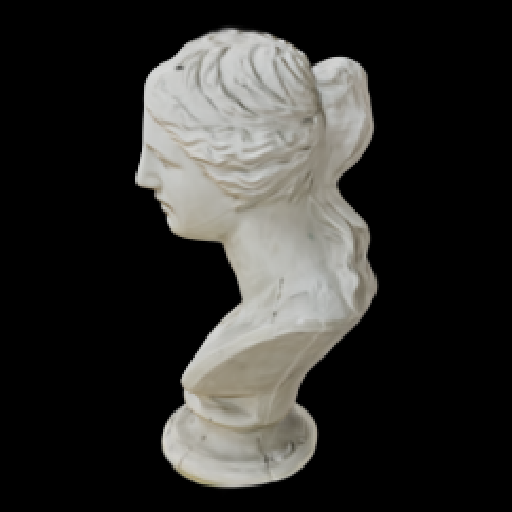} & \qualimg{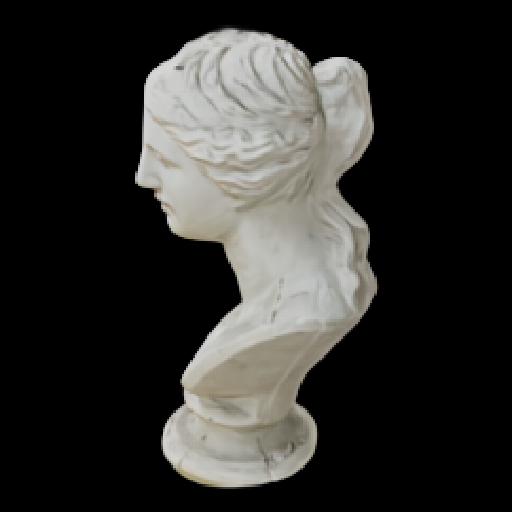} & \qualimg{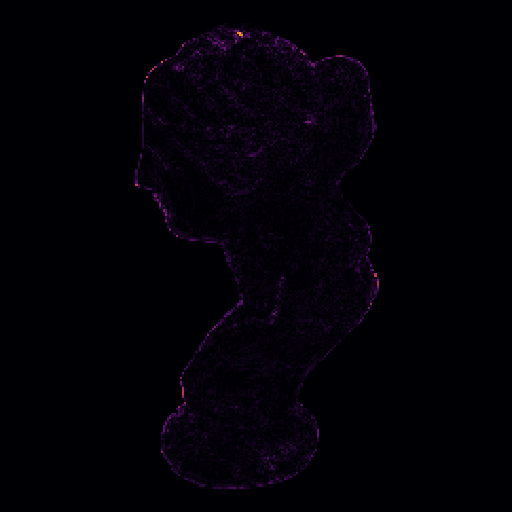} & \qualimg{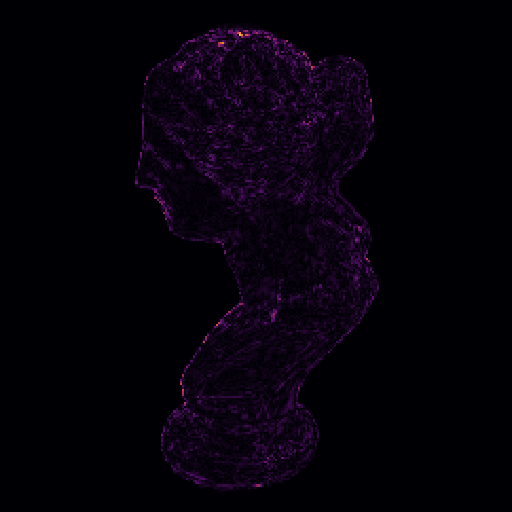} \\
  \end{tabular}
  \caption{Three fresh held-out objects, chosen at the 10th, 50th and 90th percentile of \method{}'s per-object PSNR.
  From left to right: the reference (the full, unpruned splat rasterized with gsplat), gsplat on the pruned
  20,000-Gaussian input, \method{} on the same input, and the absolute error of each against the reference, scaled
  by 5. \method{} scores 34.5, 39.5 and 41.7 dB on these views, and gsplat on the input 39.7, 45.5 and 45.7 dB.
  \method{} gets shapes and colors right, and its error concentrates on edges and fine texture.}
  \label{fig:qualitative}
  \vspace{-8pt}  %
\end{figure}

\begin{table}[t]
  \centering
  \caption{Main result, against the unpruned render. PSNR over the 4 views at distance 1.7 used in the rest of the
  paper and over all 28 views at the three training distances, masked PSNR over the object's pixels on the whole
  image, and SSIM and LPIPS over the 28 views. Fresh held-out: 300 objects never used during development. Development
  held-out: the 300 held-out objects of the other sections. No full splat is kept for the training objects, so they
  have no masked PSNR. Without the footprint: the same recipe and schedule without the projected footprint
  (Section~\ref{sec:method_footprint}). Higher PSNR and SSIM and lower LPIPS are better.}
  \label{tab:main}
  \small
  \begin{tabular}{@{}lccccc@{}}
    \toprule
    & \multicolumn{2}{c}{PSNR (dB)} & Masked & & \\
    Renderer and objects & 4 views & 28 views & PSNR (dB) & SSIM & LPIPS \\
    \midrule
    gsplat on the input, fresh held-out & 44.97 & 44.02 & 39.19 & 0.9954 & 0.0046 \\
    gsplat on the input, training objects & 45.01 & 44.12 & -- & 0.9954 & 0.0046 \\
    \midrule
    Ours, fresh held-out & 38.36 & 38.08 & 32.68 & 0.9814 & 0.0113 \\
    Ours, development held-out & 38.36 & 38.02 & 32.66 & 0.9802 & 0.0116 \\
    Ours, training objects & 38.55 & 38.25 & -- & 0.9821 & 0.0108 \\
    \midrule
    Ours without the footprint, fresh held-out & 36.87 & 36.67 & 31.34 & 0.9779 & 0.0140 \\
    \bottomrule
  \end{tabular}
  \vspace{-8pt}  %
\end{table}

\noindent\textbf{Cost.} \method{} is much slower than a rasterizer (Table~\ref{tab:cost}). On one L40S GPU, a single view takes 275 ms,
more than half of it in the scene encoder. The encoder runs once per object, so every further view in the same call
adds about 135 ms. gsplat renders the same input in under a millisecond.

\begin{table}[t]
  \centering
  \caption{Rendering time and memory on one NVIDIA L40S GPU, 20,000 Gaussians, $512 \times 512$ pixels. Median over
  10 held-out objects. Of the 275 ms of one view, 151 ms are the scene encoder and 124 ms the view decoder. The model has
  195M parameters, 113M in the scene encoder and 82M in the view decoder.}
  \label{tab:cost}
  \begin{tabular}{@{}rrrrr@{}}
    \toprule
    Views per call & Ours (ms) & per view (ms) & peak memory (GB) & gsplat (ms) \\
    \midrule
    1 & 275 & 275 & 2.8 & 0.7 \\
    4 & 685 & 171 & 7.7 & 0.9 \\
    14 & 2,012 & 144 & 13.0 & 1.2 \\
    \bottomrule
  \end{tabular}
  \vspace{-8pt}  %
\end{table}

The gap in cost grows with the number of Gaussians (Appendix~\ref{sec:appendix_cost}). The scene encoder attends from
every Gaussian to every other, so the model's time grows roughly with the square of the input size: one view of
640,000 Gaussians takes 75 s instead of 275 ms at 20,000. Its memory grows roughly linearly, to 23.7 GB at 640,000
Gaussians, while gsplat stays under 3 ms and 0.1 GB of working memory.

\subsection{Ablations}
\label{sec:exp_ablations}

Training the final model took 1,327 GPU-hours, so we do not repeat it for every change. Our earlier probes trained
on 10 objects, and their results did not carry over to larger training sets. We therefore test every change on one
harness of moderate scale. It trains on 1,000 objects from the end of stage 2 (Table~\ref{tab:stages}), with
short versions of stages 3 and 4 (Appendix~\ref{sec:appendix_supporting}). Every arm is evaluated on the 300 held-out objects and on its own training objects, and differs from the
harness with all changes, the recipe of the final model, in one change only (Table~\ref{tab:ablations}). The end of
stage 2 has $8 \times 8$ patches, no 2D embedding and no footprint, so the arm with all changes has to learn all three
in these steps, while the ablated arms have less to learn. This favors the ablated arms. Every arm is a single run.
To see how far two runs of the same recipe differ, we repeat the arm with all changes with another seed for the data
order, the drawn views and the random rotations. The two runs differ by 0.02 dB on the held-out objects
(Table~\ref{tab:ablations}), so differences of a few tenths of a dB between arms are larger than the variation
between runs.

\begin{table}[t]
  \centering
  \caption{Ablations on the 1,000-object harness. PSNR in dB against the unpruned render, on the 300 held-out objects
  and on 300 of the training objects (their training views). Difference: held-out PSNR minus that of the arm with all
  changes, with a 95\% bootstrap interval over objects for this single run. The last rows vary how the footprint is
  given and repeat the arm with all changes with another seed, which differs from the first run by 0.10 dB per object on
  average.}
  \label{tab:ablations}
  \small
  \begin{tabular}{@{}lcccc@{}}
    \toprule
    & \multicolumn{3}{c}{Held-out objects} & Training objects \\
    Model & PSNR & Difference & LPIPS & PSNR \\
    \midrule
    All changes & 34.42 & & 0.0224 & 35.69 \\
    No footprint & 32.71 & $-1.71 \pm 0.07$ & 0.0316 & 34.43 \\
    No 2D rotary embedding & 29.22 & $-5.20 \pm 0.18$ & 0.0467 & 30.50 \\
    $8 \times 8$ patches & 31.37 & $-3.05 \pm 0.10$ & 0.0379 & 32.49 \\
    Windowed cross-attention & 34.62 & $+0.20 \pm 0.02$ & 0.0218 & 35.85 \\
    \midrule
    Footprint without the 0.3-pixel$^2$ term & 34.34 & $-0.08 \pm 0.02$ & 0.0225 & 35.63 \\
    Footprint on keys and values & 35.10 & $+0.68 \pm 0.03$ & 0.0197 & 36.29 \\
    All changes, second seed & 34.44 & $+0.02 \pm 0.02$ & 0.0223 & 35.68 \\
    \bottomrule
  \end{tabular}
  \vspace{-8pt}  %
\end{table}

All three changes to how the decoder reads the scene matter. Without the footprint, the model loses 1.7 dB on the
held-out objects, and 299 of the 300 objects are worse. Without the 2D embedding it loses 5.2 dB, and with
$8 \times 8$ patches instead of $4 \times 4$ it loses 3.0 dB. Both cost more than before the footprint was added,
4.1 and 2.3 dB on the same harness, so with the footprint the decoder relies more on knowing where each Gaussian
lands and on small patches, not less. Each change costs about as
much on the training objects as on the held-out ones, so they help the model fit its training images, not only
generalize. The changes matter just as much at full scale (Table~\ref{tab:fullscale}). Trained on all objects with
the recipe before the footprint, to the first milestone of the final run, the model loses 4.3 dB without the 2D
embedding and 2.4 dB with $8 \times 8$ patches. The final recipe without the footprint scores 1.4 dB lower than the
final model on the fresh held-out objects (Table~\ref{tab:main}), and lower on every one of the 300.
Figure~\ref{fig:ablation_renders} in the appendix shows how each change looks.

Giving the decoder the footprint in its values as well as its keys, through a second zero-initialized layer per
decoder layer, adds another 0.7 dB on the harness (Table~\ref{tab:ablations}), so the footprint can also help the decoder read the Gaussians it selects. We nevertheless keep it in the keys only. There,
it only decides which Gaussians a patch attends to, which keeps a clear split between the geometry we compute and
what the network learns. The values variant is also more sensitive to inputs denser than in training
(Sections~\ref{sec:exp_gap} and~\ref{sec:exp_beyond}): with 40,000 Gaussians, it loses 0.3 dB more than the
keys-only model (95\% interval 0.2 to 0.4 dB), and for the keys-only model this loss grows with training
(Section~\ref{sec:exp_gap}). We leave the values variant to future work.

Without the small isotropic term
that we take from gsplat (0.3 square pixels), the model loses only 0.08 dB (Table~\ref{tab:ablations}), against
1.7 dB without the footprint, so its gain does not come from a rasterizer convention.

Restricting each patch's cross-attention to the Gaussians that reach its tile, the way a rasterizer bins Gaussians by
tile, adds only 0.2 dB and saves little time (Appendix~\ref{sec:appendix_negative}), so we keep dense cross-attention
and let the network decide which Gaussians each patch attends to.

The first two stages change the model the most, since they turn a mesh renderer into a Gaussian one, so we test
them separately, at 256 pixels on all 26,820 objects. Each arm trains for 50,000 steps, so all of them get the same
number of steps: from RenderFormer's weights with the input layer trained alone for the first 5,000 steps (our
recipe), from RenderFormer's weights with the whole model trained from the start, and from random weights. We evaluate each at 256 pixels on the held-out objects, against the unpruned render
downsampled as in training (Table~\ref{tab:stage_ablation} in Appendix~\ref{sec:appendix_stages}). Training the input layer alone first gives 3.1 dB more
than training the whole model from the start, in the same number of steps. The gain is not the input layer's higher
learning rate: training the whole model from the start with the input layer at the first stage's learning rate
($10^{-3}$) does worse still, 0.55 dB below the whole model at one learning rate. Its validation loss stays flat for
the first 24,000 steps before it starts to learn. We did not find a
recipe that trains the model from random weights: with the loss of stages 1 and 2, it collapsed to a constant black
image.

\subsection{Data and training length}
\label{sec:exp_scaling}

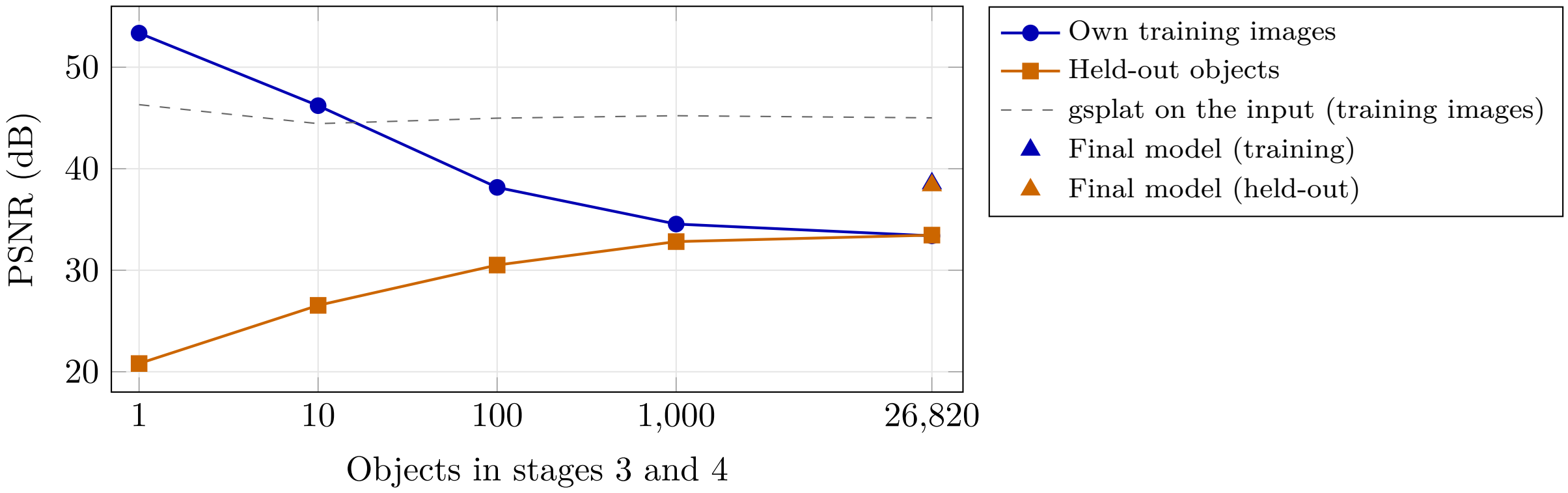
\begin{figure}[t]
  \centering
  \begin{tikzpicture}
    \begin{axis}[width=0.62\textwidth, height=5.5cm, xmode=log, log basis x=10, xmin=0.7, xmax=40000,
                 ymin=18, ymax=56, xlabel={Objects in stages 3 and 4}, ylabel={PSNR (dB)},
                 xtick={1,10,100,1000,26820}, xticklabels={1,10,100,1{,}000,26{,}820},
                 legend style={font=\footnotesize, at={(1.03,1)}, anchor=north west}, legend cell align=left,
                 grid=major, grid style={black!10}]
      \addplot[thick, mark=*, blue!70!black] table[x=n, y=train] {figures/data/nsweep.dat};
      \addplot[thick, mark=square*, orange!80!black] table[x=n, y=heldout] {figures/data/nsweep.dat};
      \addplot[dashed, mark=none, black!60] table[x=n, y=raster] {figures/data/nsweep.dat};
      \addplot[only marks, mark=triangle*, mark size=3pt, blue!70!black] coordinates {(26820, 38.55)};
      \addplot[only marks, mark=triangle*, mark size=3pt, orange!80!black] coordinates {(26820, 38.36)};
      \legend{Own training images, Held-out objects, gsplat on the input (training images), Final model (training), Final model (held-out)}
    \end{axis}
  \end{tikzpicture}
  \caption{PSNR against the unpruned render when the end of stage 2, which has seen all 26,820 objects, is fine-tuned
  on fewer objects, with the same number of steps for every point. With one object the model fits its training images
  and forgets the others. The two curves meet at the full set, far below rasterizing the same input. The triangles
  show the final model, trained about 10 times longer with the footprint.}
  \label{fig:nscaling}
  \vspace{-8pt}  %
\end{figure}

To see how much the amount of data matters, we fine-tune the end of stage 2 with the recipe of the harness
(Section~\ref{sec:exp_ablations}), before the footprint was added, on 1, 10, 100 and 1,000 objects and on all 26,820, with the same number of steps
(Figure~\ref{fig:nscaling}). Stage 2 has already seen all 26,820 objects, so every point is a fine-tune of a model that
renders thousands of objects, and the passes over each training image differ by four orders of magnitude: about
9,400 at one object and 0.35 at 26,820. On a single object, the model fits its own training images 7 dB above
rasterizing its input and forgets the other objects. With more objects the two curves approach each other and meet at
the full set, about 12 dB below rasterizing the input. At our compute, the model is not data-limited. The final model,
trained about 10 times longer on all objects and with the footprint, still scores about the same on its training and
held-out objects (the triangles).

The final run was still improving when it ended. It gained about 0.9 dB with each doubling of its steps, from
94,000 to 375,000, with no slowdown (Table~\ref{tab:milestones}), so longer training would narrow the gap,
by an amount these points cannot tell.

Appendix~\ref{sec:appendix_per_object} shows how much detail the network can output when it is trained on a single
object.

\subsection{Where the gap comes from}
\label{sec:exp_gap}

Denser inputs than in training do not help the model. With 40,000 Gaussians instead of 20,000,
rasterization gains 4.6 dB while the model loses 3.1 dB (Table~\ref{tab:generalization}), and on the real captures,
more Gaussians likewise help the rasterizer and hurt the model (Table~\ref{tab:real_captures}). It never saw inputs of that size in training. The same recipe
without the footprint also loses 1.6 dB at 40,000 Gaussians, so dense cross-attention, which spreads each patch's
attention over twice as many Gaussians, is part of the cause, and the footprint, whose values shift as denser splats
use smaller Gaussians, adds to it. An earlier model with windowed cross-attention, in which each patch sees only the
Gaussians that reach it, lost nothing at 40,000 Gaussians, but gained nothing either. With less input, the model
degrades more slowly than rasterization (Table~\ref{tab:generalization}), most likely because it was not reproducing
that detail in the first place.

The errors are largest in fine detail. The 10\% of pixels with the strongest edges in the reference hold 71\% of the
model's squared error. Yet the model trails the rasterizer by more off the edges than on them, 8.1 against 6.2 dB, so
part of its error lies at low frequencies, at the scale of the object's parts (Appendix~\ref{sec:appendix_edges}).
At high frequencies it both blurs and adds detail that is not in the reference: above 1/4 cycles per pixel it
reproduces 85\% of the reference's content, against 93\% for the rasterizer (Appendix~\ref{sec:appendix_frequency}).
The gap also grows with the number of Gaussians the model has to blend at a pixel, and is largest where 32 or more
reach it, as on most of the object (Appendix~\ref{sec:appendix_overlap}).

A reachable target does not close the gap, at least late in training. The model is trained to produce renders of the full splat, which hold
detail its pruned input does not. We therefore trained the harness to reproduce the rasterization of its own input
instead, a target that holds nothing the input lacks. Measured against that target, it scores only 0.19 dB more than
the harness trained on the full splat (Appendix~\ref{sec:appendix_supporting}). Both are short fine-tunes of a model
that was trained on the full splat, though, so this shows what a reachable target changes late in training, not from the start.

Together, these results are consistent with a bottleneck in the view decoder: the model barely uses more detail when
its input has it, its fine detail is blurred and partly spurious, and a reachable target does not help. The ablations
agree: all three changes that improve the model make the decoder more precise in the image
(Section~\ref{sec:exp_ablations}), and the footprint gives the most direct evidence. Each Gaussian's footprint follows
from its center, scale and rotation and the camera, all of which the decoder receives, yet giving it explicitly adds
1.4 dB at full scale, on the held-out and the training objects alike (Table~\ref{tab:fullscale}). The decoder did not
recover the footprint on its own in our training. It also reduces both kinds of error above at high frequencies: the
model reproduces more of the reference's content and adds less that the reference does not have
(Appendix~\ref{sec:appendix_supporting}).

Under-training is not ruled out, however. The final run was still
improving, each training image was seen about four times, the larger inputs were never seen in training, and the
controls above are short fine-tunes of a model trained on the full-splat target. The evidence that more encoder layers
and a wider network do not help comes from an earlier model on 100 objects (Appendix~\ref{sec:appendix_negative}).

To test the encoder directly, we fine-tune the final model on the 14 views of one held-out object, once with only
the view decoder trained and the scene encoder frozen, and once with the whole model, for 10,000 steps each. The two
fit the object about equally well, both several dB above rasterizing its input (Appendix~\ref{sec:appendix_supporting}).
On one object, the frozen scene tokens hold what the decoder
needs to exceed the rasterizer.

Appendix~\ref{sec:appendix_failures} shows the four held-out objects where the model does worst. On all four the
shapes are right, and the error comes from dense texture and soft, low-contrast structure.

\subsection{Beyond the training setup}
\label{sec:exp_beyond}

\noindent\textbf{One camera configuration.} The model was trained in a single camera configuration: $512 \times 512$ pixels, a $45^\circ$ field of view, and
cameras at distances from 1.15 to 2.45. Table~\ref{tab:generalization} in Appendix~\ref{sec:appendix_generalization}
changes one factor at a time, on 30 held-out objects, and shows that the model is a renderer for that configuration, not a general one. At 256 and 1,024 pixels it
drops to about 33 dB. Rasterization also loses about 4 dB at these resolutions, but the model loses about 6. For
cameras much farther away than in training it breaks, at 24.09 dB. Changing the field of view costs only 1.1 dB at
$35^\circ$ and 1.2 dB at $60^\circ$, even when the object fills the frame as in training, while rasterization is
unaffected. Without the footprint, the same model lost 2.4 and 3.2 dB at these fields of view, so telling the decoder how large each Gaussian appears also makes it far less tied to
the training field of view. Cameras closer than in training cost 1.6 dB. Inputs denser than in training do not work
as well: with 40,000 Gaussians the model loses 3.1 dB, against 1.6 dB without the footprint. Varying the resolution, field of view and distance during training would most likely fix the rest. Training on three camera distances instead of one mainly improved close cameras
(Appendix~\ref{sec:appendix_history}), and we expect the same for the other factors. Covering more configurations
takes more training.

\noindent\textbf{Camera paths.} Rendered along a smooth 96-frame orbit that avoids every training camera, the model keeps its quality, 38.52 dB per
frame on average over 30 held-out objects. It renders every frame on its own and was never trained on sequences, yet
the videos show no flicker (project page, Appendix~\ref{sec:appendix_camera_path}).

\noindent\textbf{Real captures.} We also apply the model to regions of four real scenes from the Mip-NeRF 360 dataset~\citep{barron2022mipnerf360}, cut
out of splats we train with gsplat and normalized like our objects, with inputs of up to 160,000 Gaussians
(Appendix~\ref{sec:appendix_real}). The model gets shapes, colors and layout right and
softens fine texture, as on the objects. With 20,000 Gaussians, its training input size, it stays 1.6 to 4.7 dB below
rasterizing the same input on the evaluation crop. With more Gaussians the rasterizer improves and the model gets
worse (Section~\ref{sec:exp_gap}). The regions are viewed from outside on an orbit, so this is not yet rendering of
full scenes.

\section{Limitations}
\label{sec:limitations}

\noindent\textbf{Quality and speed.} \method{} is not yet a practical replacement for a rasterizer. It stays several dB below rasterizing the same input
(Table~\ref{tab:main}), is several hundred times slower and needs far more memory (Table~\ref{tab:cost}), and its
time and memory grow quickly with the input (Appendix~\ref{sec:appendix_cost}).

\noindent\textbf{Objects, not scenes.} We train and evaluate on single objects viewed from outside, and our real captures are regions of scenes viewed the
same way, so rendering a full scene, with the camera inside it, remains open. The full self-attention of the scene
encoder also limits how many Gaussians the model can take (Appendix~\ref{sec:appendix_cost}).

\noindent\textbf{One camera configuration.} \method{} renders well only in its training configuration. Other input sizes, camera distances, resolutions
and fields of view cost it quality, but not the rasterizer (Section~\ref{sec:exp_beyond}).

\noindent\textbf{Appearance.} We use only the view-independent color of each Gaussian, so \method{} cannot render the view-dependent effects that
3DGS stores as spherical harmonics. Unlike RenderFormer, it also renders the colors stored in the Gaussians under
fixed lighting and cannot relight them.

\noindent\textbf{Input preparation.} The renderer itself has no rasterizer, but preparing an input from a larger splat does use one. Pruning to 20,000
Gaussians scores each Gaussian by rasterizing it from 64 views, and the following fine-tune is ordinary 3DGS
optimization (Section~\ref{sec:data}). Inputs of up to 20,000 Gaussians need neither step.

\noindent\textbf{Bound to RenderFormer's design.} Starting from RenderFormer's weights limits how far the network can change. Changes to its pretrained parts undo what
the weights encode: in early experiments, even doubling the frequency of the existing position embedding hurt as much
as replacing it. Only additive changes, new parts initialized to zero such as our 2D embedding, kept the pretrained
behavior. More capacity on top of the pretrained layers trained without trouble but gave nothing: in an earlier
version of the model, on 100 objects, 9 extra layers (287M parameters in total) and an 8 times wider feed-forward
network performed exactly like the original. We could therefore add to RenderFormer's design but not freely redesign
it, and we cannot tell how a network designed for Gaussians from the start would do.

\noindent\textbf{Cost and dependence on RenderFormer.} Training the model is expensive (Appendix~\ref{sec:appendix_reproducibility}), and we found no recipe that
trains it from random weights (Section~\ref{sec:exp_ablations}), so it depends on a pretrained mesh renderer.

\section{Conclusion}
\label{sec:conclusion}

We presented \method{}, a transformer that renders a given set of 3D Gaussians into an image with no rasterizer in the
renderer. Starting from RenderFormer's weights, a token per Gaussian, a 2D rotary embedding of where each Gaussian
lands in the image, its projected footprint and smaller patches are enough for a network to render objects it has
never seen, in the camera configuration it was trained in. A clear gap to rasterizing the same input remains. It is not a generalization gap,
and our diagnostics are consistent with a bottleneck in the view decoder, though under-training is not ruled out. A
learned renderer for Gaussians is feasible, but not yet competitive with a rasterizer.

Each limitation in Section~\ref{sec:limitations} points to future work. Giving
each Gaussian material parameters, as relightable Gaussian methods do, and adding light sources to the scene
would let a model like ours learn relighting and global illumination, the goal that motivates this work (Section~\ref{sec:introduction}).

\bibliography{references}
\bibliographystyle{tmlr}

\clearpage
\appendix
\section{Camera paths}
\label{sec:appendix_camera_path}

We render 30 held-out objects along a smooth path of 96 frames that makes one full turn around the object. The
camera distance varies smoothly between 1.35 and 2.05, inside the training range, and the elevation rises and falls
twice per turn. The angles are offset so that no frame repeats a training camera. Every frame is rendered by the
model from the 20,000-Gaussian input, by gsplat from the same input, and by gsplat from the full splat (the
reference).

We report the PSNR of each frame against the reference, on the evaluation crop as in the main evaluation, and
a simple measure of consistency between frames. For each pair of consecutive frames we take the change from one frame
to the next, for the model and for the reference, and compute the PSNR between the two changes over the whole frame.
A model that flickers adds changes the reference does not have and scores lower. This measure does not separate
flicker from ordinary per-frame error, so we compare it with gsplat on the input and also inspect the videos.
Table~\ref{tab:camera_path} shows the results and Figure~\ref{fig:filmstrip} shows frames of one path.

\begin{table}[ht]
  \centering
  \caption{Rendering along a 96-frame camera path, mean over 30 held-out objects.}
  \label{tab:camera_path}
  \begin{tabular}{@{}lcc@{}}
    \toprule
     & Ours & gsplat on the input \\
    \midrule
    PSNR per frame (dB) & 38.52 & 44.12 \\
    Worst frame per object (dB) & 35.70 & \\
    Spread over frames, standard deviation (dB) & 1.38 & \\
    Frame-to-frame consistency, PSNR (dB) & 42.98 & 47.74 \\
    \bottomrule
  \end{tabular}
\end{table}

The model's per-frame quality along the path matches its quality on the evaluation views (38.36 dB on the 4 views of
the held-out objects), so it does not depend on the training cameras. Its consistency score is 4.8 dB below gsplat's,
a little less than its per-frame gap of 5.6 dB, and the videos on the project page show no visible
flicker.

\newcommand{\filmimg}[1]{\includegraphics[width=0.118\textwidth]{figures/camera_path/#1}}
\begin{figure}[ht]
  \centering
  \setlength{\tabcolsep}{0.5pt}
  \begin{tabular}{@{}cccccccc@{}}
    \filmimg{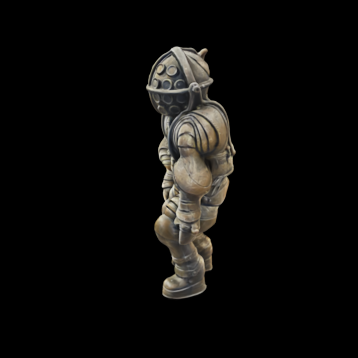} & \filmimg{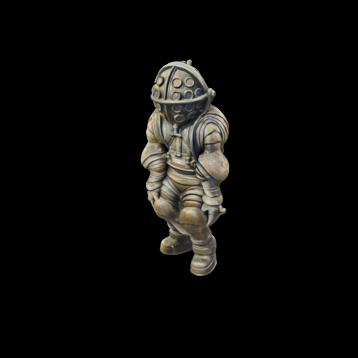} & \filmimg{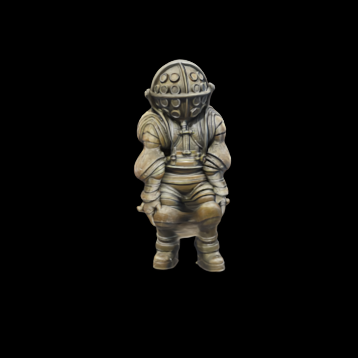} & \filmimg{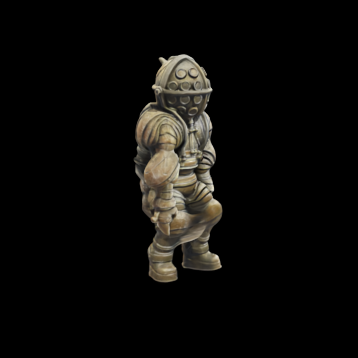} &
    \filmimg{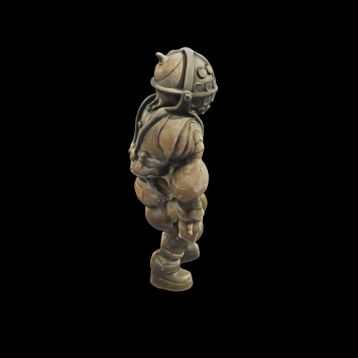} & \filmimg{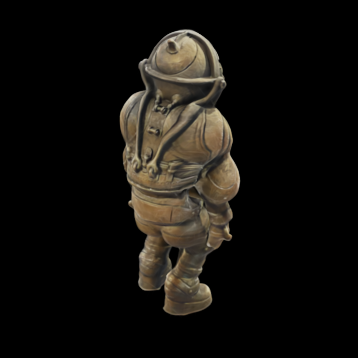} & \filmimg{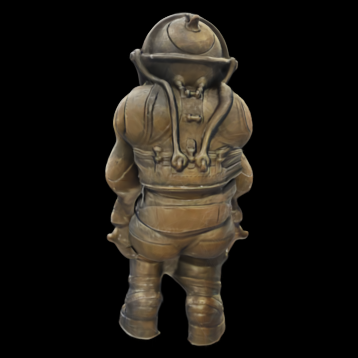} & \filmimg{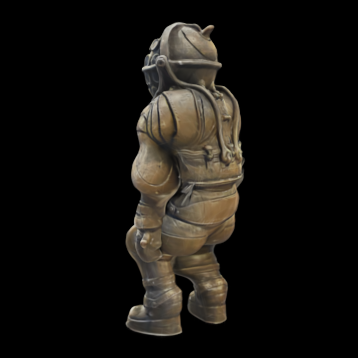} \\
    \filmimg{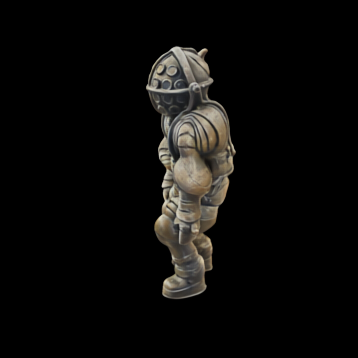} & \filmimg{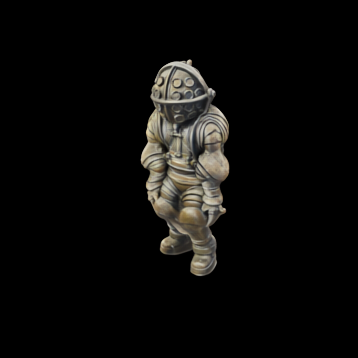} & \filmimg{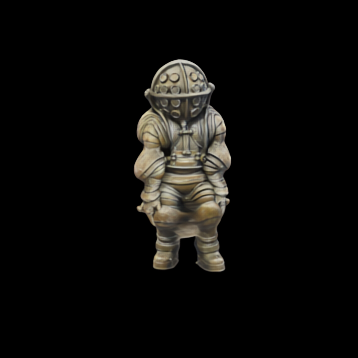} & \filmimg{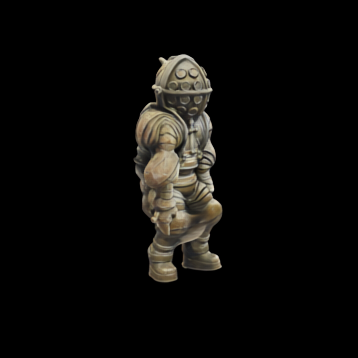} &
    \filmimg{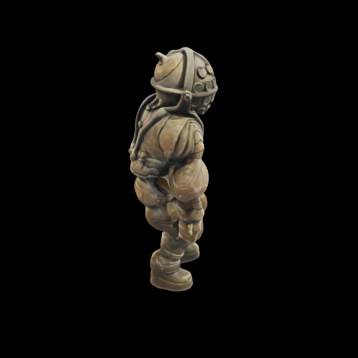} & \filmimg{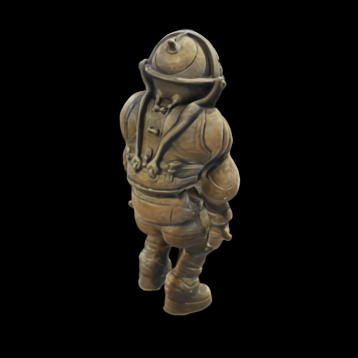} & \filmimg{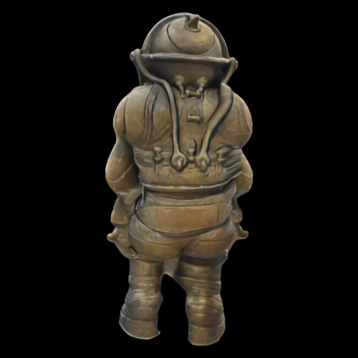} & \filmimg{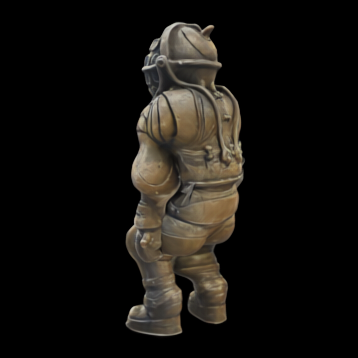} \\
  \end{tabular}
  \caption{Every 12th frame of a 96-frame camera path around a held-out object. Top: the reference. Bottom:
  \method{}, which renders each frame independently.}
  \label{fig:filmstrip}
\end{figure}

\section{Real captures}
\label{sec:appendix_real}

We use four scenes of the Mip-NeRF 360 dataset: bonsai, counter, garden and kitchen. The dataset's project page states no
license, but the project page of SMERF, by an overlapping group of authors, states that the Mip-NeRF
360 scenes are released under CC BY 4.0.\footnote{\url{https://smerf-3d.github.io/\#data}} No trained splat of them
carries a clear license, so we train our own. We use gsplat with its standard
recipe and densification, 30,000 steps on the images downsampled 4 times, with only the view-independent color, and
hold out every 8th image. We then cut out a cube around the point the cameras look at, drop Gaussians larger than a
quarter of the cube, and normalize the region like our training objects. The reference is a rasterization of every
Gaussian in the region, the input is the region pruned with our pipeline (Section~\ref{sec:data}), and we render
7 orbit views above the horizon at distance 1.7. Table~\ref{tab:real_captures} lists the results. The PSNR here is
over the whole frame, which is mostly background, so it is not comparable to the object-level numbers in the main
text.

\begin{table}[ht]
  \centering
  \caption{Real captures from Mip-NeRF 360. Splat: our gsplat reconstruction and its PSNR on held-out photos.
  Region: the cube cut out of it. PSNR in dB over the whole frame, mean over 7 views.}
  \label{tab:real_captures}
  \small
  \begin{tabular}{@{}lrcrccc@{}}
    \toprule
    Scene & Splat & Photos & Region & Input & Ours & gsplat on the input \\
    \midrule
    Bonsai & 2.00M & 31.04 & 346,027 & 20k / 80k / 160k & 34.12 / 33.84 / 31.09 & 35.70 / 43.67 / 48.93 \\
    Counter & 1.81M & 28.11 & 732,862 & 20k / 80k / 160k & 31.78 / 29.92 / 28.40 & 34.37 / 37.96 / 40.86 \\
    Garden & 7.81M & 26.84 & 994,061 & 20k / 80k / 160k & 30.56 / 29.65 / 27.33 & 32.58 / 37.53 / 40.23 \\
    Kitchen & 2.38M & 30.63 & 1,284,974 & 20k / 80k / 160k & 31.99 / 28.61 / 25.81 & 36.74 / 41.55 / 44.86 \\
    \bottomrule
  \end{tabular}
\end{table}

The model does best at its training input size of 20,000 Gaussians, where it stays 1.6 to 4.7 dB below rasterizing
the same input on the evaluation crop (29.8 against 31.4 dB on bonsai, 29.7 against 32.3 on counter, 28.9 against
30.9 on garden and 30.6 against 35.4 on kitchen, means over the 7 views). With 80,000 and 160,000 Gaussians the
rasterizer improves by up to 13 dB, while the model gets worse (Section~\ref{sec:exp_gap}).
Figure~\ref{fig:real_all} shows the best and the worst view of each scene, by the model's PSNR on the crop, with
20,000 Gaussians. In all four scenes the model gets the shapes, colors and layout right and softens fine texture,
most visibly on the bonsai's foliage and on the objects on the counter. Gaussians cut in half by the region's
boundary, and long thin Gaussians at the edge of what the photos covered, show up as blurry streaks in both the
reference and the model, which smears them further.

\newcommand{\realallimg}[1]{\raisebox{-.5\height}{\includegraphics[width=0.135\textwidth]{figures/real/#1}}}
Figure~\ref{fig:real_scene} shows the bonsai region at the three input sizes.

\newcommand{\realimg}[1]{\includegraphics[width=0.3\textwidth]{figures/bonsai/#1}}
\begin{figure}[t]
  \centering
  \setlength{\tabcolsep}{1.5pt}
  \footnotesize
  \begin{tabular}{@{}ccc@{}}
    \realimg{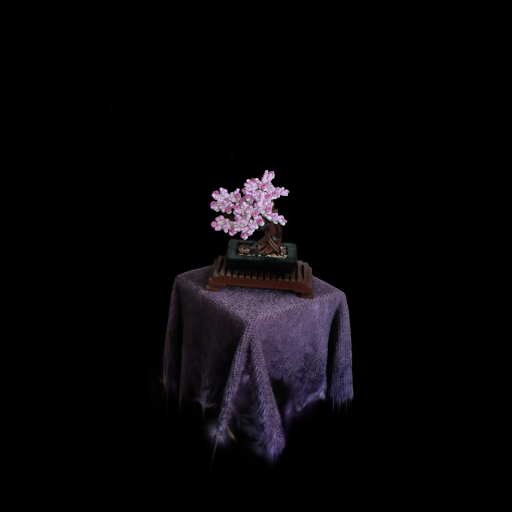} & \realimg{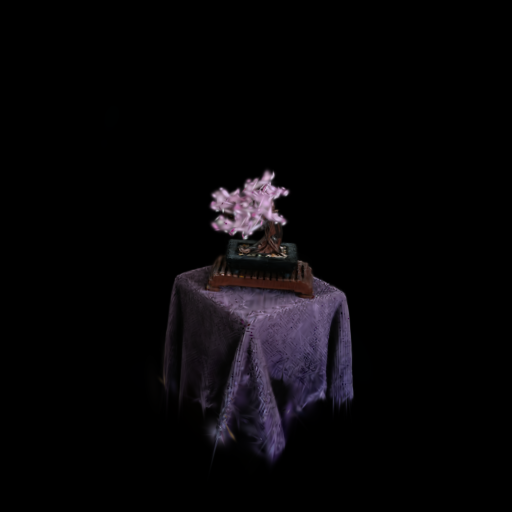} & \realimg{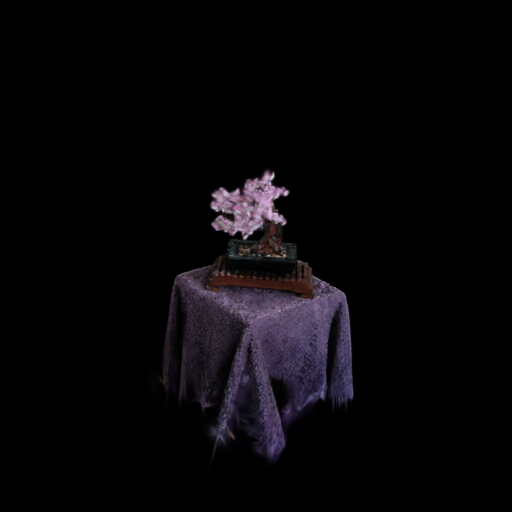} \\
    \realimg{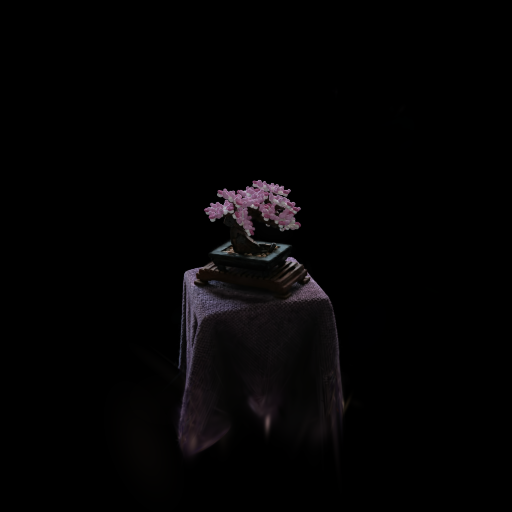} & \realimg{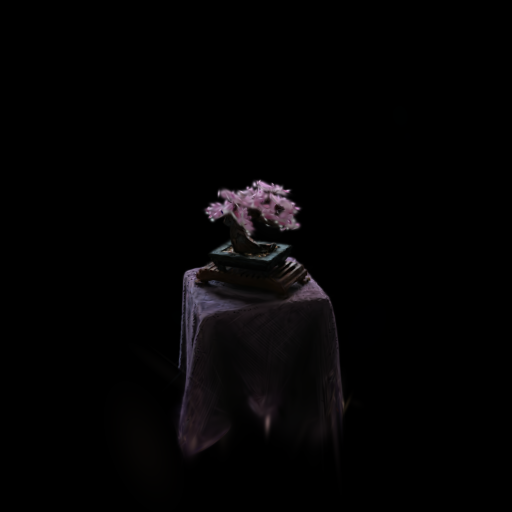} & \realimg{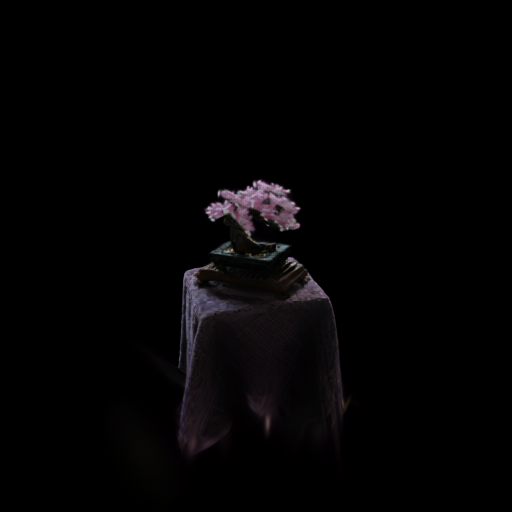} \\
    Full region, rasterized & 20k-Gaussian input, rasterized & Ours, 20k-Gaussian input \\
  \end{tabular}
  \caption{A region of the bonsai scene from the Mip-NeRF 360 dataset, a real capture, seen from two orbit cameras.
  Left: the full region (346,027 Gaussians) rasterized with gsplat, used as the reference. Middle: the region pruned
  to 20,000 Gaussians, \method{}'s training input size, rasterized with gsplat. Right: \method{} on the same pruned
  input. Over the full frame, \method{} scores 34.1 dB and gsplat 35.7 dB against the reference.}
  \label{fig:real_scene}
\end{figure}

\begin{figure}[p]
  \centering
  \setlength{\tabcolsep}{1pt}
  \renewcommand{\arraystretch}{0.6}
  \footnotesize
  \begin{tabular}{@{}c@{\hspace{3pt}}ccccc@{}}
    & Reference & gsplat on the input & Ours & Error, gsplat & Error, Ours \\[2pt]
    \rotatebox[origin=c]{90}{Bonsai, best} & \realallimg{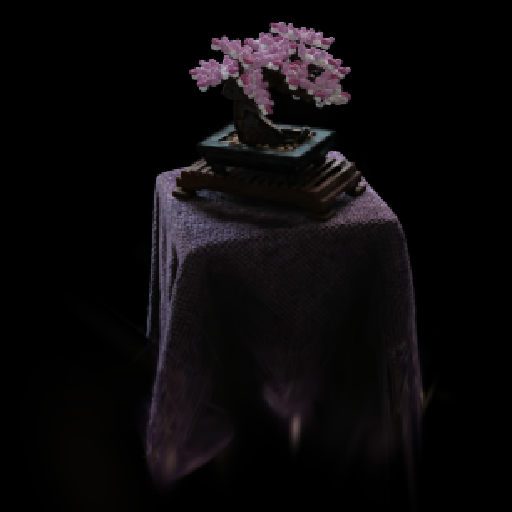} & \realallimg{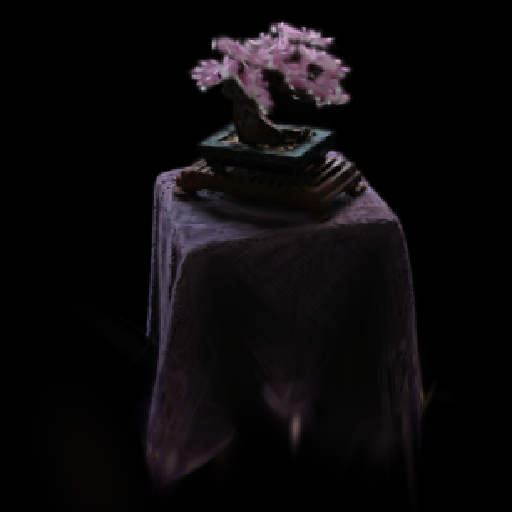} & \realallimg{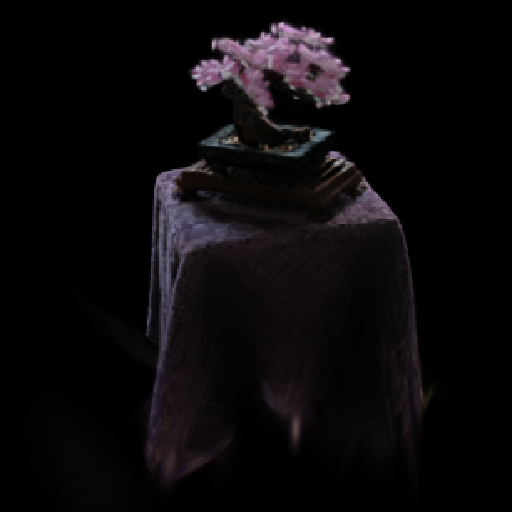} & \realallimg{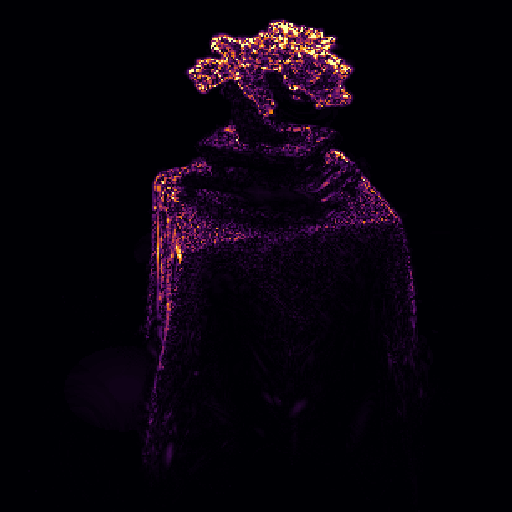} & \realallimg{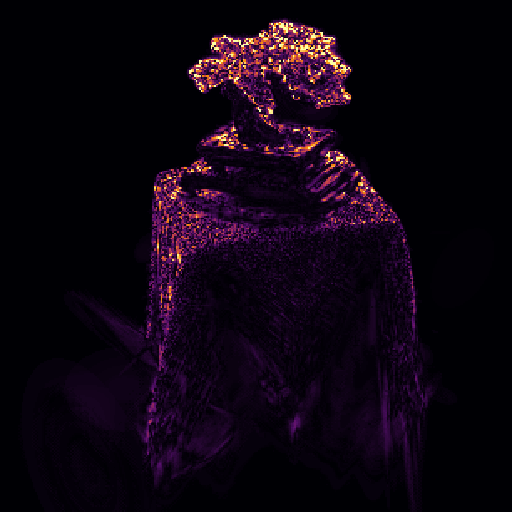} \\
    \rotatebox[origin=c]{90}{Bonsai, worst} & \realallimg{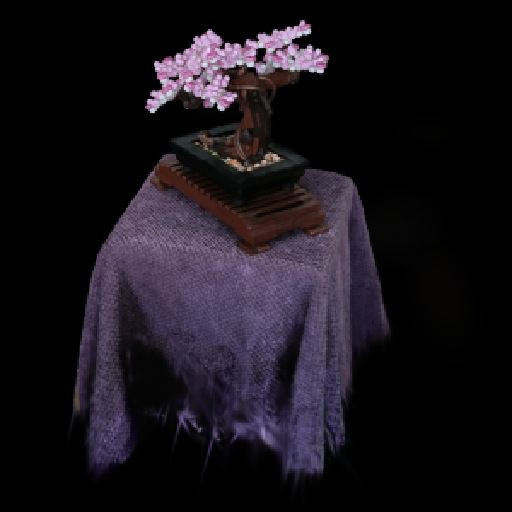} & \realallimg{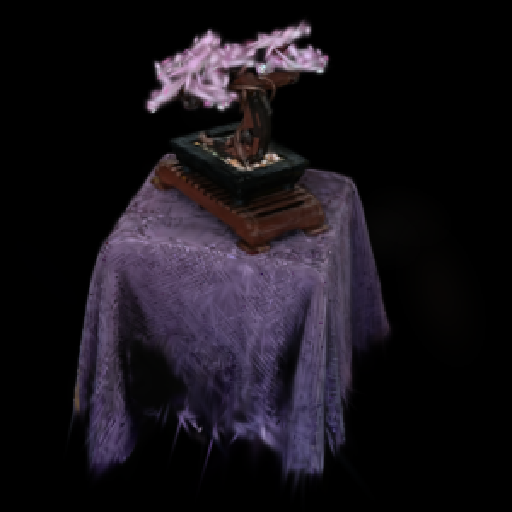} & \realallimg{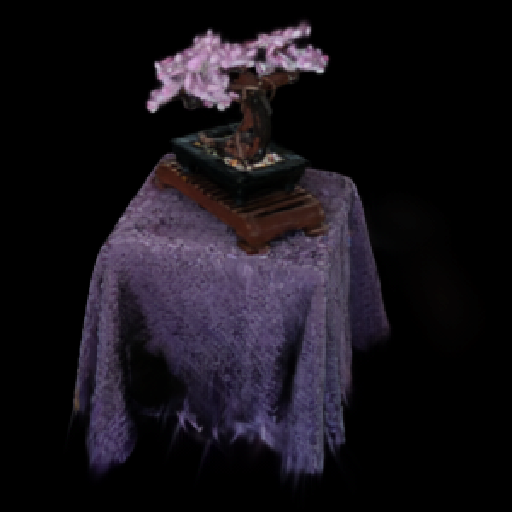} & \realallimg{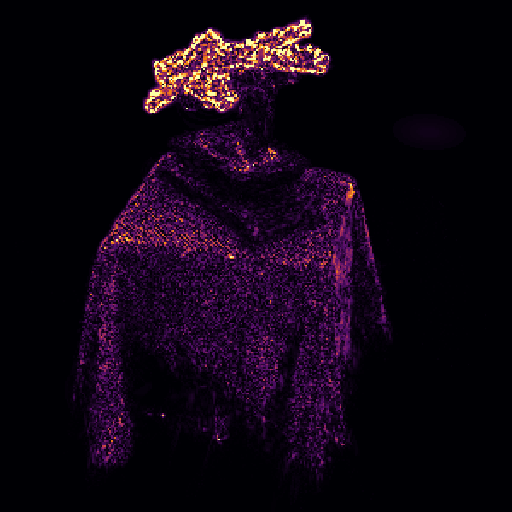} & \realallimg{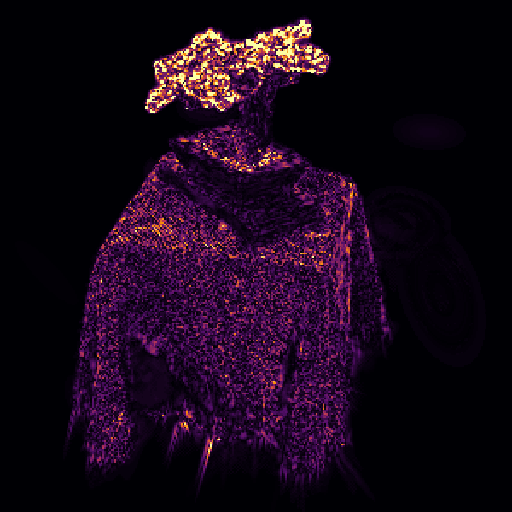} \\
    \rotatebox[origin=c]{90}{Counter, best} & \realallimg{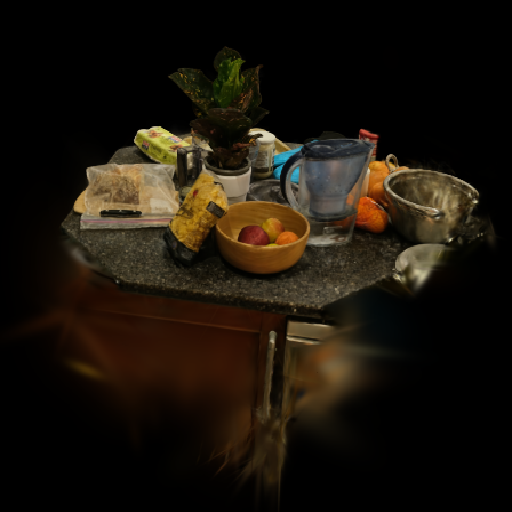} & \realallimg{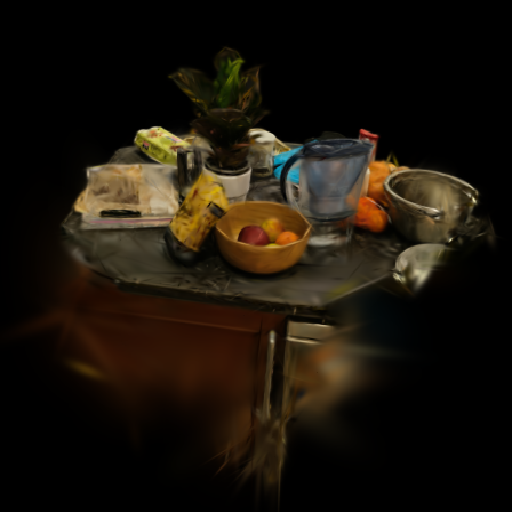} & \realallimg{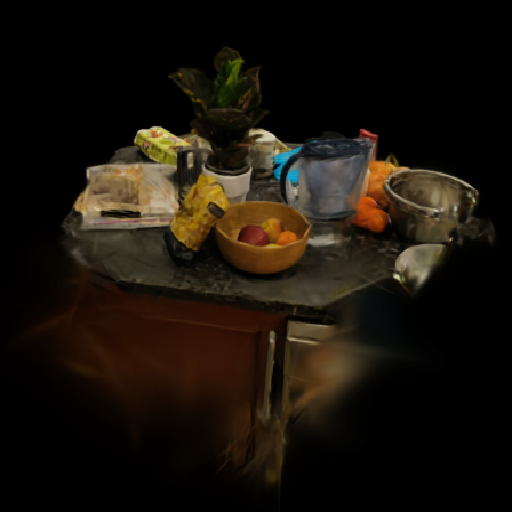} & \realallimg{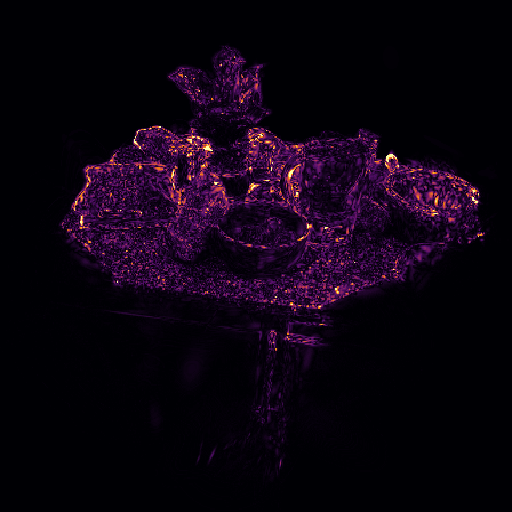} & \realallimg{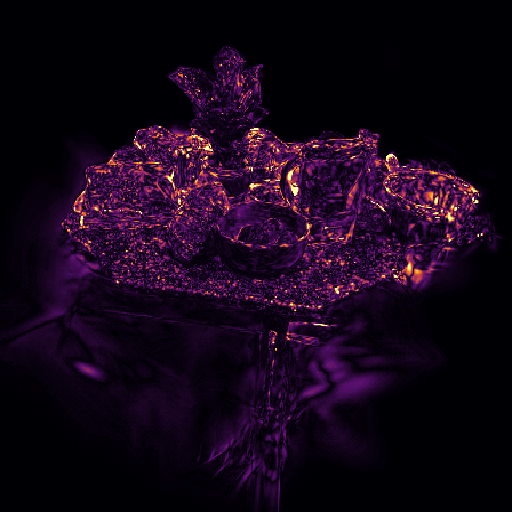} \\
    \rotatebox[origin=c]{90}{Counter, worst} & \realallimg{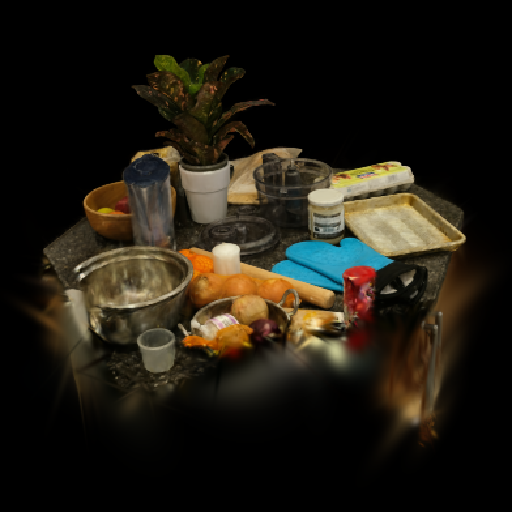} & \realallimg{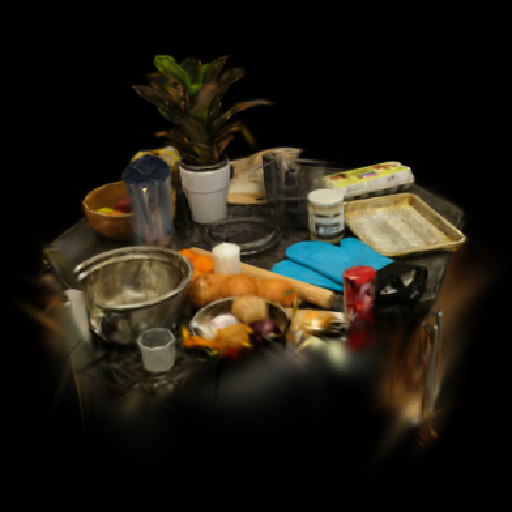} & \realallimg{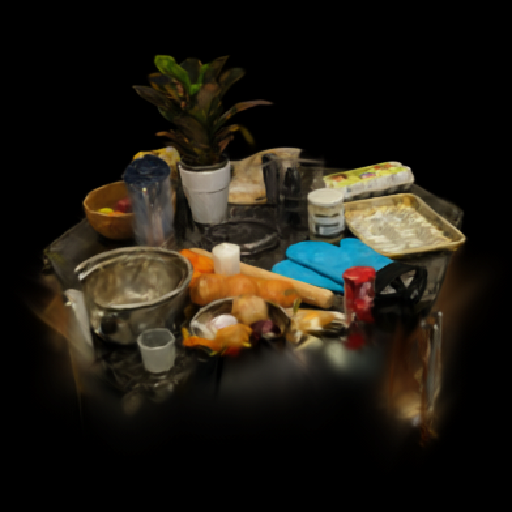} & \realallimg{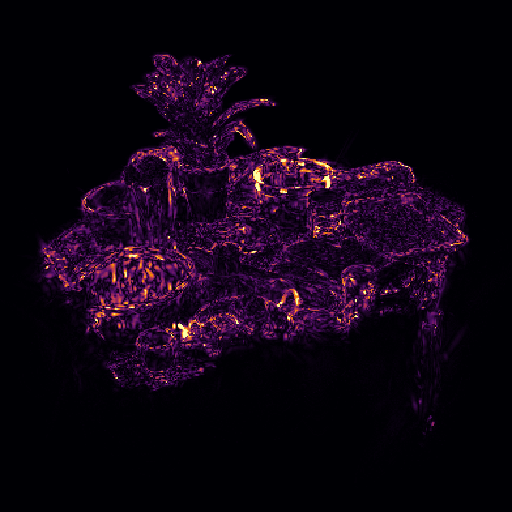} & \realallimg{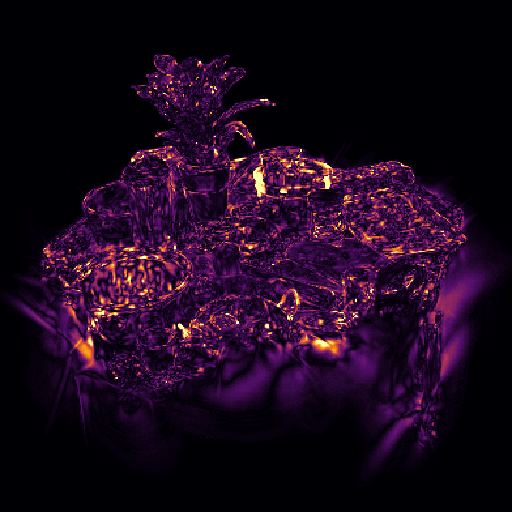} \\
    \rotatebox[origin=c]{90}{Garden, best} & \realallimg{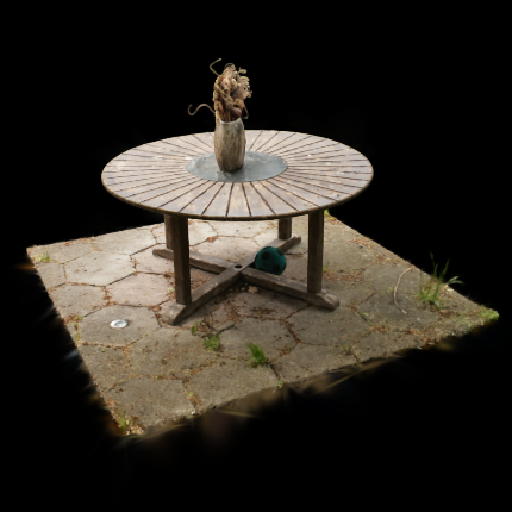} & \realallimg{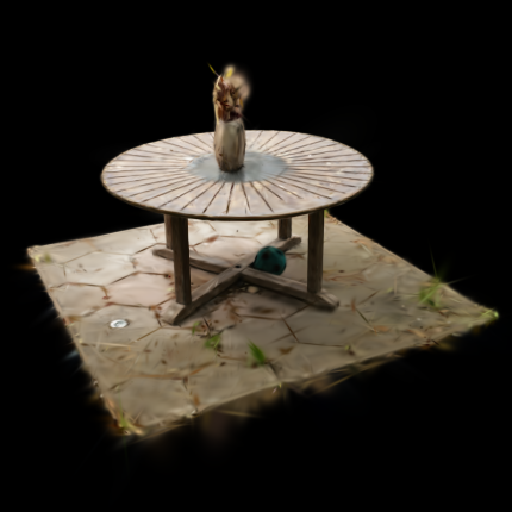} & \realallimg{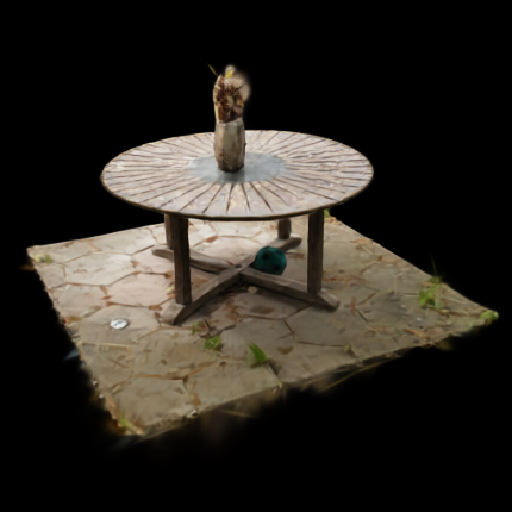} & \realallimg{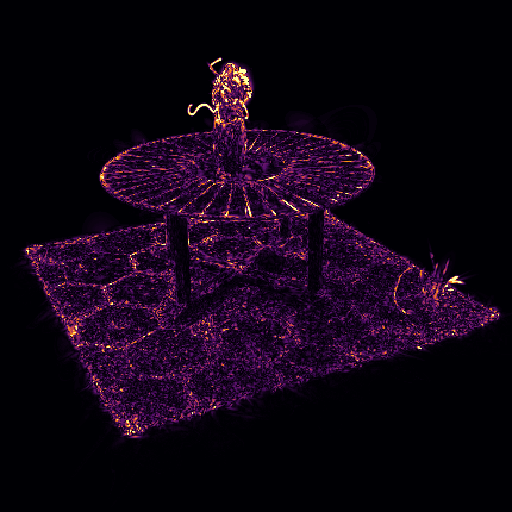} & \realallimg{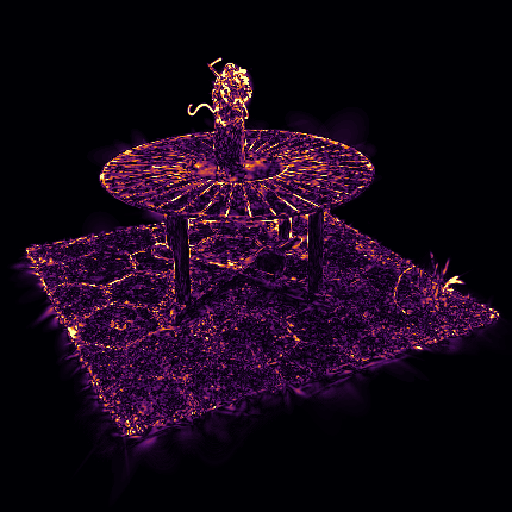} \\
    \rotatebox[origin=c]{90}{Garden, worst} & \realallimg{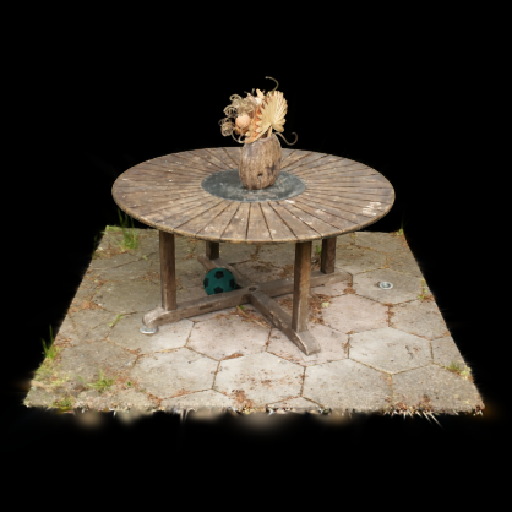} & \realallimg{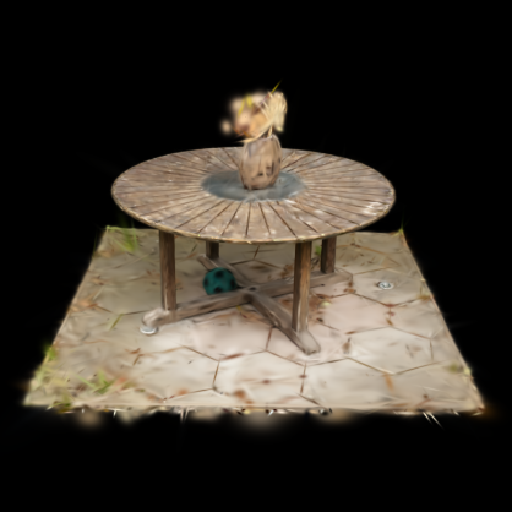} & \realallimg{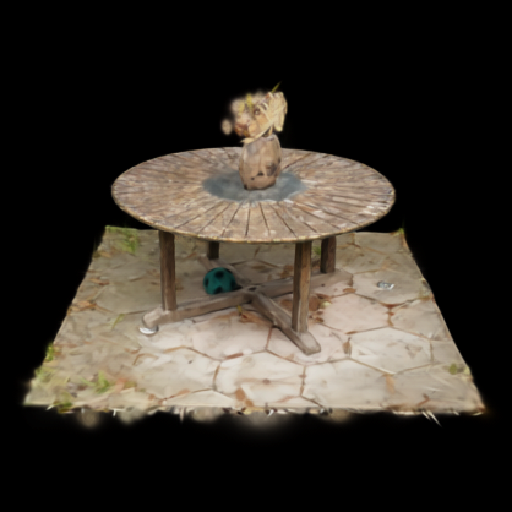} & \realallimg{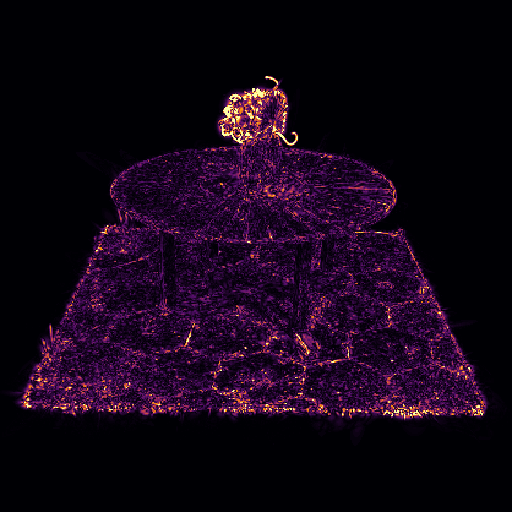} & \realallimg{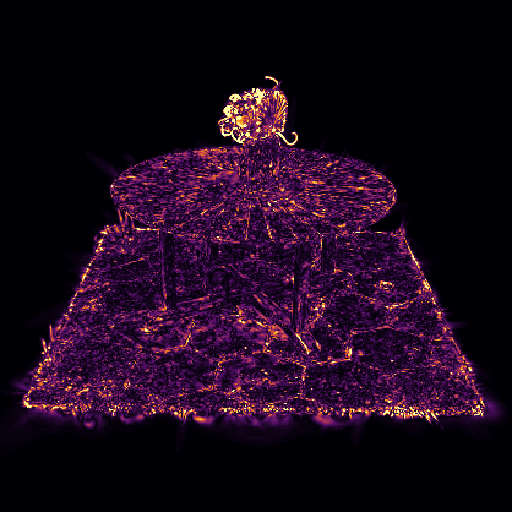} \\
    \rotatebox[origin=c]{90}{Kitchen, best} & \realallimg{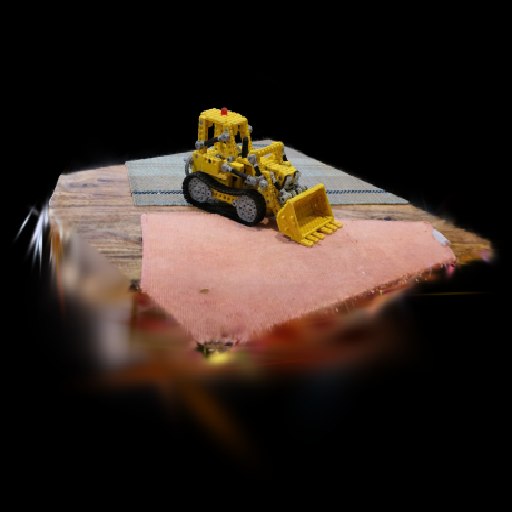} & \realallimg{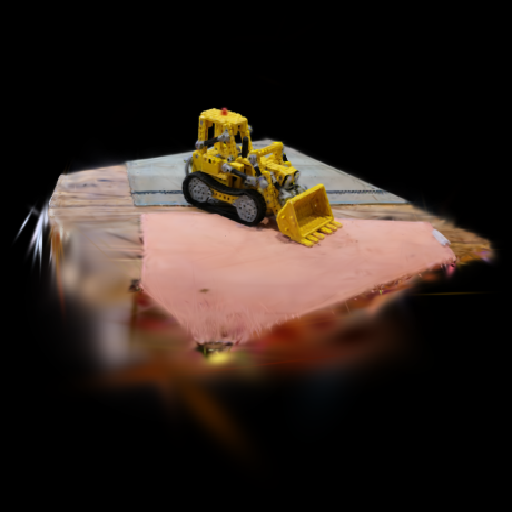} & \realallimg{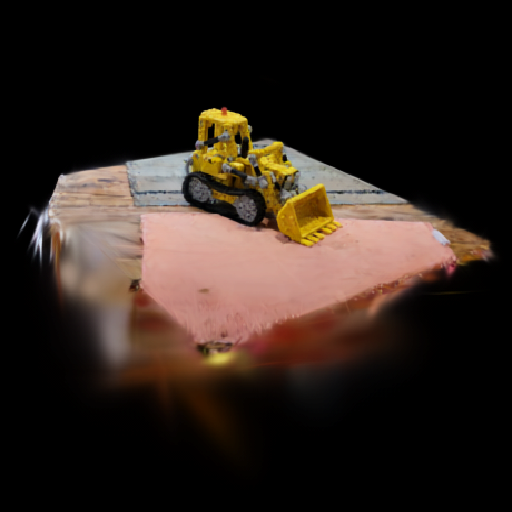} & \realallimg{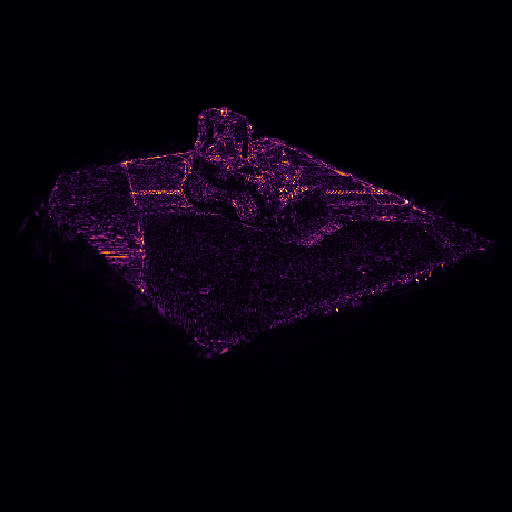} & \realallimg{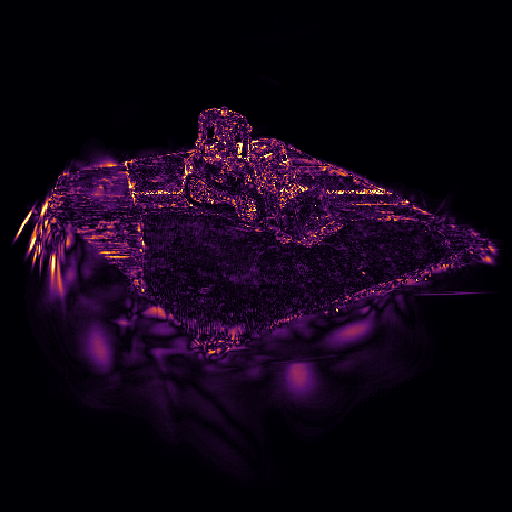} \\
    \rotatebox[origin=c]{90}{Kitchen, worst} & \realallimg{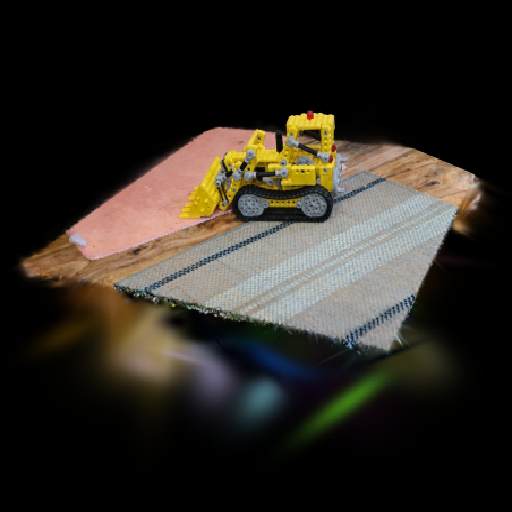} & \realallimg{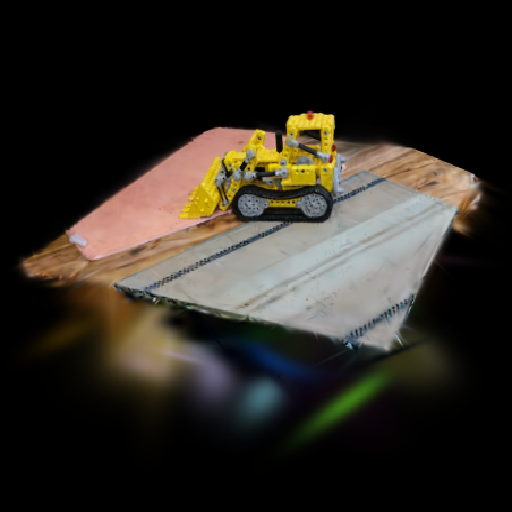} & \realallimg{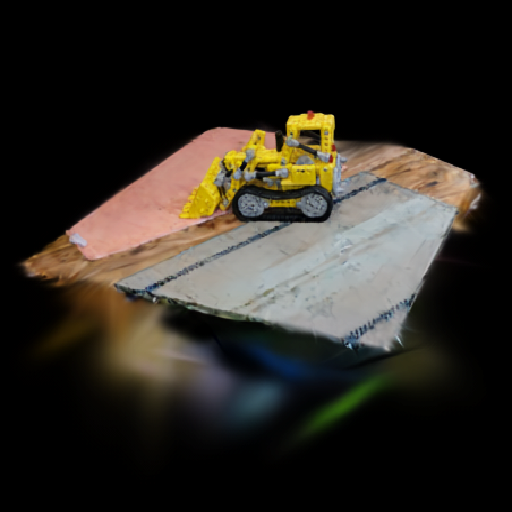} & \realallimg{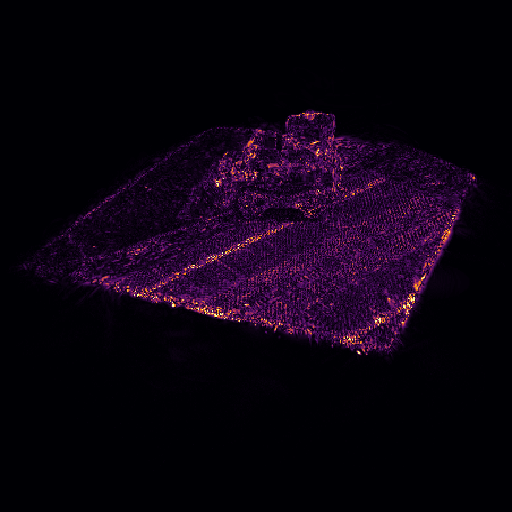} & \realallimg{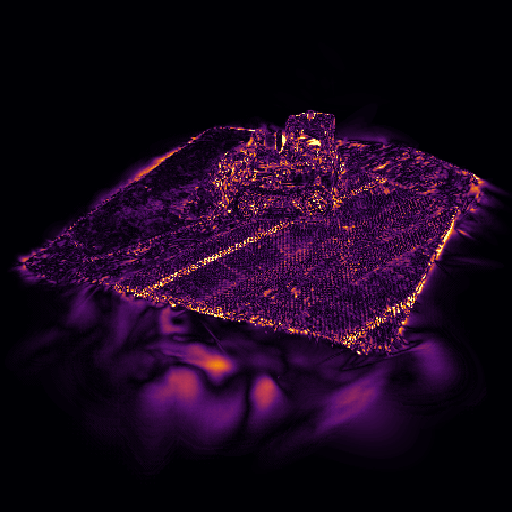} \\
  \end{tabular}
  \caption{The four real captures with 20,000 Gaussians, each at its best and its worst view by \method{}'s PSNR on the
  evaluation crop, with errors scaled by 5 as in Figure~\ref{fig:qualitative}. Bonsai: views 4 and 0, \method{} 31.6
  and 28.3 dB, gsplat on the input 32.8 and 29.8 dB. Counter: views 6 and 4, 30.7 and 29.1 dB against 33.2 and 31.9 dB.
  Garden: views 4 and 0, 29.4 and 28.4 dB against 31.4 and 30.3 dB. Kitchen: views 6 and 3, 31.9 and 29.8 dB against
  36.9 and 33.7 dB.}
  \label{fig:real_all}
\end{figure}

\section{Additional results}
\label{sec:appendix_additional}

\subsection{Supporting measurements}
\label{sec:appendix_supporting}
\paragraph{Harness schedule.} Every harness arm starts from the end of stage 2 and trains for 3,000 steps at 256
pixels with log-L1 only (warmup of 101 steps, cosine decay over the last 1,000), then for 33,000 steps at 512 pixels
with log-L1 and LPIPS (warmup of 1,000 steps, decay over the last 3,000), at a learning rate of $5 \times 10^{-5}$ with
a batch of 8 views on 8 GPUs.

\paragraph{Full scale.} Table~\ref{tab:fullscale} repeats the ablations of Section~\ref{sec:exp_ablations} at full
scale. The arms before the footprint are trained on all 26,820 objects for 94,000 steps plus a 10,000-step decay,
the first milestone of the final run, and evaluated on the 4 views of the development held-out objects. The footprint
pair is the final model and the same recipe trained without the footprint, on all 28 views. With the footprint, the
model reproduces more of the reference's content above 1/4 cycles per pixel, 85\% against 83\% without it (measured
as in Appendix~\ref{sec:appendix_frequency}), and the content it adds that is not in the reference falls by 24 to 38\%
between 1/16 and 1/4 cycles per pixel.

\begin{table}[ht]
  \centering
  \caption{Ablations at full scale. PSNR in dB against the unpruned render, the difference from the model with all
  changes with a 95\% bootstrap interval over objects, and LPIPS.}
  \label{tab:fullscale}
  \small
  \begin{tabular}{@{}llccc@{}}
    \toprule
    Objects and views & Model & PSNR & Difference & LPIPS \\
    \midrule
    Development held-out, 4 views & Before the footprint, 94k steps & 34.73 & & 0.022 \\
     & \quad without the 2D embedding & 30.40 & $-4.3$ (4.2 to 4.5) & 0.039 \\
     & \quad with $8 \times 8$ patches & 32.31 & $-2.4$ (2.3 to 2.5) & 0.033 \\
    \midrule
    Fresh held-out, 28 views & Final model & 38.08 & & 0.0113 \\
     & \quad without the footprint & 36.67 & $-1.40$ (1.33 to 1.48) & 0.0140 \\
    Training objects, 28 views & Final model & 38.25 & & 0.0108 \\
     & \quad without the footprint & 36.86 & $-1.39$ & 0.0135 \\
    \bottomrule
  \end{tabular}
\end{table}

\paragraph{Reachable target.} Measured against the rasterization of its own input, the harness trained on that
target scores 33.24 dB on the held-out objects and 35.04 dB on its own training images, against 33.05 and 34.80 dB
for the harness trained on the full splat: 0.19 dB more (95\% interval over objects 0.17 to 0.21 dB, one run per arm).

\paragraph{Frozen encoder.} Fine-tuned on the 14 views of object 604 for 10,000 steps, the final model reaches 45.7 dB
on these views with only the view decoder trained and 46.4 dB with the whole model (47.2 and 46.8 dB a thousand steps
earlier), against 41.8 dB for rasterizing its input and 34.5 dB before the fine-tune.

\subsection{Beyond the training setup}
\label{sec:appendix_generalization}
Table~\ref{tab:generalization} changes one factor at a time from the training setup (Section~\ref{sec:exp_beyond}).

\begin{table}[ht]
  \centering
  \caption{One change at a time from the training setup (20,000 Gaussians, distance 1.7, $512 \times 512$ pixels,
  $45^\circ$ field of view), on 30 held-out objects. PSNR in dB against a rasterization
  of the full splat. Training distances range from 1.15 to 2.45. For the field of view, the camera distance is scaled
  so the object fills the frame as at $45^\circ$. The inputs of other sizes are pruned again from the full splat with
  our pipeline, and pruning again to 20,000 Gaussians gives 38.76 and 45.28 dB, the reference for their
  differences in the text.}
  \label{tab:generalization}
  \small
  \begin{tabular}{@{}lcc@{\hspace{2.5em}}lcc@{}}
    \toprule
    Change & Ours & gsplat on the input & Change & Ours & gsplat on the input \\
    \midrule
    Training setup & 38.74 & 45.24 & & & \\
    \midrule
    5,000 Gaussians & 31.66 & 36.20 & 256 pixels & 33.39 & 41.03 \\
    10,000 Gaussians & 35.59 & 40.10 & 1,024 pixels & 32.72 & 41.18 \\
    40,000 Gaussians & 35.68 & 49.91 & & & \\
    \midrule
    Distance 0.9 & 37.17 & 40.28 & Field of view $35^\circ$ & 37.60 & 45.18 \\
    Distance 1.15 & 38.88 & 42.63 & Field of view $40^\circ$ & 38.25 & 45.20 \\
    Distance 2.45 & 37.63 & 43.69 & Field of view $50^\circ$ & 38.55 & 45.20 \\
    Distance 3.2 & 24.09 & 41.47 & Field of view $55^\circ$ & 38.14 & 45.13 \\
     & & & Field of view $60^\circ$ & 37.56 & 45.10 \\
    \bottomrule
  \end{tabular}
\end{table}

\subsection{Cost and input size}
\label{sec:appendix_cost}
Table~\ref{tab:scaling} gives the time and memory of one view as the input grows from 20,000 to 640,000 Gaussians.

\begin{table}[ht]
  \centering
  \caption{Time and peak memory for one $512 \times 512$ view as the input grows, on one NVIDIA L40S GPU. Inputs are a
  region of the Mip-NeRF 360 garden scene pruned to each size. The model's memory includes its weights, gsplat's
  excludes the input.}
  \label{tab:scaling}
  \begin{tabular}{@{}rrrrr@{}}
    \toprule
    Gaussians & Ours (ms) & Ours memory (GB) & gsplat (ms) & gsplat memory (GB) \\
    \midrule
    20,000 & 275 & 2.9 & 0.7 & 0.007 \\
    40,000 & 613 & 3.0 & 0.8 & 0.009 \\
    80,000 & 1,793 & 3.9 & 1.2 & 0.014 \\
    160,000 & 5,294 & 6.7 & 1.5 & 0.030 \\
    320,000 & 19,367 & 12.4 & 1.9 & 0.044 \\
    640,000 & 74,853 & 23.7 & 2.5 & 0.082 \\
    \bottomrule
  \end{tabular}
\end{table}

\subsection{First training stages}
\label{sec:appendix_stages}
Table~\ref{tab:stage_ablation} compares the initializations and schedules of the first two stages (Section~\ref{sec:exp_ablations}).

\begin{table}[ht]
  \centering
  \caption{The first training stages, 50,000 steps each at $256 \times 256$ pixels on all 26,820 objects. Held-out
  PSNR and LPIPS at 256 pixels against the unpruned render, the difference from our recipe with a 95\% bootstrap
  interval over objects, and validation loss at the end of training.}
  \label{tab:stage_ablation}
  \begin{tabular}{@{}lcccc@{}}
    \toprule
    Initialization and schedule & PSNR (dB) & Difference & LPIPS & Validation loss \\
    \midrule
    RenderFormer, input layer first (ours) & 29.98 & & 0.0783 & 0.00100 \\
    RenderFormer, whole model from the start & 26.85 & $-3.13 \pm 0.10$ & 0.1257 & 0.00148 \\
    RenderFormer, whole model, input layer at $10^{-3}$ & 26.30 & $-3.68 \pm 0.12$ & 0.1383 & 0.00162 \\
    Random weights & 13.12 & & 0.5733 & 0.01357 \\
    \bottomrule
  \end{tabular}
\end{table}

\subsection{Training curve}
\begin{figure}[ht]
  \centering
  \begin{tikzpicture}
    \begin{axis}[width=0.8\textwidth, height=5cm, xlabel={Training step (thousands)}, ylabel={Validation loss},
                 xmin=0, xmax=375, ymin=0.003, legend style={font=\footnotesize}, grid=major, grid style={black!10}]
      \addplot[thick, blue!70!black] table[x=step, y=total] {figures/data/v22_val.dat};
      \addplot[thick, orange!80!black] table[x=step, y=logl1] {figures/data/v22_val.dat};
      \addplot[thick, black!45, dashed] table[x=step, y=total] {figures/data/v21_val.dat};
      \legend{Total, Log-L1 term, Total without the footprint}
    \end{axis}
  \end{tikzpicture}
  \caption{Validation loss of the final run at $512 \times 512$ pixels (stage 4), on the 100 validation objects,
  every 2,000 steps, and of the same recipe trained without the footprint (dashed). The log-L1 term of the first
  16,000 steps was not logged.}
  \label{fig:training_curve}
\end{figure}
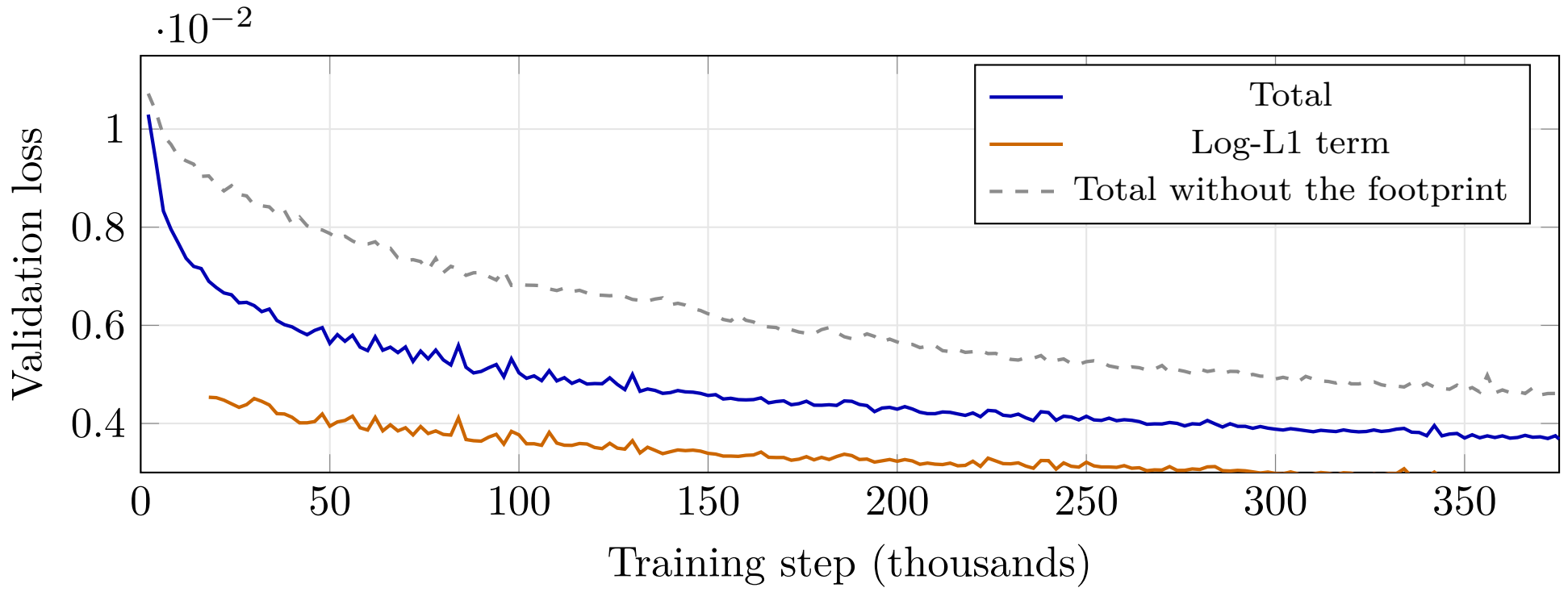

Figure~\ref{fig:training_curve} shows the validation loss of the final run at 512 pixels. Table~\ref{tab:milestones}
shows the held-out PSNR at four points of the run, each after a 10,000-step learning-rate decay. The run was still
improving when it ended, by less in each quarter. Section~\ref{sec:exp_scaling} discusses what that means for the gap.

\begin{table}[ht]
  \centering
  \caption{PSNR of the final run on the fresh held-out objects over their 28 views, at four points, each after a
  10,000-step decay.}
  \label{tab:milestones}
  \begin{tabular}{@{}lcccc@{}}
    \toprule
    Steps before the decay & 94,000 & 188,000 & 282,000 & 375,000 \\
    \midrule
    PSNR (dB) & 36.20 & 37.12 & 37.65 & 38.08 \\
    \bottomrule
  \end{tabular}
\end{table}

\subsection{Renders of the ablations}
Figure~\ref{fig:ablation_renders} shows the harness arms and the full-scale models of Section~\ref{sec:exp_ablations} on objects of
Figure~\ref{fig:qualitative}.

\newcommand{\ablimg}[1]{\includegraphics[width=0.105\textwidth]{figures/ablations/#1}}
\newcommand{\ablhead}[1]{\parbox[b]{0.105\textwidth}{\centering #1}}
\begin{figure}[ht]
  \centering
  \setlength{\tabcolsep}{0.5pt}
  \renewcommand{\arraystretch}{0.6}
  \scriptsize
  \begin{tabular}{@{}ccccccccc@{}}
    & \multicolumn{2}{c}{Full scale} & \multicolumn{4}{c}{Harness} & \multicolumn{2}{c}{Before the footprint, 94k} \\[3pt]
    \cmidrule(lr){2-3} \cmidrule(lr){4-7} \cmidrule(lr){8-9}
    \ablhead{Reference} & \ablhead{Final model} & \ablhead{No footprint} & \ablhead{All changes} & \ablhead{No footprint} & \ablhead{No 2D emb.} & \ablhead{$8{\times}8$ patches} & \ablhead{2D emb.} & \ablhead{No 2D emb.} \\[2pt]
    \ablimg{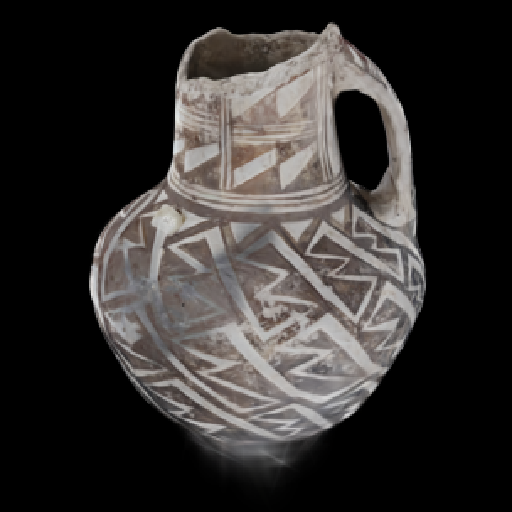} & \ablimg{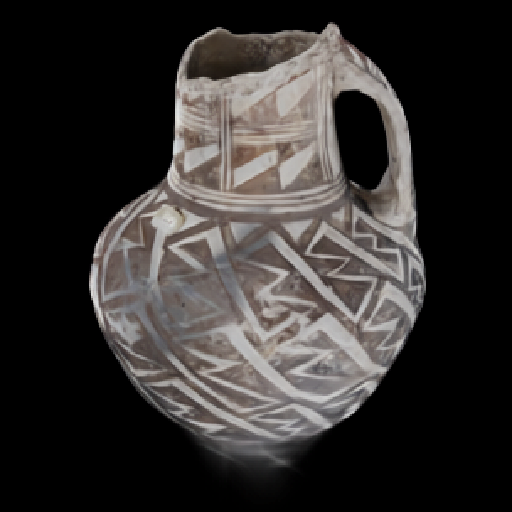} & \ablimg{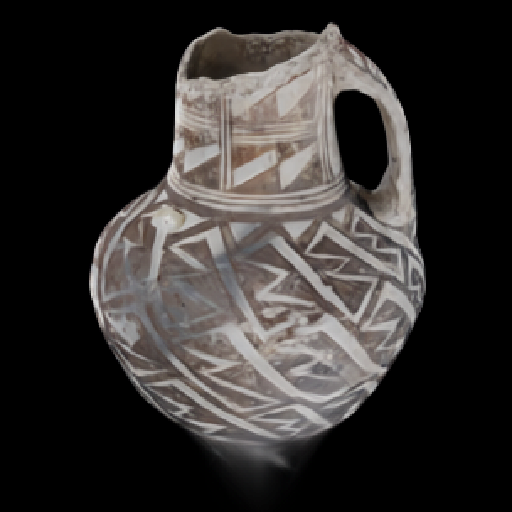} & \ablimg{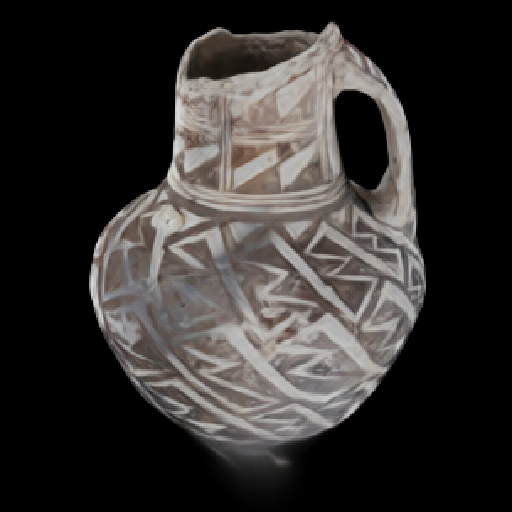} & \ablimg{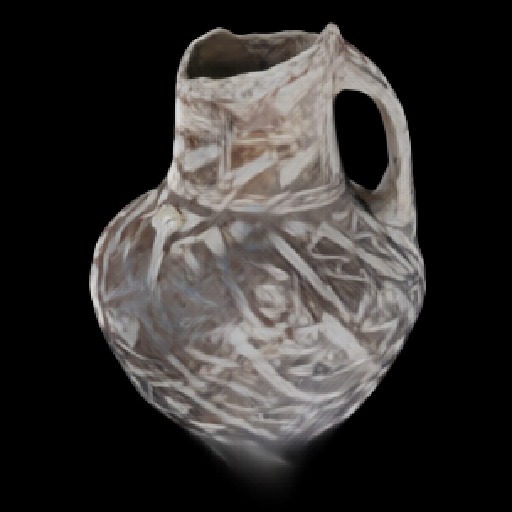} & \ablimg{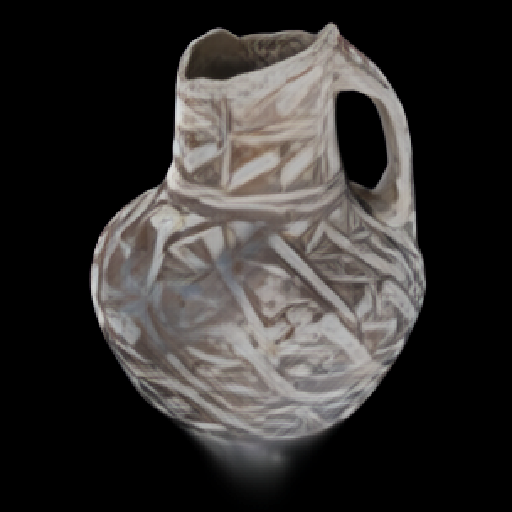} & \ablimg{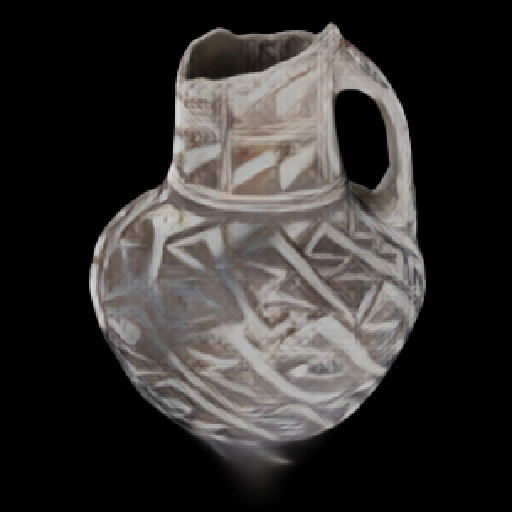} & \ablimg{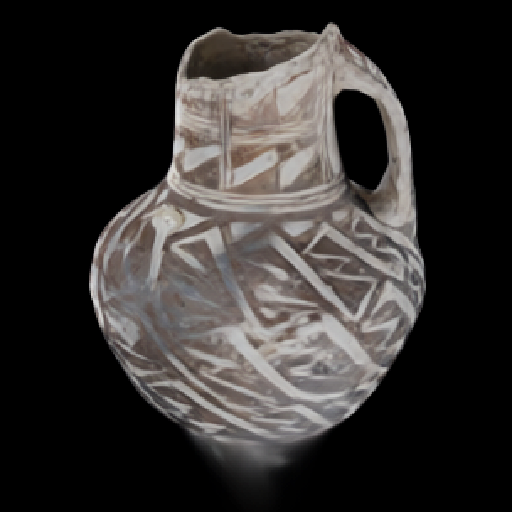} & \ablimg{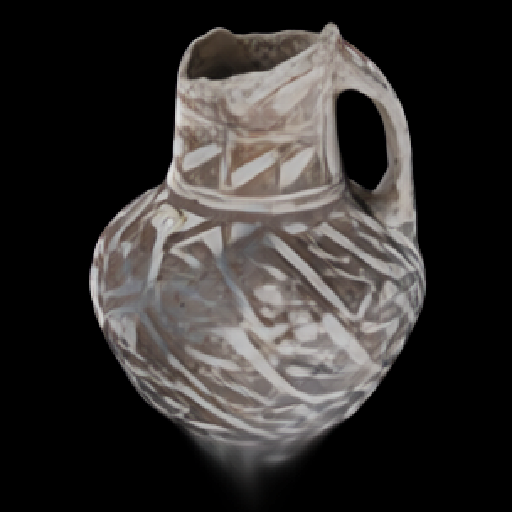} \\
    \ablimg{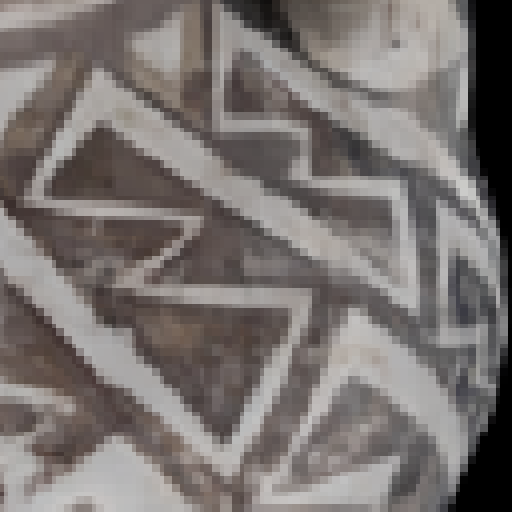} & \ablimg{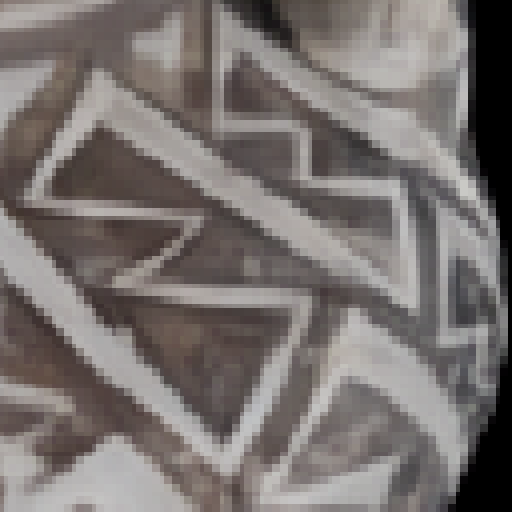} & \ablimg{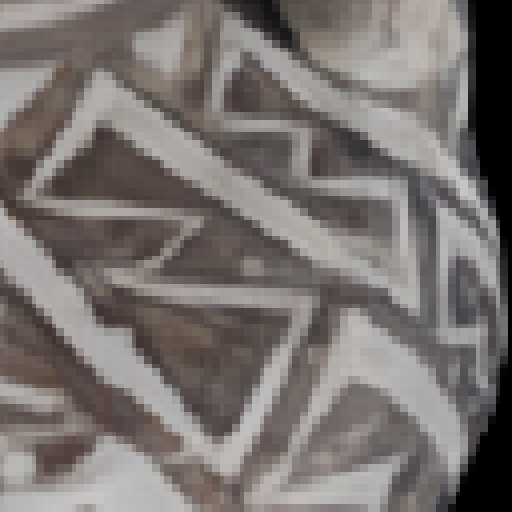} & \ablimg{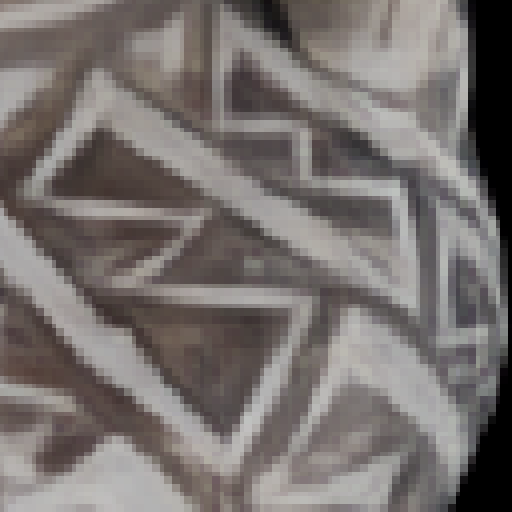} & \ablimg{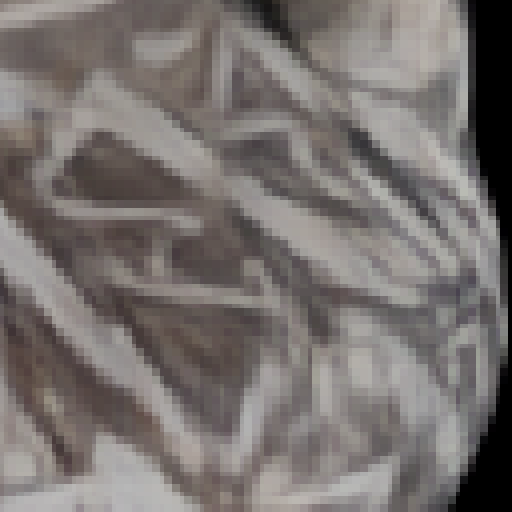} & \ablimg{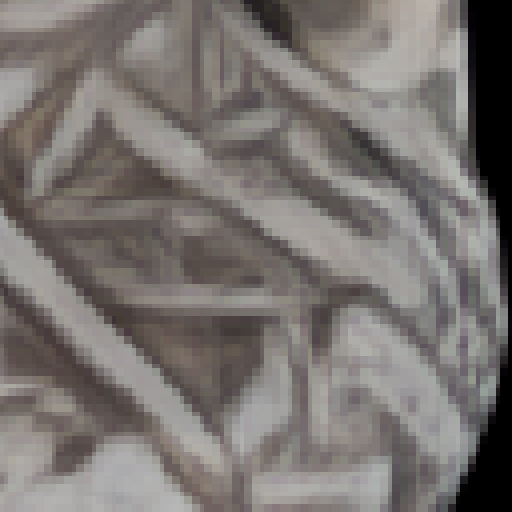} & \ablimg{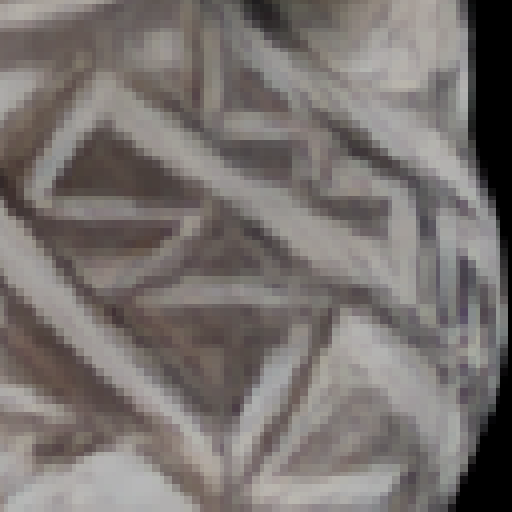} & \ablimg{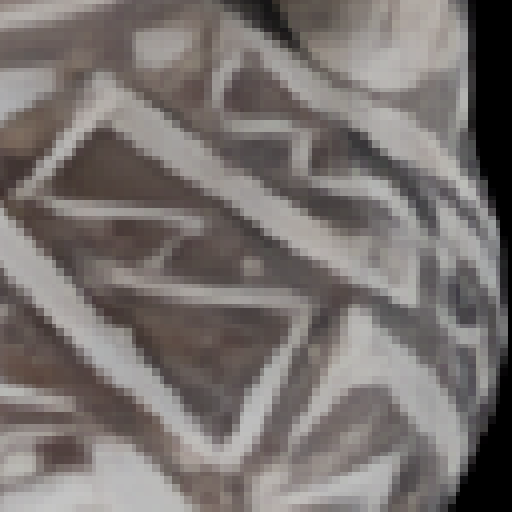} & \ablimg{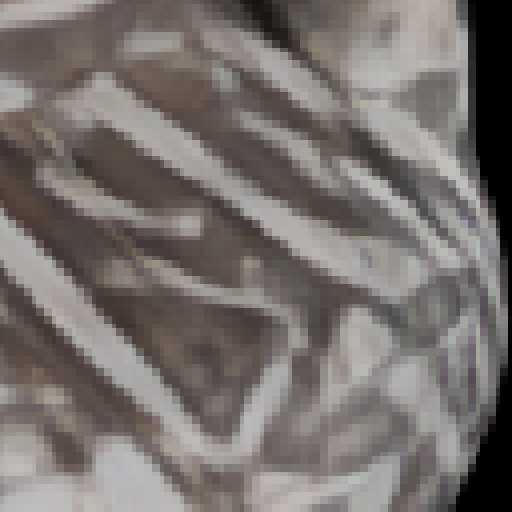} \\
    \ablimg{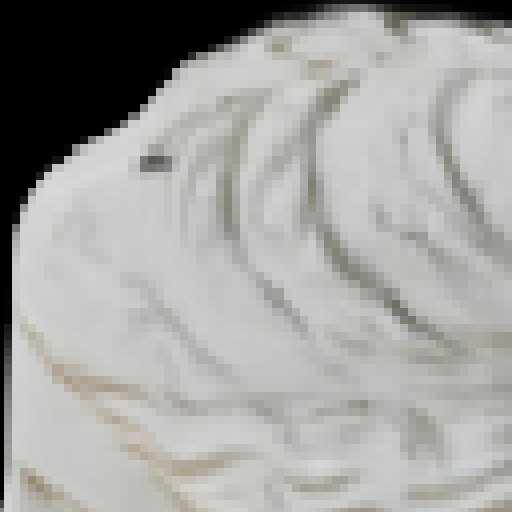} & \ablimg{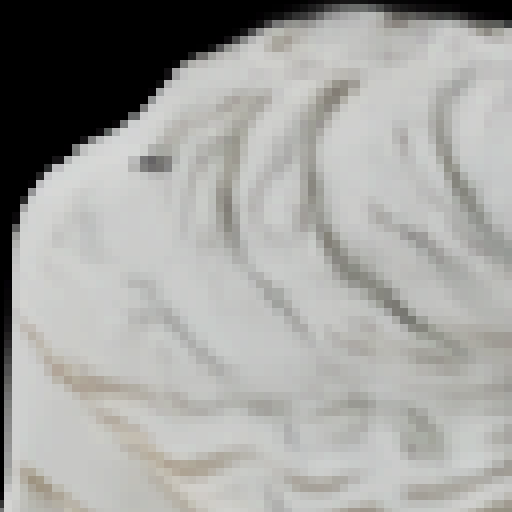} & \ablimg{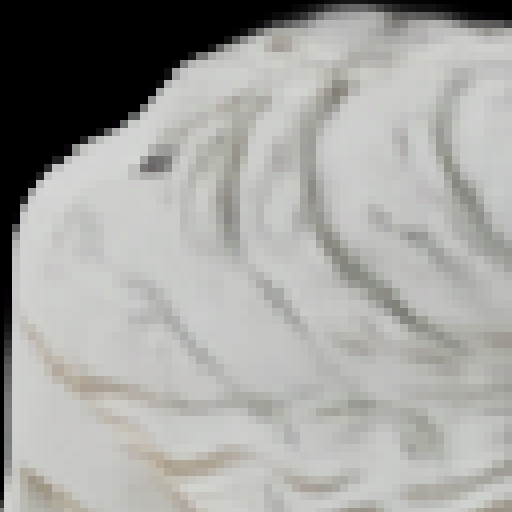} & \ablimg{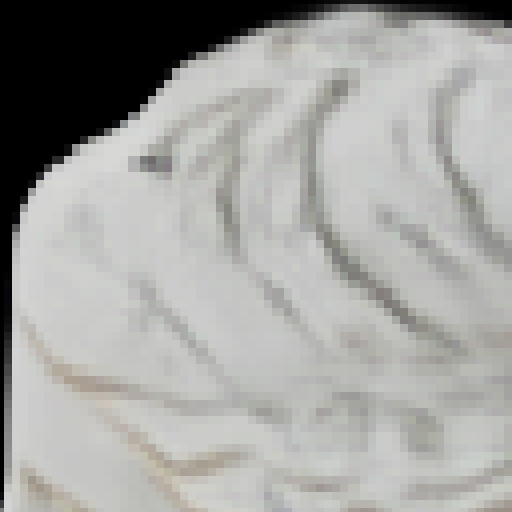} & \ablimg{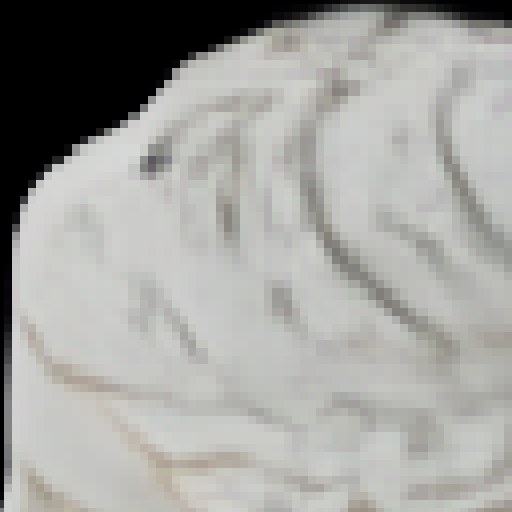} & \ablimg{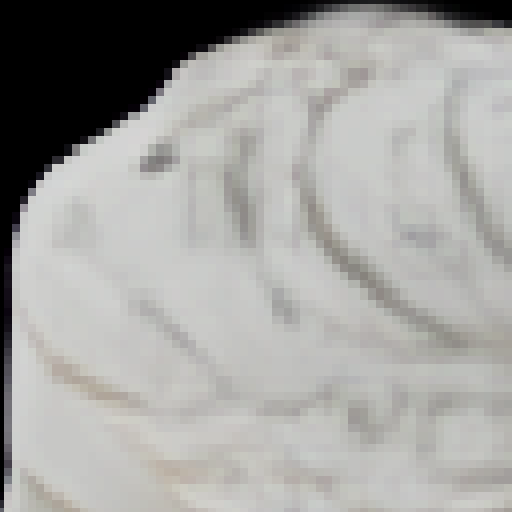} & \ablimg{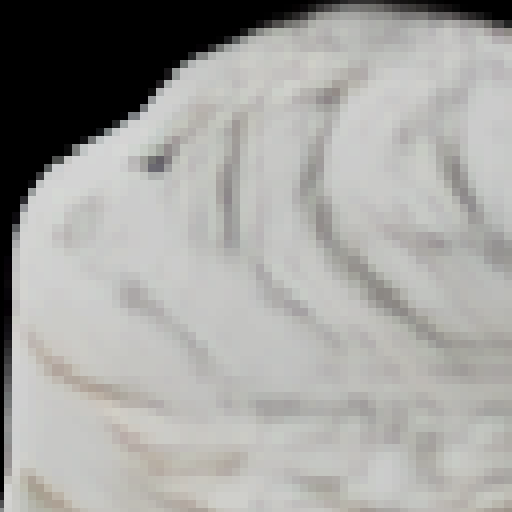} & \ablimg{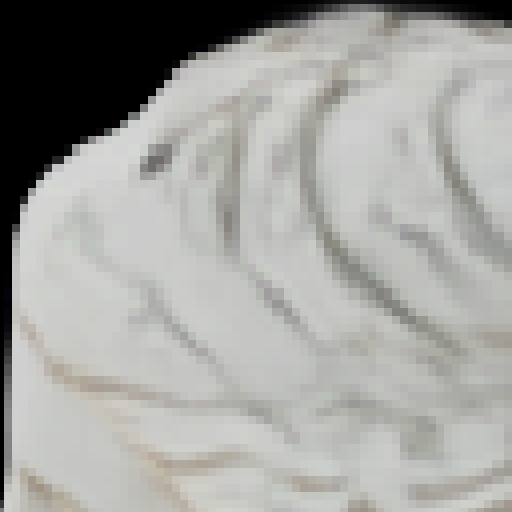} & \ablimg{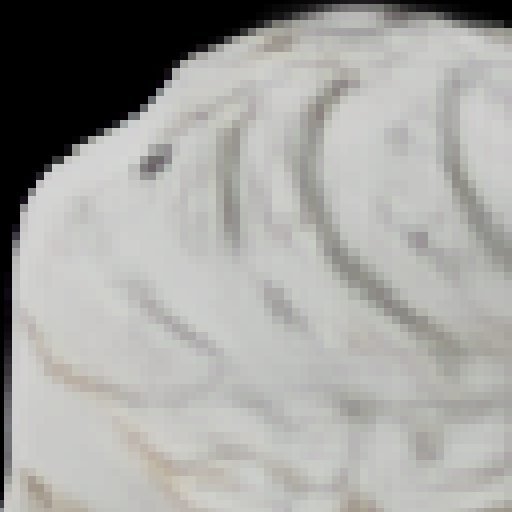} \\
  \end{tabular}
  \caption{The arms of Table~\ref{tab:ablations} and the full-scale models on the first and third objects of
  Figure~\ref{fig:qualitative}: the first object (top), and $4\times$ close-ups of the region of each of the two objects
  with the most detail (middle and bottom). The final model keeps the painted pattern of the first object, and the
  same recipe trained without the footprint renders it slightly softer. The models of the 1,000-object harness,
  trained for about a tenth as long, are softer, and lose much of the pattern without the footprint, without the 2D
  embedding and with $8 \times 8$ patches. The last two columns are the recipe before the footprint, trained on all
  the data for 94,000 steps, with and without the 2D embedding.}
  \label{fig:ablation_renders}
\end{figure}

\subsection{Spread over objects}
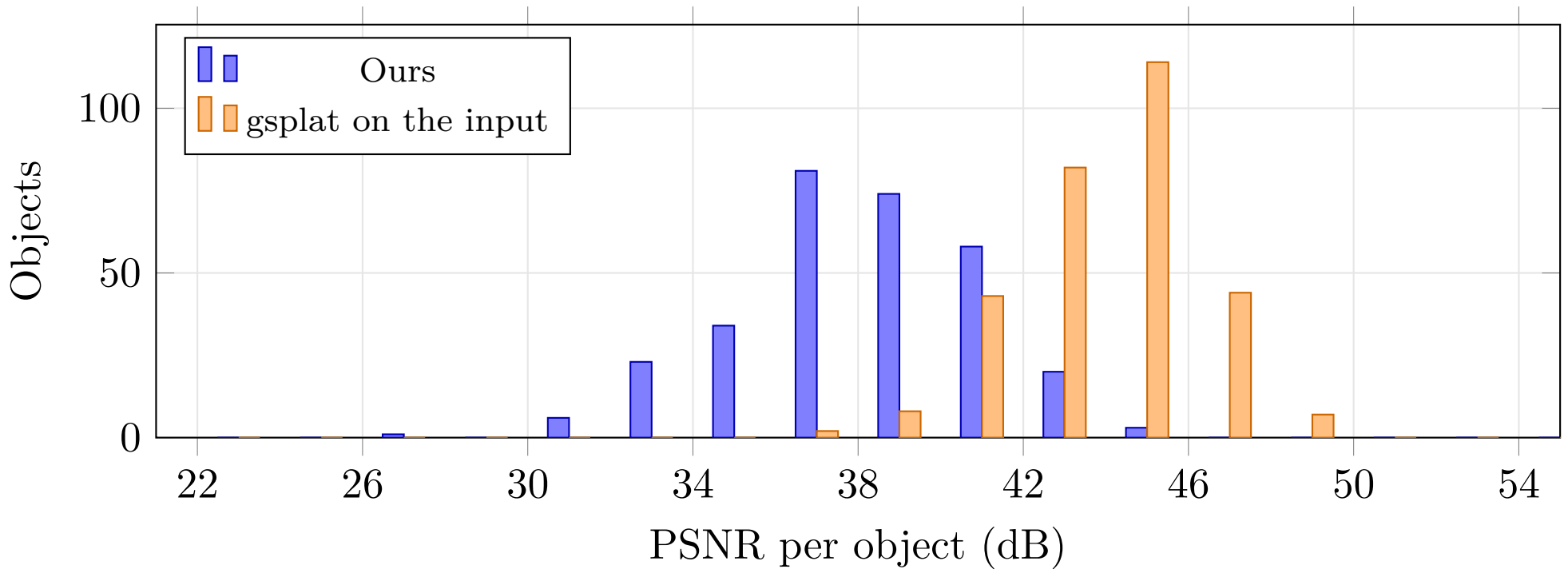
\begin{figure}[ht]
  \centering
  \begin{tikzpicture}
    \begin{axis}[width=0.8\textwidth, height=5cm, ybar=0pt, bar width=5pt, xlabel={PSNR per object (dB)},
                 ylabel={Objects}, xmin=21, xmax=55, ymin=0, xtick={22,26,...,54},
                 legend style={font=\footnotesize, at={(0.02,0.97)}, anchor=north west}, grid=major, grid style={black!10}]
      \addplot[fill=blue!50, draw=blue!70!black] table[x expr=\thisrow{lo}+1, y=model] {figures/data/heldout_hist.dat};
      \addplot[fill=orange!50, draw=orange!80!black] table[x expr=\thisrow{lo}+1, y=gsplat] {figures/data/heldout_hist.dat};
      \legend{Ours, gsplat on the input}
    \end{axis}
  \end{tikzpicture}
  \caption{PSNR of each of the 300 fresh held-out objects (mean over its 28 views), in bins of 2 dB.}
  \label{fig:histogram}
\end{figure}

Figure~\ref{fig:histogram} shows how the PSNR is spread over the 300 fresh held-out objects, over all 28 views. The
model ranges from 26.5 to 45.0 dB, with a median of 38.2 dB, and gsplat on the input ranges from 36.4 to 49.3 dB.

\subsection{Error on edges}
\label{sec:appendix_edges}
We rank the pixels of each held-out crop by the Sobel gradient of the gray reference. On the 10\% of pixels with the
strongest edges, the model scores 30.04 dB, against 45.19 dB on the rest, and these pixels hold 71\% of its squared
error. Rasterizing the input shows the same pattern, with 78\% of its error on these pixels. The model trails the
rasterizer by 6.2 dB on the edges and by 8.1 dB on the rest of the object. In the frequency domain, 72\% of the
model's error energy lies above 1/8 cycles per pixel, but its error also exceeds the rasterizer's at low frequencies
(Table~\ref{tab:frequency}).

\subsection{Error by frequency band}
\label{sec:appendix_frequency}
Table~\ref{tab:frequency} splits the error of each held-out view into frequency bands. We take the square crop
around the object at its native resolution, apply a Hann window, and sum the squared magnitude of the 2D Fourier
transform over octave bands of radial frequency. For each band we report the share of each renderer's error energy, its
in-phase gain, the part of the reference's content that it reproduces ($g = \mathrm{Re}\sum F_x F_r^{*} / \sum
|F_r|^2$, 1 when all of it is kept), its unmatched power, the content that is not in the reference ($\sum |F_x - g
F_r|^2 / \sum |F_r|^2$, 0 when there is none), and the model's error energy divided by the rasterizer's. Means over
the 1,200 views of the 300 held-out objects.

\begin{table}[ht]
  \centering
  \caption{Error by frequency band, \method{} against gsplat on the same pruned input, both compared with the unpruned
  render. Frequencies in cycles per pixel, the last band includes the corners of the spectrum.}
  \label{tab:frequency}
  \small
  \begin{tabular}{@{}lccccccc@{}}
    \toprule
    & \multicolumn{2}{c}{Share of error} & \multicolumn{2}{c}{In-phase gain} & \multicolumn{2}{c}{Unmatched power} & Error, Ours \\
    Band & Ours & gsplat & Ours & gsplat & Ours & gsplat & over gsplat \\
    \midrule
    0 to 1/64 & 0.046 & 0.015 & 0.997 & 1.001 & 0.000 & 0.000 & 16.4 \\
    1/64 to 1/32 & 0.039 & 0.022 & 0.999 & 1.000 & 0.002 & 0.000 & 9.1 \\
    1/32 to 1/16 & 0.069 & 0.049 & 0.997 & 0.999 & 0.006 & 0.001 & 6.7 \\
    1/16 to 1/8 & 0.131 & 0.113 & 0.989 & 0.997 & 0.015 & 0.003 & 5.4 \\
    1/8 to 1/4 & 0.270 & 0.255 & 0.960 & 0.985 & 0.048 & 0.010 & 4.9 \\
    1/4 and above & 0.445 & 0.546 & 0.845 & 0.928 & 0.162 & 0.044 & 4.1 \\
    \bottomrule
  \end{tabular}
\end{table}

At the highest frequencies the model reproduces 85\% of the reference's content, against 93\% for the rasterizer, and
adds almost four times as much content that is not in the reference: it both blurs and adds spurious detail. This is
not a small misalignment: if every $16 \times 16$ tile of an image may shift by up to 2 pixels to best match the
reference, the model's error drops by only 4\%, and the rasterizer's by 0.7\%. Its excess error over the rasterizer
is largest, relative to the rasterizer's, at the lowest frequencies. This is not a global error in brightness or
contrast: fitting a gain and a bias per view and color channel to the reference raises the model's PSNR by only 0.05
dB. It is error at the scale of the object's parts.

\clearpage
\subsection{Error by overlap}
\label{sec:appendix_overlap}
A typical pixel of an object is reached by about 130 of its 20,000 Gaussians, and 77\% of the object's pixels are
reached by 32 or more. Table~\ref{tab:overlap} groups the pixels of each held-out view by the number of input Gaussians that reach them,
counted as the Gaussians whose opacity at that pixel is at least $1/255$, the ones a rasterizer blends there,
without stopping at opaque ones. Both renderers score best on pixels reached by a single Gaussian, and lowest on
pixels reached by 16 to 63, which lie mostly on the silhouette and on thin parts. The gap between the two is
smallest where one to three Gaussians reach a pixel.

\begin{table}[ht]
  \centering
  \caption{PSNR in dB against the unpruned render, by the number of input Gaussians reaching a pixel, over all
  object pixels of the 1,200 held-out views.}
  \label{tab:overlap}
  \small
  \begin{tabular}{@{}lrrrr@{}}
    \toprule
    Gaussians per pixel & Share of pixels & Ours & gsplat & Difference \\
    \midrule
    1 & 7.5\% & 53.30 & 57.36 & 4.06 \\
    2 to 3 & 5.6\% & 46.95 & 49.42 & 2.48 \\
    4 to 7 & 4.3\% & 35.18 & 43.25 & 8.07 \\
    8 to 15 & 3.1\% & 32.01 & 37.38 & 5.37 \\
    16 to 31 & 2.8\% & 30.25 & 33.84 & 3.59 \\
    32 to 63 & 5.0\% & 28.99 & 34.94 & 5.95 \\
    64 to 127 & 22.5\% & 31.42 & 38.53 & 7.11 \\
    128 to 255 & 41.9\% & 33.05 & 40.65 & 7.60 \\
    256 or more & 7.4\% & 34.71 & 42.97 & 8.26 \\
    \bottomrule
  \end{tabular}
\end{table}

\clearpage
\subsection{Failure cases}
\label{sec:appendix_failures}
Figure~\ref{fig:failures} shows the four held-out objects where the model does worst.

\newcommand{\failimg}[1]{\includegraphics[width=0.19\textwidth]{figures/failures/#1}}
\begin{figure}[ht]
  \centering
  \setlength{\tabcolsep}{1pt}
  \footnotesize
  \begin{tabular}{@{}ccccc@{}}
    Reference & gsplat on the input & Ours & Error, gsplat & Error, Ours \\
    \failimg{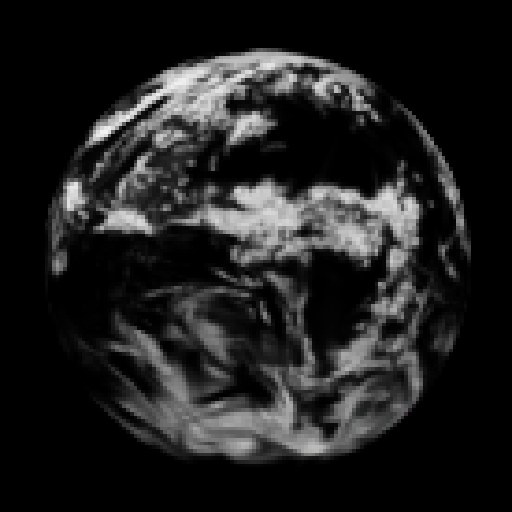} & \failimg{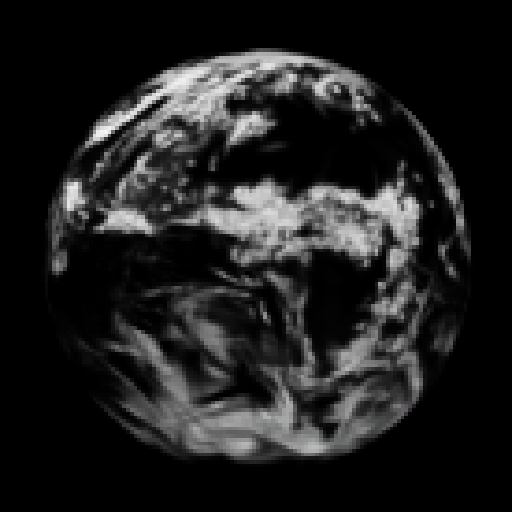} & \failimg{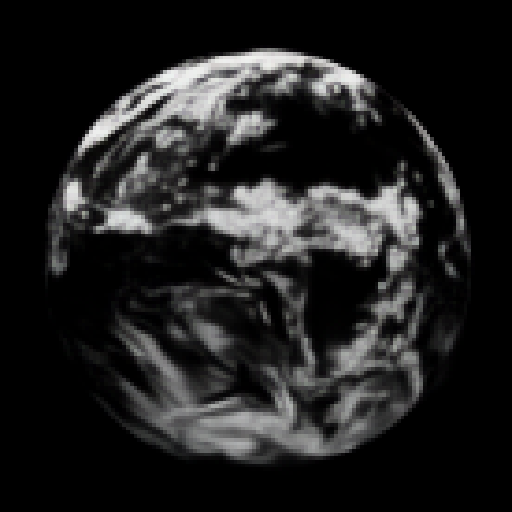} & \failimg{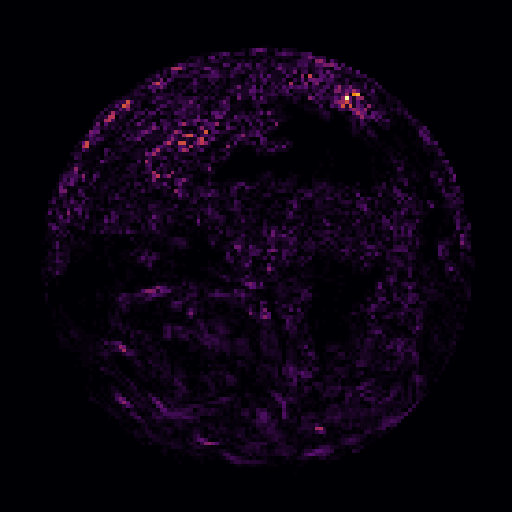} & \failimg{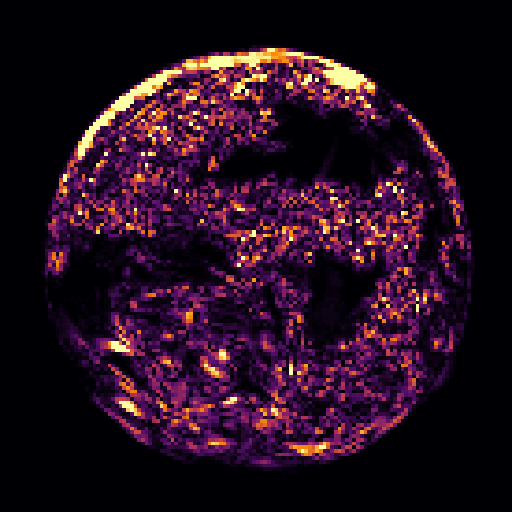} \\
    \failimg{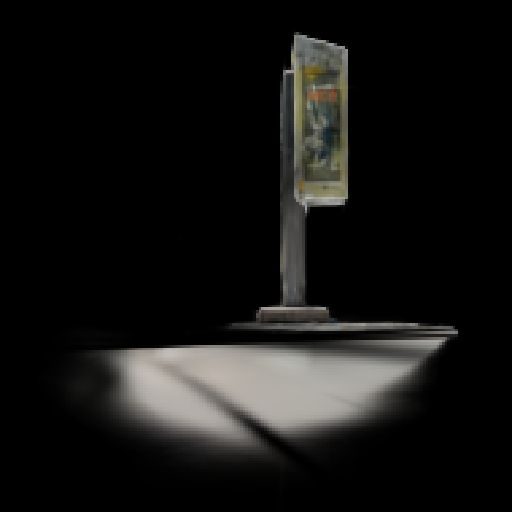} & \failimg{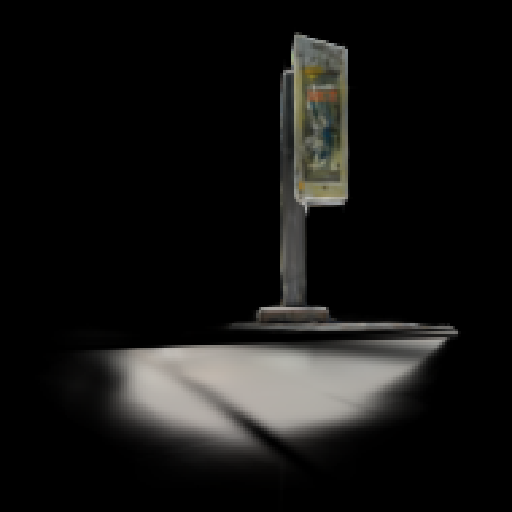} & \failimg{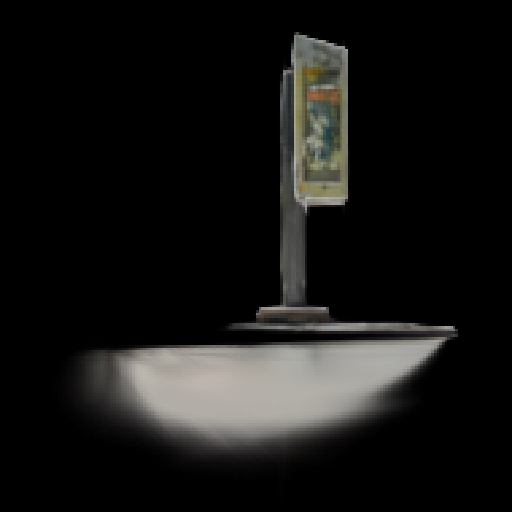} & \failimg{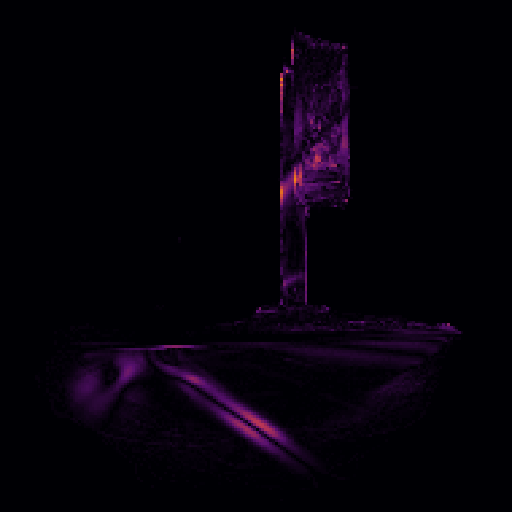} & \failimg{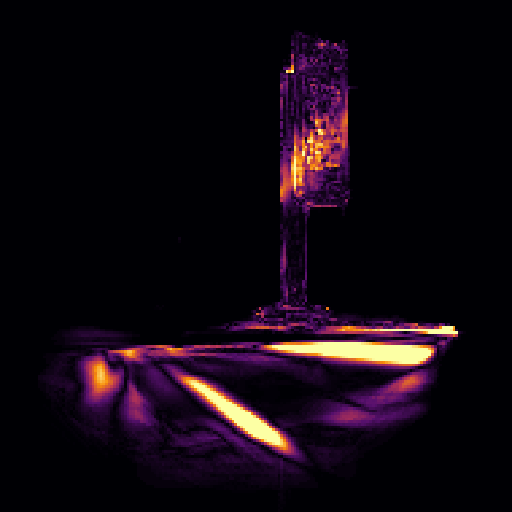} \\
    \failimg{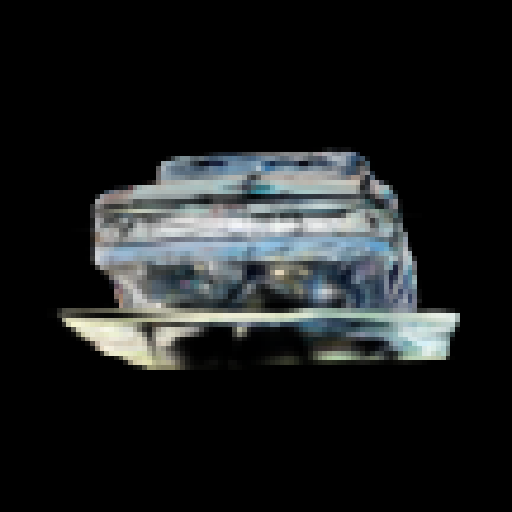} & \failimg{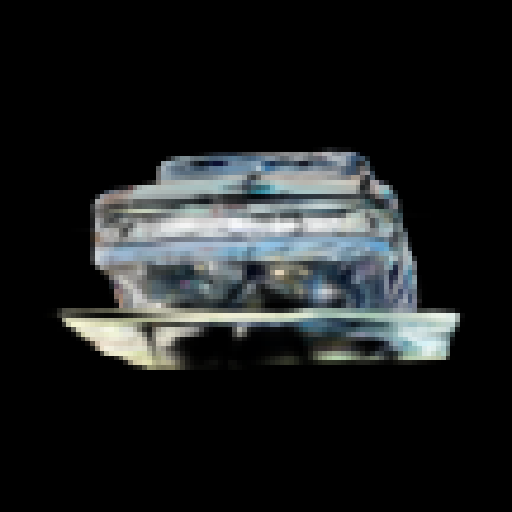} & \failimg{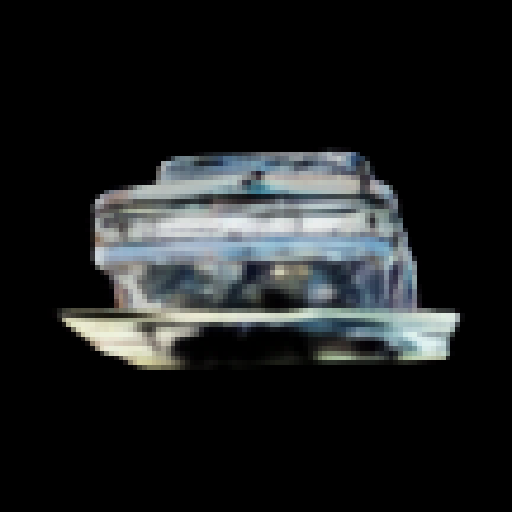} & \failimg{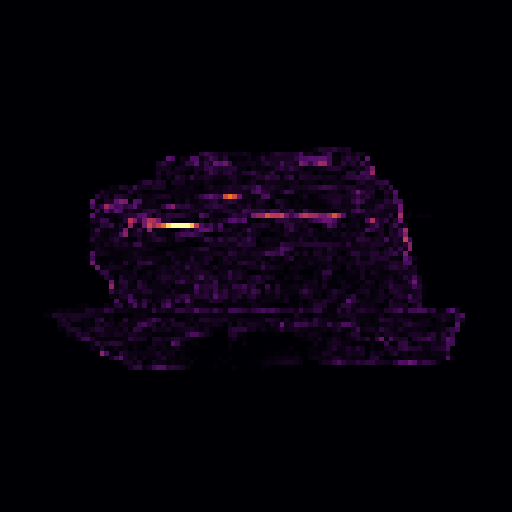} & \failimg{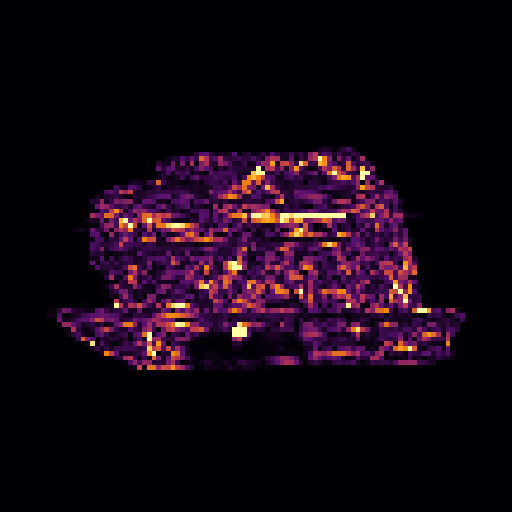} \\
    \failimg{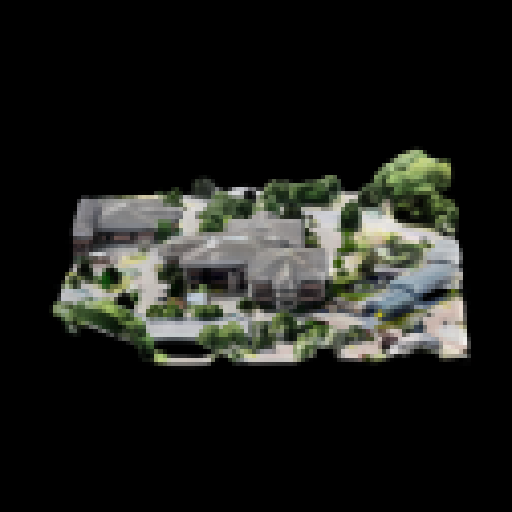} & \failimg{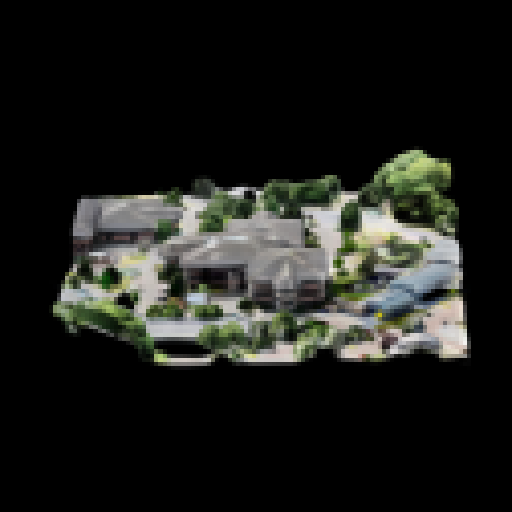} & \failimg{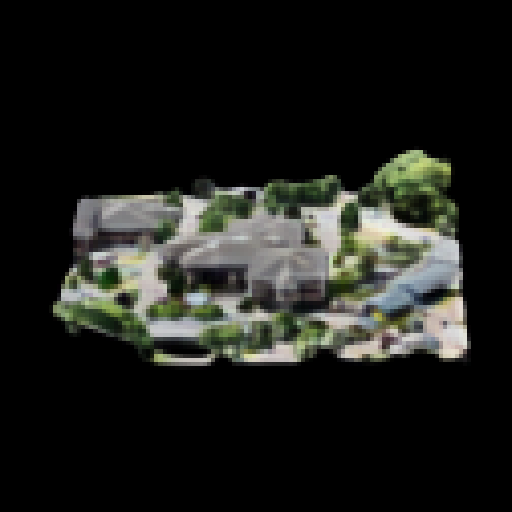} & \failimg{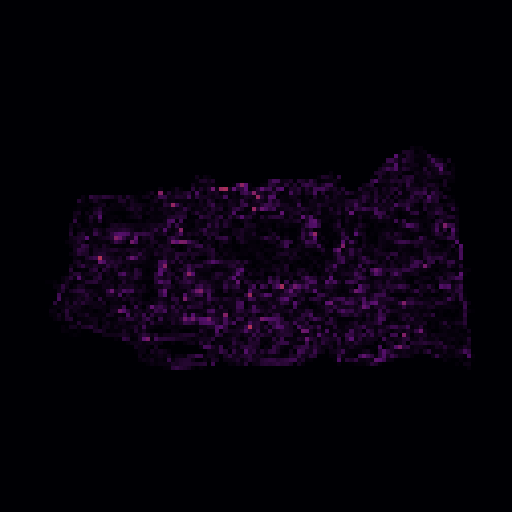} & \failimg{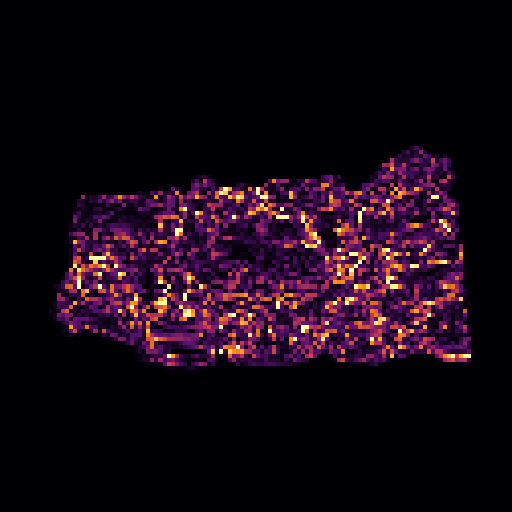} \\
  \end{tabular}
  \caption{Failure cases. The four fresh held-out objects with the lowest mean PSNR of \method{}, each at its worst
  view. Columns as in Figure~\ref{fig:qualitative}, with errors scaled by 5. \method{} scores 25.4, 27.2, 29.1 and
  28.5 dB on these views, and gsplat on the input 37.9, 40.0, 38.6 and 40.4 dB. The shapes are right on all four. The
  error comes from dense texture that \method{} renders less precisely than the reference (the clouds on the globe,
  the facade of the building, the houses and trees of the town) and from soft, low-contrast structure (the streaks on
  the ground under the sign).}
  \label{fig:failures}
\end{figure}

\section{The road to the final model}
\label{sec:appendix_history}

Several choices in Section~\ref{sec:method} only make sense in light of what did not work, so we summarize the
experiments that led to them. Early results used other protocols, so we report most of them only qualitatively.

\subsection{Data}
Our first training set followed RenderFormer's examples, meshes sampled into Gaussians with RenderFormer's own
renders as the target. The model learned to imitate RenderFormer's lighting instead of rendering its input, so we
switched to real splats rendered by gsplat. Scenes composed of several small splats then plateaued whatever we
changed, and tests on outside captures showed that the quality of the input mattered more than its size. We moved to
single Objaverse\_Splats objects, from 2,667 to 13,405 and finally 26,820, and added the fine-tune of the pruned
input (Section~\ref{sec:data}).

\subsection{Training schedule}
Training the input layer alone first was part of our design from the start, since RenderFormer's pretrained layers
expect its own triangle tokens. A series of runs that skipped this stage because of a bug all stalled at the same
loss, whatever encoding of the Gaussian parameters they used, and restoring the stage fixed them. The stage ablation
of Section~\ref{sec:exp_ablations} tests this directly. Larger inputs (20,000 Gaussians instead of 5,000), random rotations of every
sample and much longer training with a 256 to 512 pixel curriculum each helped clearly. The best model of this phase,
trained on 13,405 objects, scored 30.39 dB on our held-out set.

\subsection{The decoder}
That model scored only 0.55 dB better on its training objects than on held-out ones, models trained on 10 to 1,000
objects could not fit even their own training objects, and more layers or a wider network did not help
(Appendix~\ref{sec:appendix_negative}). We therefore tested changes to the view decoder on 10 training objects. The
2D rotary embedding of projected positions was the first change that improved both training and held-out objects.
Smaller patches improved them further, and became affordable once the view decoder ran in bf16 with gradient
checkpointing, which made a training step about three times faster. We also tried windowed cross-attention at this
point and later dropped it (Section~\ref{sec:exp_ablations}). Showing
every object from three distances mainly improved close cameras. The final model combines these changes and
trains on all 26,820 objects for 375,000 steps, from the second-stage checkpoint of the previous phase
(Appendix~\ref{sec:appendix_reproducibility}).

\section{Changes that did not help}
\label{sec:appendix_negative}

Table~\ref{tab:negative} lists changes we tried that did not improve the model. Most were tested on small training
sets with earlier versions of the model, so they do not rule a change out at full scale, but none of them justified a
full run. Margins are the PSNR of the rasterized input minus the model's, so lower is better.

\begin{table}[ht]
  \centering
  \caption{Changes that did not help, with the setup they were tested in. ``Fit'' is the margin on the training
  objects and ``held-out'' the margin on the 300 held-out objects, in dB.}
  \label{tab:negative}
  \small
  \begin{tabular}{@{}p{0.37\textwidth}p{0.25\textwidth}p{0.3\textwidth}@{}}
    \toprule
    Change & Setup & Effect \\
    \midrule
    More depth: 18 + 9 layers (287M parameters) & 100 objects, earlier model & Fit 13.43 vs 13.40 \\
    Wider feed-forward network ($8\times$, 322M) & 100 objects, earlier model & Fit 13.35 vs 13.40 \\
    Small MLP as input layer, log-scale input & 100 objects, earlier model & Fit 13.43 vs 13.40 \\
    Training from scratch & Several widths, with and without a 256-pixel warm-up & Never left a poor plateau \\
    Changing the pretrained position embedding (doubled frequency) & 10 objects & Destroyed the pretrained behavior \\
    Geometry-biased cross-attention & 100 objects & Fit and held-out unchanged, the model learned to ignore it \\
    Proximity bias and depth features in cross-attention & 10 objects & Fit 7.60 vs 7.57, held-out 20.45 vs 20.47 \\
    Log-covariance input & 10 objects & Held-out $+0.04$ dB \\
    Rotary embedding of the endpoints of each Gaussian's projected axes (unlike the footprint on the keys, Section~\ref{sec:method_decoder}) & 1 object & Within 0.1 dB \\
    Rotary embedding of values & 10 objects & Held-out unchanged, 40\% slower \\
    $2 \times 2$-pixel patches & 10 objects & Held-out $-0.50$ dB, at 4 to 5 times the cost \\
    LPIPS off (log-L1 only) & 10 objects & Better fit, held-out PSNR 0.21 dB lower \\
    Several views of each object per step & 1,000 objects, equal time & $+0.01$ dB \\
    Windowed cross-attention (each tile of patches sees only the Gaussians that reach it) & 1,000 objects & Held-out 34.62 against 34.42 dB without it (32.82 against 32.71 before the footprint), 8\% less time per step, 25.3 against 20.1 GB of memory, 2 to 7\% faster at up to 640,000 Gaussians \\
    Per-object fine-tuning with LoRA adapters~\citep{hu2022lora} & Single objects & Worse than rasterization at equal storage \\
    \bottomrule
  \end{tabular}
\end{table}

Per-object training, which can beat the rasterizer on single objects but does not transfer, is described in
Appendix~\ref{sec:appendix_per_object}.

\section{Training on a single object}
\label{sec:appendix_per_object}

Our main model shares one set of weights across all objects. Here we ask what the same architecture can do when it
only has to render one object, which separates what the network can represent from what it can learn to generalize.
All of these experiments used earlier versions of the model.

Trained on a single object with 9,000 views of its full splat, an earlier version of the model ended above
rasterizing the pruned input on 4 of the 13 objects we tried, by up to 1.8 dB.
It does so by learning the object and filling in what pruning removed, not by rendering its input better: the one
such model we kept scores 44.59 dB against the full splat but only 42.18 dB against a rasterization of its own input.
These runs show how much detail the network can output for one object, not what sharing the weights costs.

\subsection{How much detail can the network output}
We first trained the model on a single object, a scan of tomatoes, on its 14 training views and without augmentation,
until it had memorized them. On those views it reached 52.5 to 53.0 dB on average against the full splat, with an LPIPS of
0.0003, and its renders cannot be told apart from the reference (Figure~\ref{fig:tomatoes}). The decoder can
therefore output this much detail when it is trained to, so the error of our general model is not a hard limit of
its output head. This experiment used an earlier input: the 219,000-Gaussian scan pruned to 20,000 with the same
importance score, but without the fine-tune of Section~\ref{sec:data}, so rasterizing it gives a blurry image. On new
views between the training views, the model's renders are much sharper than that rasterization. They are closer to
the full splat in PSNR (30.5 against 29.7 dB) and far closer in LPIPS (0.09 against 0.15 on the crops of
Figure~\ref{fig:tomatoes}), and far from the rasterization of the model's own input (27.4 dB). From 14 views of the
full splat, the model learned enough of the object to restore much of what pruning removed. It did not learn all of
it: on new views it stays far below its 52.5 dB on the training views.

\newcommand{\tomimg}[1]{\includegraphics[width=0.2\textwidth]{figures/tomatoes/#1}}
\begin{figure}[ht]
  \centering
  \setlength{\tabcolsep}{1pt}
  \footnotesize
  \begin{tabular}{@{}cccc@{}}
    & Reference & gsplat on the input & Trained on this object \\
    \raisebox{0.09\textwidth}{\rotatebox{90}{Training view}} & \tomimg{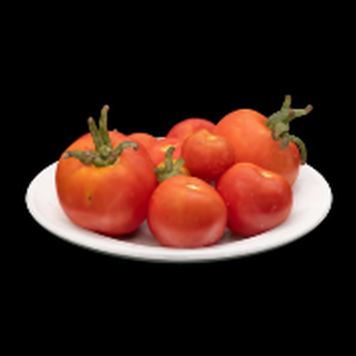} & \tomimg{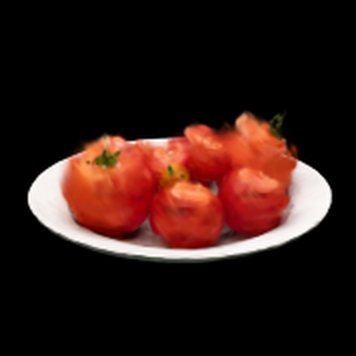} & \tomimg{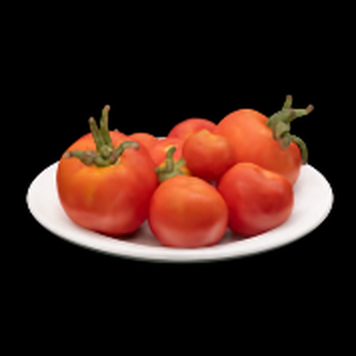} \\
    & & 29.2 dB & 51.0 dB \\
    \raisebox{0.09\textwidth}{\rotatebox{90}{New view}} & \tomimg{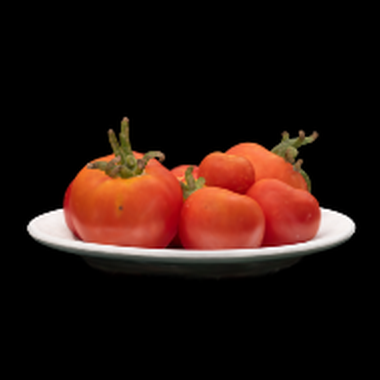} & \tomimg{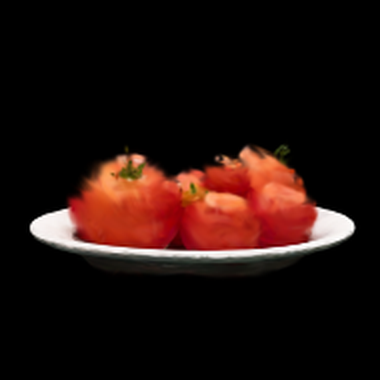} & \tomimg{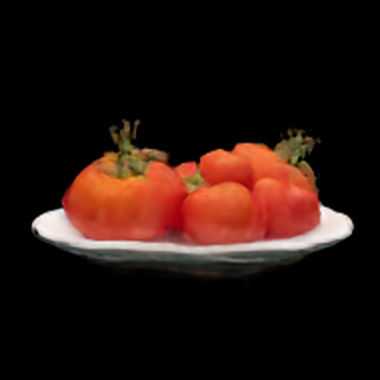} \\
    & & 29.9 dB & 30.2 dB \\
    \raisebox{0.09\textwidth}{\rotatebox{90}{New view}} & \tomimg{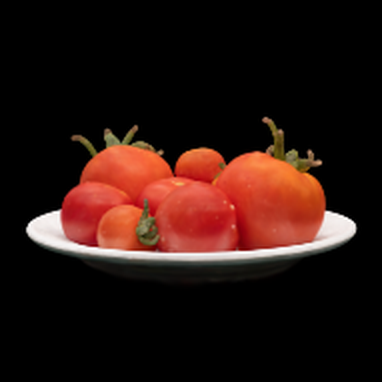} & \tomimg{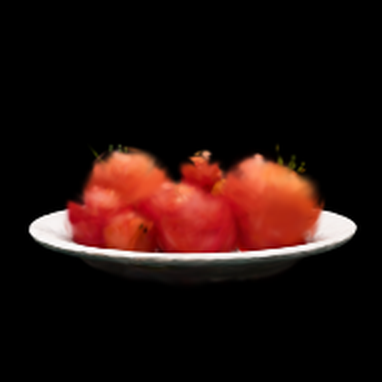} & \tomimg{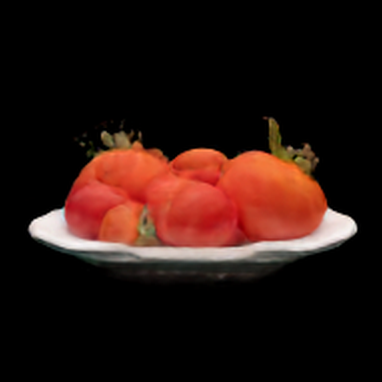} \\
    & & 29.4 dB & 30.9 dB \\
  \end{tabular}
  \caption{A model trained on 14 views of one object until it memorized them. On a training view (top) its render
  cannot be told apart from the reference. On new views between the training views (middle and bottom), it is much
  sharper than rasterizing its input, and restores much of the detail that pruning removed. This experiment used an
  earlier input, pruned without the fine-tune of Section~\ref{sec:data}, which is why the rasterized input is
  blurry. PSNR over the whole frame, on this view (52.5 to 53.0 dB on average over the 14 training views). Tomatoes scan by
  simonbethke, CC BY 4.0.}
  \label{fig:tomatoes}
\end{figure}

\subsection{Compensating for pruning on one object}
With thousands of views of the object instead of 14, per-object training can beat the rasterizer on new views. The
training images are renders of the full splat, so the model learns that one object and can fill in what pruning
removed from its input. This is a prior on the object, not a better rendering of the input. We trained on 9,000 random views per object, in cycles of 30,000 steps with a cosine learning rate, and
evaluated on 40 views not used in training (24 random, 8 close-up and 8 distant), against gsplat on the same
20,000-Gaussian input. Table~\ref{tab:per_object} and Figure~\ref{fig:slipper} show the results. Only on 4 of the 13
objects we tried did the model end above the rasterizer on average, by up to 1.8 dB, and most of its gain was in
close-up views, where the pruned input shows its gaps. On the other 9 it stayed below, by 1 to 9 dB, typically on
simple objects that the 20,000 Gaussians already render almost perfectly.

\begin{table}[ht]
  \centering
  \caption{Per-object training: PSNR of the model minus gsplat on the same 20,000-Gaussian input, mean over 40 views
  not used in training (higher is better), and the number of views where the model is better.}
  \label{tab:per_object}
  \small
  \begin{tabular}{@{}lllrc@{}}
    \toprule
    Object & Source & Training & Difference (dB) & Views better \\
    \midrule
    Slipper & real scan, 1.17M Gaussians & 120,000 steps & $+1.82$ & 35 / 40 \\
    Molecule & Objaverse\_Splats & 150,000 steps & $+1.47$ & 29 / 40 \\
    Tomatoes & real scan, 219,000 Gaussians & several cycles & $+1.30$ & 34 / 40 \\
    Tent & Objaverse\_Splats & 150,000 steps & $+0.71$ & 28 / 40 \\
    Octopus & real scan, 1.84M Gaussians & 90,000 steps & $-1.21$ & 11 / 40 \\
    8 other objects & Objaverse\_Splats & 60,000 steps & $-2.1$ to $-9.1$ & 0 to 7 / 40 \\
    \bottomrule
  \end{tabular}
\end{table}

\newcommand{\slipperimg}[1]{\includegraphics[width=0.2\textwidth]{figures/slipper/#1}}
\begin{figure}[ht]
  \centering
  \setlength{\tabcolsep}{1pt}
  \footnotesize
  \begin{tabular}{@{}ccc@{}}
    Reference & gsplat on the input & Trained on this object \\
    \slipperimg{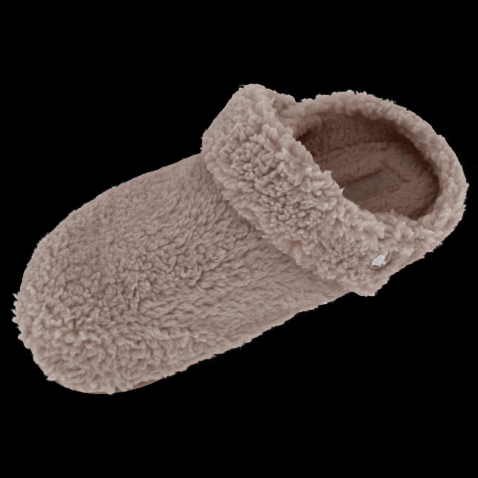} & \slipperimg{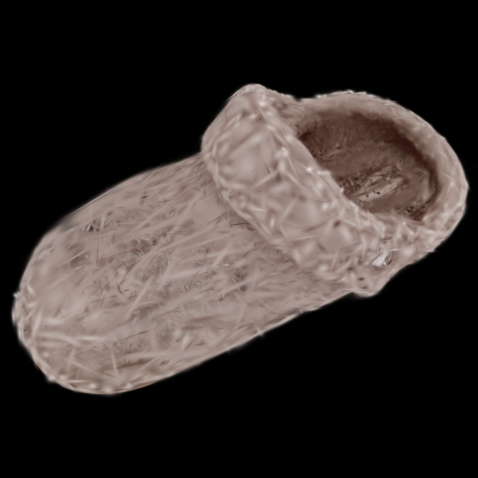} & \slipperimg{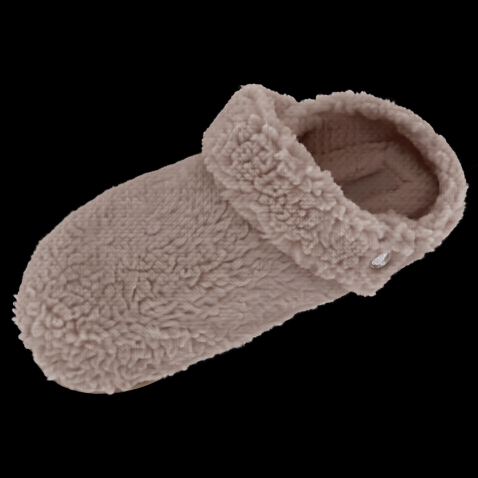} \\
    & 26.8 dB & 31.5 dB \\
    \slipperimg{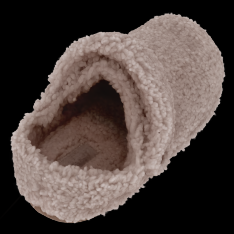} & \slipperimg{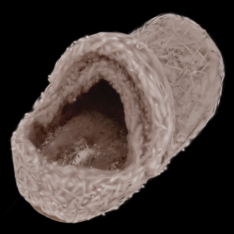} & \slipperimg{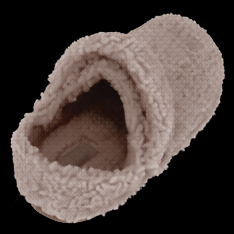} \\
    & 34.5 dB & 36.5 dB \\
    \slipperimg{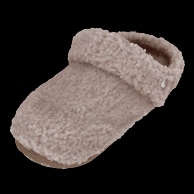} & \slipperimg{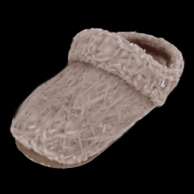} & \slipperimg{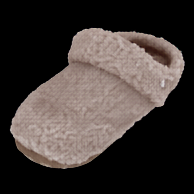} \\
    & 37.6 dB & 39.6 dB \\
  \end{tabular}
  \caption{A model trained on the slipper scan alone, on three views not used in training: close-up (top), at a
  random distance (middle) and distant (bottom). The reference is the full scan of 1.17M Gaussians, and both gsplat
  and the model get it pruned to 20,000. Close up, the pruned input shows long streaks where detail was removed,
  which the model fills in with the fleece texture. Slipper scan by Andre, CC BY 4.0.}
  \label{fig:slipper}
\end{figure}

Of these models, we kept only the one trained on the tomatoes. On the same 40 views, it scores 44.59 dB against the
full splat and 42.18 dB against a rasterization of its own input, while that rasterization scores 43.29 dB against the
full splat. The model is further from its own input than from the full splat on all 40 views, and most of all close
up (37.90 against 41.90 dB). It has learned to reproduce the object, not to render its input more faithfully.

\subsection{Why we do not pursue it}
These results show that the network can output far more detail than our general model does, and that trained on one
object it can learn to fill in what pruning removed. We do not consider per-object training a practical use of it,
for three reasons. It does not transfer to shared weights (Section~\ref{sec:exp_scaling}). It is
expensive to store and to run: every object needs its own copy of the 195M-parameter network, about 745 MB, to
improve on an input of 20,000 Gaussians that takes 1.1 MB, and rendering a view took 476 ms against 1 ms for gsplat
in our measurement on an NVIDIA A40. It is also slow to train, with 4 to 5 cycles of 30,000 steps per object. We tried
to cut the storage with low-rank adapters (LoRA) on a shared network, but they fell short of the rasterizer at equal
storage. The gain, under 2 dB on the objects where it works, does not justify this cost.

\section{Reproducibility}
\label{sec:appendix_reproducibility}
\label{sec:appendix_details}

The code, the trained weights, the checkpoint after the first two stages, the lists of training and evaluation
objects, and the scripts for every experiment in this paper are available at
\url{https://github.com/SVLwoof/gaussianformer}. Its file \texttt{paper/README.md} gives the command and the expected numbers for every table and
figure. The three trained models are on Hugging Face: the final model (\texttt{shahafvl/gaussianformer-v22}), the
model without the footprint (\texttt{shahafvl/gaussianformer-v21}) and the checkpoint after stage 2
(\texttt{shahafvl/gaussianformer-256px}). Table~\ref{tab:stages} describes the released
training recipe. The final model was produced by that recipe, with two differences in its first two stages, which
were run during earlier phases of the project. Its first stage was trained on 13,405 objects instead of 26,820, and
its first two stages saw only the 14 views at distance 1.7, not all 28. Starting from the released checkpoint after
the second stage reproduces stages 3 and 4 exactly, which is about 620 of the 1,327 GPU-hours. Running all four stages
from RenderFormer's weights with the released recipe is documented, but we have not run it end to end.

The stages of the final model took 46 GPU-hours (stage 1) and 662 GPU-hours (stage 2) on NVIDIA L40S GPUs, 592
GPU-hours for stages 3 and 4 up to 375,000 steps (91 on A40 and 501 on RTX Pro 6000 GPUs, as the cluster allowed)
and 27 GPU-hours for the final decay on L40S GPUs.

Table~\ref{tab:hyperparameters} lists the settings of the final model. All of them are options of the released
code.

\begin{table}[ht]
  \centering
  \caption{Settings of the final model.}
  \label{tab:hyperparameters}
  \small
  \begin{tabular}{@{}lp{0.52\textwidth}@{}}
    \toprule
    \multicolumn{2}{@{}l}{\textit{Model}} \\
    Token width, attention heads, feed-forward width & 768, 6, 3,072 \\
    Scene encoder & 12 layers, 16 register tokens, 3D RoPE on Gaussian centers \\
    View decoder & 6 layers, $4 \times 4$-pixel patches, DPT head with 128 features \\
    Ray embedding & pretrained $8 \times 8$ embedding, rays upsampled from $4 \times 4$ \\
    2D RoPE of projected positions & 16 channel pairs (32 channels) per head, patch coordinates scaled by 0.25 \\
    Projected footprint & 4 numbers per Gaussian on the cross-attention keys, a zero-initialized linear layer per decoder layer (23,040 parameters) \\
    Cross-attention & dense, every patch attends to every Gaussian \\
    Parameters & 194.9M (scene encoder 113.3M, view decoder 81.6M) \\
    \midrule
    \multicolumn{2}{@{}l}{\textit{Training}} \\
    Optimizer & AdamW, weight decay 0.01, gradient clipping at 1.0 \\
    Learning rate & $5 \times 10^{-5}$ ($10^{-3}$ for stage 1), 1,000 warmup steps, cosine decay to 1\% \\
    Batch & 8 views, one per GPU, 8 GPUs \\
    Precision & bf16 autocast, gradient checkpointing in the view decoder \\
    Loss & $(1-\lambda)$ weighted log-L1 $+ \lambda$ LPIPS (VGG), $\lambda = 0.5$, background weight 0.05 \\
    Augmentation & one random 3D rotation of object and camera per sample \\
    Sampling & one random view of every object per epoch, 28 views per object \\
    Stage 4 schedule & 375,000 steps at a constant learning rate, then 10,000 steps of decay \\
    Hardware and cost & NVIDIA L40S, A40 and RTX Pro 6000 GPUs, 1,327 GPU-hours in total \\
    \bottomrule
  \end{tabular}
\end{table}

\subsection*{Licenses}
Objaverse\_Splats states no license of its own. Its splats are fit to Objaverse objects, a collection released under
ODC-By 1.0 in which each object keeps its creator's Creative Commons license. We release our code and trained weights
under the MIT license, as RenderFormer does.

\end{document}